\documentclass[runningheads]{llncs}

\usepackage{eccv24toolkit/eccv}

\usepackage{eccv24toolkit/eccvabbrv}
\usepackage{graphicx}
\usepackage{booktabs}
\usepackage[accsupp]{axessibility}

\usepackage{amsmath,amssymb,amsfonts}
\usepackage{multirow,multicol}
\usepackage{algorithmic}
\usepackage{textcomp}
\usepackage{xspace}
\usepackage{url}
\usepackage{dsfont}
\usepackage{balance}
\usepackage{soul}
\usepackage{todonotes}
\usepackage{pifont}
\usepackage{colortbl}
\usepackage{overpic}
\usepackage{enumitem}
\usepackage{nccmath}
\usepackage{courier}
\usepackage{array}
\usepackage{tabularray}
\usepackage{pifont}
\usepackage{microtype}

\usepackage[dvipsnames]{xcolor}
\usepackage{dsfont}
\usepackage{enumitem}
\usepackage{wrapfig}
\usepackage{lipsum}
\usepackage{scrextend}
\usepackage{pict2e}

\usepackage{hyperref}
\usepackage{orcidlink}

\begin{document}

\pagestyle{headings}
\mainmatter

\newcommand{\cmark}{\ding{51}}
\newcommand{\xmark}{\ding{55}}

\definecolor{myblue}{rgb}{0.0000,0.7490,1.0000}
\definecolor{myyellow}{rgb}{1.0000,0.7530,0.0000}
\definecolor{mygray}{rgb}{0.9000,0.9000,0.9000}
\definecolor{myazure}{rgb}{0.8509,0.8980,0.9412}
\definecolor{mygreen}{rgb}{0.1333,0.5451,0.1333}
\definecolor{bisque}{rgb}{1.0, 0.89, 0.77}
\definecolor{cosmiclatte}{rgb}{1.0, 0.97, 0.91}

% Support for easy cross-referencing
\crefname{section}{Sec.}{Secs.}
\Crefname{section}{Section}{Sections}
\Crefname{table}{Table}{Tables}
\crefname{table}{Tab.}{Tabs.}

\newcommand{\fabiocomment}[1]{\todo[color=purple!20, inline, author=Fabio]{#1}}
\newcommand{\davide}[1]{\todo[color=blue!20, inline, author=Davide]{#1}}
\newcommand{\andreacomment}[1]{\todo[color=green!20, inline, author=Andrea]{#1}}
\newcommand{\amircomment}[1]{\todo[color=yellow!20, inline, author=Amir]{#1}}
\newcommand{\warning}[1]{\textbf{\color{red!90}{#1}}}

\newcommand{\ourmethod}{FreeZe\xspace}

\title{{\color{NavyBlue}\ourmethod}: Training-{\color{NavyBlue}free ze}ro-shot 6D pose estimation with geometric and vision foundation models}
\titlerunning{FreeZe: Training-free zero-shot 6D pose estimation}

\author{Andrea Caraffa\inst{1}%\orcidlink{009-0004-7775-924X}
\and
Davide Boscaini\inst{1}%\orcidlink{0000-0003-4887-2038}
\and
Amir Hamza\inst{1,2}%\orcidlink{1111-2222-3333-4444}
\and
Fabio Poiesi \inst{1}%\orcidlink{0000-0002-9769-1279}
}

\authorrunning{A.~Caraffa et al.}

\institute{Fondazione Bruno Kessler, Trento, Italy 
\email{\{acaraffa,dboscaini,ahamza,poiesi\}fbk.eu}
\and
University of Trento, Italy}

\maketitle

\begin{abstract}
Estimating the 6D pose of objects unseen during training is highly desirable yet challenging.
Zero-shot object 6D pose estimation methods address this challenge by leveraging additional task-specific supervision provided by large-scale, photo-realistic synthetic datasets.
However, their performance heavily depends on the quality and diversity of rendered data and they require extensive training.
In this work, we show how to tackle the same task but without training on specific data.
We propose \ourmethod, a novel solution that harnesses the capabilities of pre-trained geometric and vision foundation models.
\ourmethod leverages 3D geometric descriptors learned from unrelated 3D point clouds and 2D visual features learned from web-scale 2D images to generate discriminative 3D point-level descriptors.
We then estimate the 6D pose of unseen objects by 3D registration based on RANSAC.
We also introduce a novel algorithm to solve ambiguous cases due to geometrically symmetric objects that is based on visual features.
We comprehensively evaluate \ourmethod across the seven core datasets of the BOP Benchmark, which include over a hundred 3D objects and 20,000 images captured in various scenarios.
\ourmethod consistently outperforms all state-of-the-art approaches, including competitors extensively trained on synthetic 6D pose estimation data.
Code will be publicly available at \href{https://andreacaraffa.github.io/freeze}{andreacaraffa.github.io/freeze}.
\end{abstract}

\section{Introduction}\label{sec:intro}

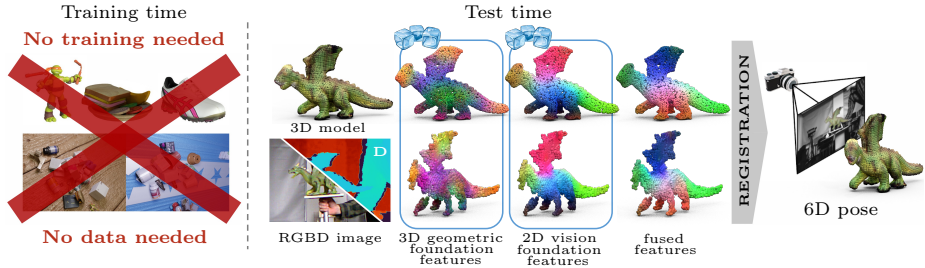
\begin{figure*}[t]
\centering
\begin{overpic}[width=1.0\textwidth,trim=0 0 0 0,clip]{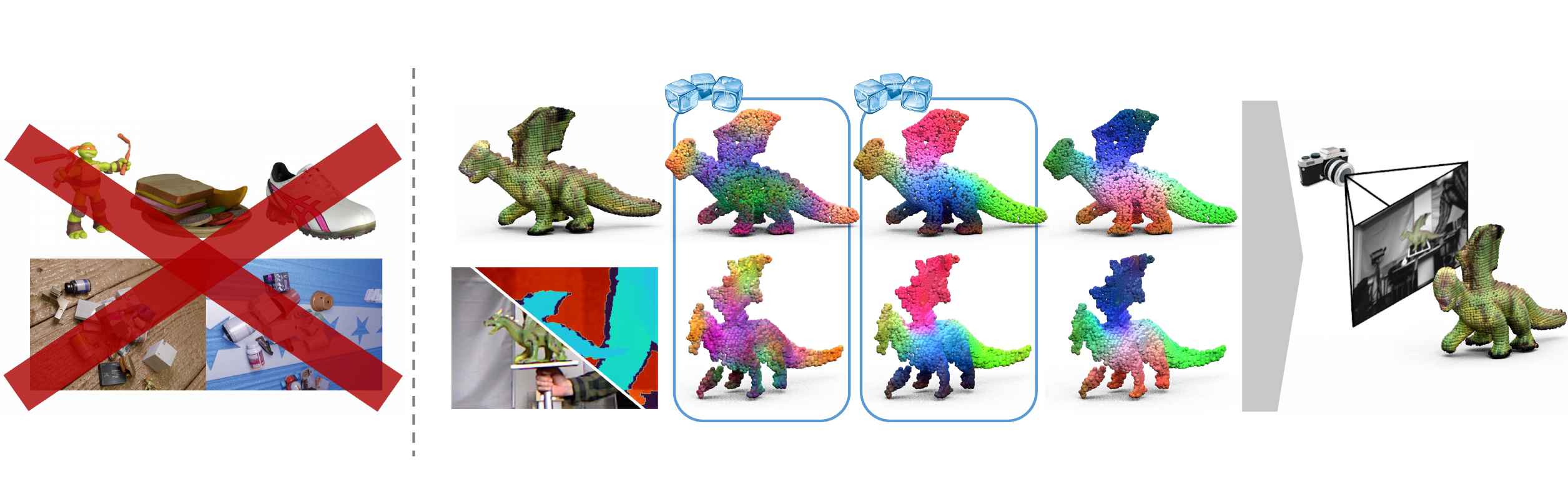}
    \put(6,27){\scriptsize {Training time}}
    \put(1.9,24){\scriptsize {\color{BrickRed}\textbf{No training needed}}}
    \put(4,2.4){\scriptsize {\color{BrickRed}\textbf{No data needed}}}
    \put(50,27){\scriptsize Test time}
    \put(31,14.5){\tiny {3D model}}
    \put(29.7,2.5){\tiny {RGBD image}}
    \put(40,11.94){\tiny \color{white} \textbf{D}}
    \put(42.8,2.5){\tiny {3D geometric}}
    \put(44,1.3){\tiny {foundation}}
    \put(45.1,0.1){\tiny {features}}
    \put(56.3,2.5){\tiny {2D vision}}
    \put(55.7,1.2){\tiny {foundation}}
    \put(56.8,0){\tiny {features}}
    \put(69.5,2.4){\tiny {fused}}
    \put(68.5,1.2){\tiny {features}}
    \put(80,6.3){\tiny \rotatebox{90}{\color{black}{\textbf{REGISTRATION}}}}
    \put(87,5.5){\scriptsize{6D pose}}
\end{overpic}

\vspace{-2mm}
\caption{
\ourmethod is designed for zero-shot object 6D pose estimation, which involves estimating the pose of an unseen 3D object within a 3D scene (\eg posing the dragon 3D model within the RGBD image displayed in the center).
Unlike most competitors, our approach eliminates the need for extensive training or the generation of large-scale tailored datasets (depicted on the left).
Instead, \ourmethod builds features suitable for 3D registration by fusing 3D geometric features with 3D-lifted 2D visual features extracted from separate pre-trained foundation models.
}
\label{fig:teaser}
\end{figure*}

In our daily interactions, we easily manipulate objects around us, whether by grasping a mug or pouring water into a glass, thanks to our subconscious ability to locate them in the real world.
In the realm of machine vision, this task is formalized as object 6D pose estimation, which involves determining the rotation and translation of an object within a scene, relative to a global reference frame. 
This becomes the key for applications such as robotic manipulation~\cite{zhuang2023instance, liu2023robotic}, augmented reality~\cite{su2019deep}, and autonomous driving~\cite{hoque2023deep}.

There exist different categories of object 6D pose estimation methods, depending on the available information about the object of interest.
The first categorization is between \emph{model-based} and \emph{model-free} methods: the former require the object's 3D model as input~\cite{zeropose, ausserlechner2023zs6d, wang2019densefusion,nguyen2023gigapose}, while the latter relax this assumption and only require a set of reference views~\cite{he2022fs6d, gao2023sa6d} or a video~\cite{he2022onepose++, sun2022onepose} of the input object.
The second distinction lies among \emph{instance-level}, \emph{category-level}, and  \emph{zero-shot} methods.
Instance-level methods~\cite{rad2018feature, peng2019pvnet, wang2019densefusion} perform pose estimation for specific object instances seen during training (\eg a particular type of drill) but do not generalize well to unseen objects.
Category-level methods~\cite{lin2021dualposenet, goodwin2022zero, li2023sd} generalize to different object instances within a category (\eg different types of drills) but do not generalize well to unseen object categories.
Zero-shot methods~\cite{labbe2022megapose,ausserlechner2023zs6d,ornek2023foundpose} address the generalization challenge by performing pose estimation for both unseen objects and categories.
This capability facilitates real-world application deployment, eliminating the need to repeatedly retrain pose estimators for every new object encountered within the application's context.

Our work focuses on the model-based zero-shot setting.
Within this context, most competitors use extensive training on large-scale pose estimation datasets, which are meticulously generated to meet specific criteria that fulfill the zero-shot paradigm.
For example, MegaPose~\cite{labbe2022megapose} utilizes physically-based rendering techniques to generate a synthetic dataset comprising two million images encompassing 20 thousand distinct objects.
However, these methods face two main limitations: firstly, their performance heavily relies on the quality and diversity of the rendered data, and secondly, their training process demands considerable time and resources.
Therefore, we aim to answer this fundamental question: ``Do we really need task-specific training at the time of foundation models?''.
To answer No! to this question, we propose a novel training-\textbf{free ze}ro-shot approach, \textit{\ourmethod}
for short, that harnesses the capabilities of pre-trained geometric and vision foundation models without requiring any training (Fig.~\ref{fig:teaser}).
As geometric foundation model we leverage GeDi~\cite{gedi}, which has been trained on 3D point cloud data of indoor scenes, while we use DINOv2~\cite{dinov2} as vision foundation model, which has been trained on web-scale 2D images with self-supervision.
\ourmethod comprises four modules: \emph{feature extraction}, \emph{feature fusion}, \emph{pose estimation}, and \emph{pose refinement}.
During feature extraction, we first compute 3D geometric features for the input object's 3D model using a frozen GeDi, and 2D visual features from the input image using a frozen DINOv2.
We feed DINOv2 with 2D images rendered from different viewpoints around the object in order to compute 2D visual features for the object's 3D model.
We feed GeDi with a 3D point cloud obtained by 3D-lifting the input depth in order to compute 3D geometric features for the scene.
During feature fusion, we compute distinctive 3D point-level features by concatenating and regularizing geometric and visual features.
The fused features are more discriminative because they encode both the geometric and visual characteristics, providing a richer and more comprehensive representation.
During pose estimation, we use RANSAC to establish valid correspondences between the 3D points of the object and the 3D-lifted scene.
In the case of geometrically-symmetric objects, we further refine their poses with a novel symmetry-aware refinement procedure based on DINOv2 visual features.
We comprehensively evaluate \ourmethod using the seven core datasets of the BOP Benchmark~\cite{BOPdatasets}, which include 
LM-O~\cite{lmo}, 
T-LESS~\cite{tless}, 
TUD-L~\cite{tudl}, 
IC-BIN~\cite{icbin}, 
ITODD~\cite{itodd}, 
HB~\cite{hb}, and 
YCB-V~\cite{ycbv}.
Together, these datasets encompass over a hundred objects and 20 thousand images, spanning a wide range of scenarios (\eg industrial vs ordinary environments) and types of noise (occlusions, variations in illumination, texture-less or symmetrical objects, cluttered scenes).
\ourmethod consistently outperforms all state-of-the-art approaches, including competitors extensively trained on large-scale synthetic pose estimation data.
In summary, our contributions are: 
\begin{itemize}[noitemsep,nolistsep,leftmargin=*]
    \item We are the first that effectively leverage the synergy between geometric and vision foundation models for the task of 6D pose estimation of unseen objects;
    \item We perform 6D pose estimation without requiring any task-specific training, resulting in a versatile solution that can be easily integrated with foundation models that may emerge in the future;
    \item We establish state-of-the-art performance on the BOP Benchmark, outperforming other competitors by a significant margin without requiring any additional training on task-specific data.
\end{itemize}
\section{Related work}\label{sec:related_work}

%%%%%%%%%%%%%%%%%%%%%%%%%%%%%%%%%%%%%%%%%%%%%%%%%%%%%%%%%%%%%%%%%%%
\noindent\textbf{3D deep descriptors} are typically used for generic point cloud registration problems~\cite{fcgf,gedi}. 
Corsetti et al.~\cite{fcgf6d} showed that these descriptors can be adapted to the challenge of object 6D pose estimation. 
Deep descriptors can be based on the Local Reference Frame (LRF), which transforms a local set of points (patch) into a canonical representation to achieve rotation and translation invariance~\cite{Yang2016}. 
Canonicalized points are then processed by a neural network to produce a compact descriptor as output.
Methods in this category include 
G3DOA~\cite{g3doa}, which utilizes multi-scale cylindrical convolutions; 
WSDesc~\cite{li2022wsdesc}, based on voxelization layers that learn optimal voxel neighborhood size; 
Li et al.~\cite{li2020end}, which employs multi-view differentiable rendering; and 
GeDi~\cite{gedi}, which processes canonicalized points through a PointNet++~\cite{qi2017pointnet++} network.
Descriptors can also be computed without relying on LRF. Methods in this category include 
SpinNet~\cite{ao2021spinnet}, which processes points projected on a cylindrical kernel using 3D cylindrical convolutions; 
FCGF~\cite{fcgf}, based on sparse convolutions for learning point-level 3D descriptors; and 
PREDATOR~\cite{predator}, which uses an attention mechanism to handle low-overlap point clouds.
These descriptors are tailored for registering point clouds of similar types, dealing with structures of points that differ significantly from those found in object 6D pose estimation benchmarks. 
In our work, we have successfully employed these types of descriptors to establish new state-of-the-art results in object 6D pose estimation.

%%%%%%%%%%%%%%%%%%%%%%%%%%%%%%%%%%%%%%%%%%%%%%%%%%%%%%%%%%%%%%%%%%%
\noindent\textbf{2D vision foundation models}, including CLIP~\cite{clip}, DINOv2~\cite{dinov2}, ImageBind~\cite{girdhar2023imagebind}, and Segment Anything Model (SAM)~\cite{sam}, can be employed across a range of tasks and domains.
CLIP~\cite{clip} connects visual concepts with textual descriptions using a ViT as image encoder and a GPT-like transformer as textual encoder, respectively.
DINOv2~\cite{dinov2} adopts self-supervised training on large-scale unlabeled data, introducing a patch-level objective to capture fine-grained image details. ImageBind~\cite{girdhar2023imagebind} integrates data from six different modalities into a shared feature space. 
This integration is achieved with image-paired data, thereby eliminating the need for combinations of all modalities.
SAM~\cite{sam} is specifically designed for object segmentation based on visual prompts such as a single point, a set of points, or a bounding box. 
SAM can effectively segment unseen object types in a variety on unseen contexts.
Although CLIP has been previously employed for zero-shot point cloud understanding~\cite{zhu2023pointclip,Peng2023}, where visual features are transferred to 3D points for downstream tasks, to the best of our knowledge, there are no existing object 6D pose estimation methods that lift 2D visual features to 3D point clouds.
In this work, we successfully achieve this by using DINOv2 visual features to boost geometric features for registration, and for effectively disambiguating object poses of geometrically symmetric objects.

%%%%%%%%%%%%%%%%%%%%%%%%%%%%%%%%%%%%%%%%%%%%%%%%%%%%%%%%%%%%%%%%%%%
\noindent\textbf{Zero-shot object 6D pose estimation} methods, while less numerous compared to fully supervised~\cite{li2019cdpn, he2021ffb6d, wang2019densefusion} or few-/one-shot learning approaches~\cite{goodwin2023you, sun2022onepose, castro2023posematcher}, are gaining attention.  
In the initial step, zero-shot image segmentation techniques like SAM~\cite{sam} can be employed to locate the object within the image.
Zephyr~\cite{okorn2021zephyr} generates candidate poses for an object and selects the best one based on point-level differences in color and geometric domains. 
MegaPose~\cite{labbe2022megapose} estimates object poses by rendering multiple views of the CAD model and matching these with the masked image to obtain a coarse pose.
ZeroPose~\cite{zeropose} uses multi-resolution geometric features to match regions between the scene's point cloud and the CAD model. 
Nguyen et al.~\cite{nguyen_2022_template} train a network to compute local features and match the image against rendered object templates.
Similarly, ZS6D~\cite{ausserlechner2023zs6d} matches DINOv2~\cite{dinov2} visual features against a database of features from rendered object templates, followed by a RANSAC-based PnP algorithm for final pose estimation. 
SAM6D~\cite{lin2023sam} generates mask proposals from images using SAM and ranks them based on a combined score of semantics, appearance, and geometry against rendered object templates.
Top proposals are then matched using 3D-3D correspondence matching.
A coarse point matching stage uses sparse correspondences to estimate the initial pose, followed by a fine point matching stage that refines it using Sparse-to-Dense Point Transformers.
GigaPose~\cite{nguyen2023gigapose} renders templates to extract dense features using ViT and then finds the best template using fast nearest neighbour search in the feature space. On top, two lightweight MLPs estimate 2D scale and in-plane rotation from a single 2D-2D correspondence using local features.
FoundPose~\cite{ornek2023foundpose} first leverages DINOv2 and bag-of-words descriptors to shortlist similar templates. 2D-3D correspondences are then established among query and the selected templates using DINOv2 patch level features.
FoundationPose~\cite{wen2023foundationpose} offers a unified approach for pose tracking and estimation in both model-based and model-free scenarios.
It trains on a large synthetic dataset generated via Large Language Models (LLMs).
The process starts with coarse pose estimation, refined by a transformer based architecture.
The refined pose hypotheses are then evaluated by a pose selection network using a hierarchical comparison and a pose ranking encoder trained on pose-conditioned triplet loss.

Unlike existing zero-shot 6D pose estimation approaches, \ourmethod employs deep neural networks that have been trained for various and different generic downstream tasks.
\ourmethod can utilize point cloud descriptors that were trained for point cloud registration, such as GeDi~\cite{gedi}, and image feature extraction networks that were self-supervised on large-scale internet data, such as DINOv2~\cite{dinov2}. 
\ourmethod is designed to be adaptable, allowing for the incorporation of new state-of-the-art feature representation methods as they become available in the future, without requiring re-training on specific data, all while maintaining the core methodology.

FoundationPose leverages LLMs to generate training synthetic data at scale. We make a step forward, and by leveraging recent geometric and vision foundation models, we do not require any training data at all. ZS6D and FoundPose are training-free competitors. However, unlike them that use visual features to find 2D-2D and 2D-3D correspondences, respectively, we exploit the depth information of the scene and we work with 3D-3D correspondences. Moreover, we leverage also geometric and not only vision foundation models.

%%%%%%%%%%%%%%%%%%%%%%%%%%%%%%%%%%%%%%%%%%%%%%%%%%%%%%%%%%%%%%%%%%%%%%%
%%%%%%%%%%%%%%%%%%%%%%%%%%%%%%%%%%%%%%%%%%%%%%%%%%%%%%%%%%%%%%%%%%%%%%%
%%%%%%%%%%%%%%%%%%%%%%%%%%%%%%%%%%%%%%%%%%%%%%%%%%%%%%%%%%%%%%%%%%%%%%%
\section{Our approach}\label{sec:method}

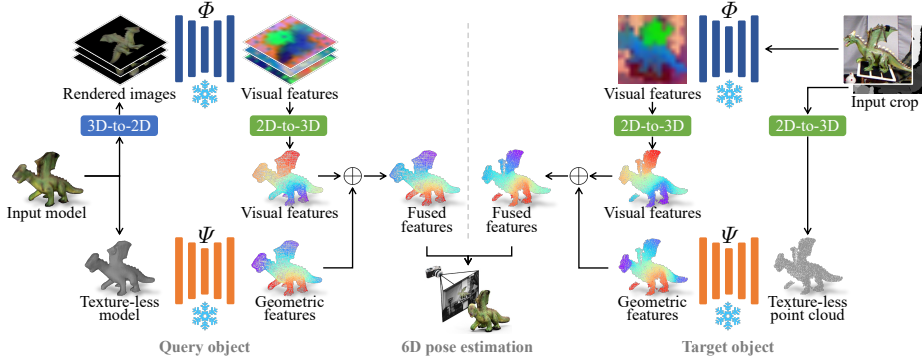
\begin{figure*}[t]
\centering
\begin{overpic}[width=\textwidth]{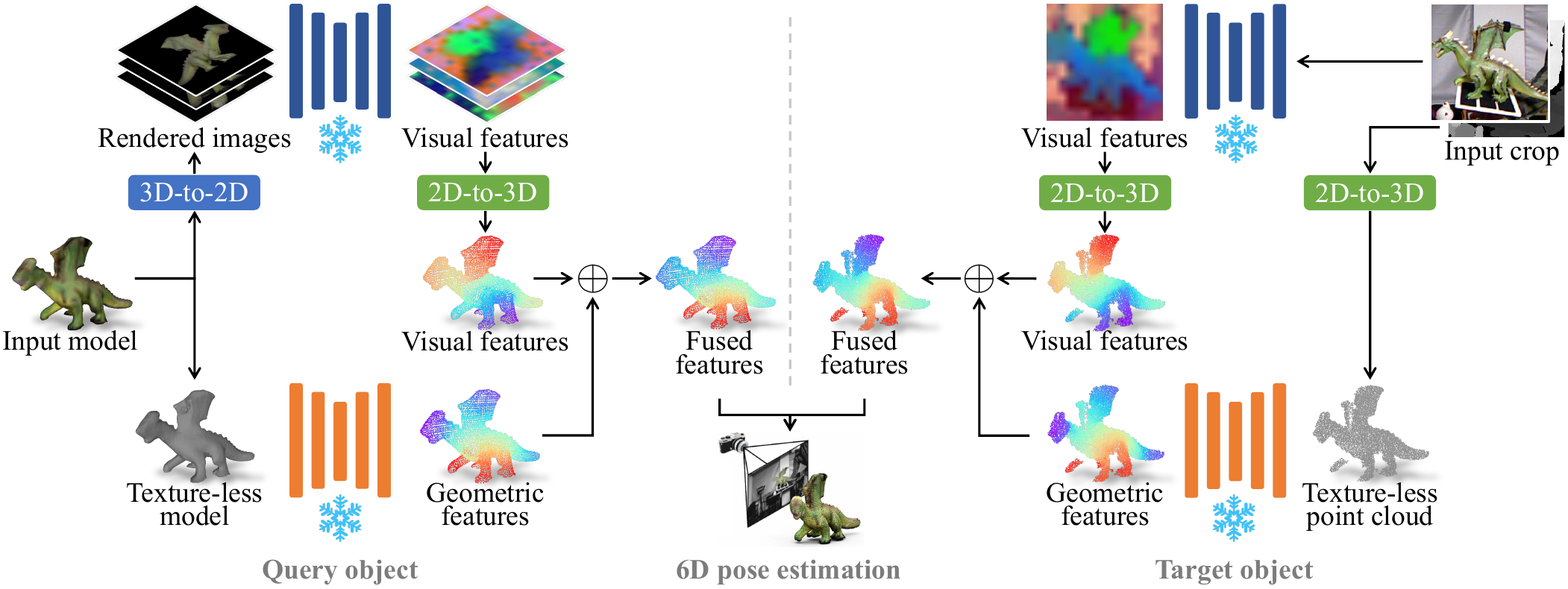}
    \put(21,37){{$\Phi$}}
    \put(78,37){{$\Phi$}}
    
    \put(21,12.5){{$\Psi$}}
    \put(78,12.5){{$\Psi$}}
\end{overpic}

\vspace{-2mm}
\caption{
Overview of \ourmethod. 
Left-hand side, we create rendered images from the input model of the query object to extract visual features, which we then back project to the object point cloud. 
Concurrently, we extract geometric features, which we then fuse with the visual features.
Similarly, right-hand side, we compute visual and geometric features of the target object imaged in the input crop, and fuse the two as before.
Although the query and target objects are from two different modalities (a 3D model the former, and a RGBD image the latter), we employ the same vision and geometric encoders to compute their features.
Lastly, we input the fused features to a registration algorithm based on feature matching to estimate the object 6D pose.
The color heatmaps used for the features are generated by reducing feature dimensionality with t-SNE~\cite{tsne}. 
The symbol $\oplus$ indicates feature concatenation.
}
\label{fig:diagram}
\end{figure*}

%%%%%%%%%%%%%%%%%%%%%%%%%%%%%%%%%%%%%%%%%%%%%%%%%%%%%%%%%%%%%%%%%%%%%%%
%%%%%%%%%%%%%%%%%%%%%%%%%%%%%%%%%%%%%%%%%%%%%%%%%%%%%%%%%%%%%%%%%%%%%%%
\subsection{Overview}\label{sec:overview}

Given the 3D model of a \textit{query object} $Q$ and an RGBD image $\textbf{I} \in \mathbb{R}^{H \times W \times 4}$ capturing $Q$, our goal is to find the 6D pose of $Q$ with respect to the camera's reference frame. 
Let the \textit{target object} $T$ be the instance of $Q$ captured in $\textbf{I}$, our goal is then to estimate the 6DoF transformation $\textbf{T} = (\textbf{R}, \textbf{t})$ that relates $Q$ to $T$, where $\textbf{R} \in \mathbb{R}^{3 \times 3}$ and   $\textbf{t} \in \mathbb{R}^3$ are the rotation and translation components.
We address the problem by leveraging geometric and vision foundation models to compute features for $Q$ and $T$, and by then estimating $\textbf{R}$ and $\textbf{t}$ from 3D-3D correspondences (Fig.~\ref{fig:diagram}). 
We sample $N$ points from the 3D model's surface to generate a point cloud $\mathcal{P}^Q \in \mathbb{R}^{N \times 3}$ from which we compute geometric features for each point through a frozen pre-trained geometric encoder $\Psi$. 
Concurrently, we use the 3D model to render templates from different viewpoints, 
which we feed to a frozen pre-trained vision encoder $\Phi$ to compute pixel-level visual features that we subsequently back-project onto $\mathcal{P}^Q$ (Sec.~\ref{sec:query_object}).
For the RGBD image, we locate $T$ in $\textbf{I}$ using segmentation masks estimated by a zero-shot object detectors.
We generate a point cloud from the pixels belonging to the mask by using the depth channel and the camera's intrinsic parameters.
We sample $M$ points to get $\mathcal{P}^T \in \mathbb{R}^{M \times 3}$ and we compute geometric features for these points. 
Concurrently, we compute pixel-level visual features from the RGB channels and back-project them onto $\mathcal{P}^T$ (Sec.~\ref{sec:target_object}). 
The geometric and visual features, for both $Q$ and $T$, are fused together (Sec.~\ref{sec:feat_fusion}) and used to perform 3D-3D matching and registration (Sec.~\ref{sec:pose_estimation}). Finally, we refine the pose of objects using our novel symmetry aware refinement algorithm (Sec.~\ref{sec:pose_refinement}).

%%%%%%%%%%%%%%%%%%%%%%%%%%%%%%%%%%%%%%%%%%%%%%%%%%%%%%%%%%%%%%%%%%%%%%%
%%%%%%%%%%%%%%%%%%%%%%%%%%%%%%%%%%%%%%%%%%%%%%%%%%%%%%%%%%%%%%%%%%%%%%%
\subsection{Query object processing}\label{sec:query_object}

We compute the geometric features of $Q$ by processing $\mathcal{P}^Q$ directly with the geometric encoder $\Psi$. 
Let $\textbf{g}^Q_n \in \mathbb{R}^{G}$ be the feature of the $n$-th point of $Q$, where $G$ is the feature size, then $\mathcal{G}^Q = \{ \textbf{g}^Q_n \}_{n=1}^{N} = \Psi(\mathcal{P}^Q)$ represents $Q$'s geometric features.
We compute the visual features of $Q$ by rendering RGBD images of the textured 3D model of $Q$ from $R$ different viewpoints, to obtain the set $\{ \textbf{I}_r \}_{r=1}^{R}$.
We also render the associated object segmentation masks, which we use to crop the portion $\textbf{I}^Q_r$ of $\textbf{I}_r$ occupied by $Q$.
We process the RGB channels of $\textbf{I}^Q_r$ with the vision encoder $\Phi$ to generate pixel-level\footnote{
Since we use a Transformer for $\Phi$, we use bilinear interpolation to convert patch-level features to pixel-level ones. For example, DINOv2~\cite{dinov2} outputs a 16$\times$16 grid of patch-level features.}
visual features $\textbf{V}^Q_r$.
$\textbf{V}^Q_r$ are then back-projected to the 3D model, and associated with $\mathcal{P}^Q$'s points as follows.
First, we compute the correspondences between the RGB pixels of $\textbf{I}^Q_r$ and the points of $\mathcal{P}^Q$.
We convert the depth channel of $\textbf{I}^Q_r$ into a viewpoint-dependent point cloud $\mathcal{P}^Q_r$ using the renderer's camera intrinsic parameters, and employ nearest neighbour search between the points of $\mathcal{P}^Q_r$ and $\mathcal{P}^Q$.
Then, we leverage this correspondence map to associate the visual features $\textbf{V}^Q_r$ extracted from the $r$-th rendered image $\textbf{I}^Q_r$ to the points of $\mathcal{P}^Q$.
Lastly, we aggregate multi-view features $\{ \textbf{V}^Q_r \}_{r=1}^R$ into a single set $\mathcal{V}^Q = \{ \textbf{v}^Q_n\}_{n=1}^{N}$ by averaging the contribution of each viewpoint as $\textbf{v}^Q_n = \sum_{r=1}^R \textbf{v}^Q_{r,n} / R$.

%%%%%%%%%%%%%%%%%%%%%%%%%%%%%%%%%%%%%%%%%%%%%%%%%%%%%%%%%%%%%%%%%%%%%%%
%%%%%%%%%%%%%%%%%%%%%%%%%%%%%%%%%%%%%%%%%%%%%%%%%%%%%%%%%%%%%%%%%%%%%%%
\subsection{Target object processing}\label{sec:target_object}

%%%%%%%%%%%%%%%%%%%%%%%%%%%%%%%%%%%%%%%%%%%%%%%%%%%%%%%%%%%%%%%%%%%%%%%
\noindent\textbf{Localisation.}~We use a zero-shot object segmentation algorithm, such as CNOS~\cite{cnos}, to localise $T$ within \textbf{I}.
CNOS computes region proposals generated using SAM~\cite{sam} (or FastSAM~\cite{fastsam}) within $\textbf{I}$, and compares them against object templates derived from the textured 3D model of query object.
It assigns a score based on the visual feature similarity with the object templates to each proposal.
CNOS produces a set of masks as output, each one with a respective confidence score.
The common practice is to select the mask with the highest confidence score. 
However, we experimentally observed that CNOS is not well calibrated, as more accurate segmentation masks might have lower confidence scores.
Hence, we keep the most confident masks as possible target object candidates. 
For each candidate, besides the mask we also extract the minimum bounding box $\textbf{I}^{T} \in \textbf{I}$ that contains the mask.

%%%%%%%%%%%%%%%%%%%%%%%%%%%%%%%%%%%%%%%%%%%%%%%%%%%%%%%%%%%%%%%%%%%%%%%
\noindent\textbf{Feature extraction.}~We compute the geometric features of $T$ by processing $\mathcal{P}^T$ with $\Psi$. 
Let $\textbf{g}^T_m \in \mathbb{R}^{G}$ be the feature of the $m$-th point of $T$, and $\mathcal{G}^T = \{ \textbf{g}^T_m \}_{m=1}^{M} = \Psi(\mathcal{P}^T)$ be the set of features of $T$, where $M$ is the number of points.s
We compute pixel-level features $\textbf{V}^{T}$ by processing the input crop $\textbf{I}^T$ with $\Phi$.
However, associating $\textbf{V}^T$ with the points of $\mathcal{P}^{T}$ is more straightforward than the process required for the query object.
Leveraging the one-to-one correspondences between the pixels of the RGB crop and the depth, we can transfer pixel-level features $\textbf{V}^T$ to the points of $\mathcal{P}^{T}$ forming the set $\mathcal{V}^T = \{ \textbf{v}_{m}^T\}_{m=1}^{M}$.
Despite $\Phi$ processes both foreground and background pixels of $\textbf{I}^T$, we only transfer foreground features to $\mathcal{P}^T$.

%%%%%%%%%%%%%%%%%%%%%%%%%%%%%%%%%%%%%%%%%%%%%%%%%%%%%%%%%%%%%%%%%%%%%%%
%%%%%%%%%%%%%%%%%%%%%%%%%%%%%%%%%%%%%%%%%%%%%%%%%%%%%%%%%%%%%%%%%%%%%%%
\subsection{Visual and geometric feature fusion}\label{sec:feat_fusion}

We fuse visual and geometric features through concatenation for both $Q$ and $T$.
We apply $L_2$ normalization to $\mathcal{V}^Q$ $\left(\mathcal{V}^{T} \right)$ and $\mathcal{G}^Q$ $\left(\mathcal{G}^T\right)$ independently to account for potential differences in norms and to balance their contribution in the final feature.
Specifically, the features we produce have the following form: 
%++++++++++++++++++++++++++++++++++
\begin{align}
\small
& \mathcal{F}^Q = \{ \textbf{f}^{Q}_{n} \}_{n=1}^{N} = \left\{ \left[ \frac{ \textbf{v}^{Q}_{n}}{\lVert \textbf{v}^{Q}_{n} \rVert_2} \bigg| \frac{\textbf{g}^{Q}_{n}}{\lVert \textbf{g}^{Q}_{n} \rVert_2} \right] \right\}_{n=1}^{N} \nonumber ,\\
& \mathcal{F}^T = \{ \textbf{f}^{T}_{m} \}_{m=1}^{M} = \left\{ \left[ \frac{ \textbf{v}^{T}_{m}}{\lVert \textbf{v}^{T}_{m} \rVert_2} \bigg| \frac{\textbf{g}^{T}_{m}}{\lVert \textbf{g}^{T}_{m} \rVert_2} \right] \right\}_{m=1}^{M} \nonumber ,
\end{align}
%++++++++++++++++++++++++++++++++++
where $\textbf{f}^{Q}_{n}, \textbf{f}^{T}_{m} \in \mathbb{R}^{V+G}$,$[\cdot|\cdot]$ is the concatenation and $\lVert\cdot\rVert_2$ is the $L_2$ norm operator.
We found this simple fusion strategy working well in practice.
However, alternative fusion strategies can be employed, and we leave this for future developments.

%%%%%%%%%%%%%%%%%%%%%%%%%%%%%%%%%%%%%%%%%%%%%%%%%%%%%%%%%%%%%%%%%%%%%%%
%%%%%%%%%%%%%%%%%%%%%%%%%%%%%%%%%%%%%%%%%%%%%%%%%%%%%%%%%%%%%%%%%%%%%%%
\subsection{Pose estimation}\label{sec:pose_estimation}
We employ an off-the-shelf registration algorithm based on 3D-3D feature matching, like RANSAC~\cite{choi2015robust}, to robustly estimate the coarse transformation $\textbf{T}_c$ from the pair $((\mathcal{P}^Q, \mathcal{F}^Q), (\mathcal{P}^T,\mathcal{F}^T))$.
RANSAC operates by sampling triplets of points from $\mathcal{P}^Q$, and searching for their corresponding points in $\mathcal{P}^T$ by performing a nearest neighbour search within the fused feature space.
False matches are pruned, while true matches are utilised to compute the transformation $(\textbf{R}, \textbf{t})$, thereby registering $\mathcal{P}^Q$ to $\mathcal{P}^T$, \ie $\mathcal{P}^Q \textbf{R} + \textbf{t} \approx \mathcal{P}^T$. 
The pair $(\textbf{R}, \textbf{t})$ represents the predicted 6D pose of $Q$.
We estimate the pose for each possible candidate $T$, and retain only the one with associated the highest number of registration inliers.

%%%%%%%%%%%%%%%%%%%%%%%%%%%%%%%%%%%%%%%%%%%%%%%%%%%%%%%%%%%%%%%%%%%%%%%
%%%%%%%%%%%%%%%%%%%%%%%%%%%%%%%%%%%%%%%%%%%%%%%%%%%%%%%%%%%%%%%%%%%%%%%
\subsection{Symmetry-aware refinement}\label{sec:pose_refinement}
We use the Iterative Closest Point (ICP) algorithm~\cite{chen1992object} to refine $\textbf{T}_\textrm{c}$ at point level and obtain a finer transformation $\textbf{T}_\textrm{f}$.
Although ICP is known to be sensitive to local minima, we find it effective to refine several of our initial poses that are close to the correct solution. Then, we use visual features computed by $\Phi$ to solve pose ambiguities that arise from geometric symmetries.
Although our fused features incorporate visual information, query objects with geometric symmetries can still lead to local-minimum registration solutions. 
We mitigate this problem by introducing a novel symmetry-aware refinement based on rendering and visual features matching. Our aim is to adjust the pose for those objects for which the estimate pose would be correct without considering the texture, \ie when $\mathcal{P}^Q$ and $\mathcal{P}^T$ match, but the texture does not.

%%%%%%%%%%%%%%%%%%%%%%%%%%%%%%%%%%%%%%%%%%%%%%%%%%%%%%%%%%%%%%%%%%%%%%%
\noindent\textbf{Symmetry estimation} aims to identify geometric symmetries of $Q$'s 3D model, regardless of its texture. 
We build a set of rotation matrices by uniformly sampling the rotation space. 
Our estimated symmetries $\{ \textbf{R}_s \in SO(3) \}^S$ are then the rotations that minimize the Chamfer distance~\cite{Zhao2019} between $\textbf{R}_s \mathcal{P}^Q$ and $\mathcal{P}^{Q}$.

\begin{figure}[t]
\centering
    \begin{tabular}{@{}c@{\,}c@{\,}c@{\,}c@{\,}c@{\,}c}
    \raggedright
        \begin{overpic}[width=0.19\columnwidth]{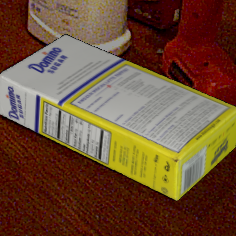}
            \put(3, 5){\color{white}\scriptsize \textbf{Initial pose}}
        \end{overpic} &
        \begin{overpic}[width=0.19\columnwidth]
        {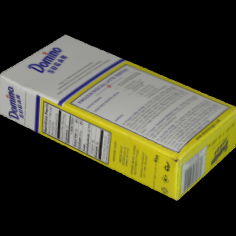}
            \put(3, 5){\color{white}\scriptsize \textbf{Sym. 0 (Id)}}
        \end{overpic} &
        \begin{overpic}[width=0.19\columnwidth]{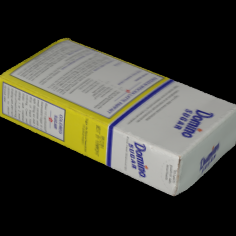}
            \put(3, 5){\color{white}\scriptsize \textbf{Sym. 1 }}
        \end{overpic} &
        \begin{overpic}[width=0.19\columnwidth]{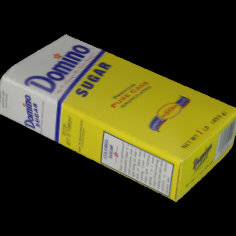}
            \put(3, 5){\color{white}\scriptsize \textbf{Sym. 2}}
        \end{overpic} &
        \begin{overpic}[width=0.19\columnwidth]{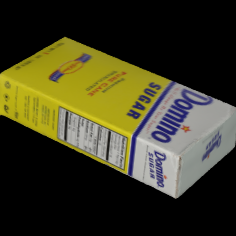}
            \put(3, 5){\color{white}\scriptsize \textbf{Sym. 3}}
        \end{overpic} & \\
        \begin{overpic}[width=0.19\columnwidth]{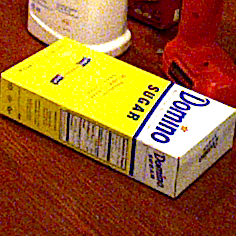}
            \put(3, 5){\color{white}\scriptsize \textbf{Input crop}}
            \put(30, -15){\footnotesize Score}
        \end{overpic} &  
        \begin{overpic}[width=0.19\columnwidth]{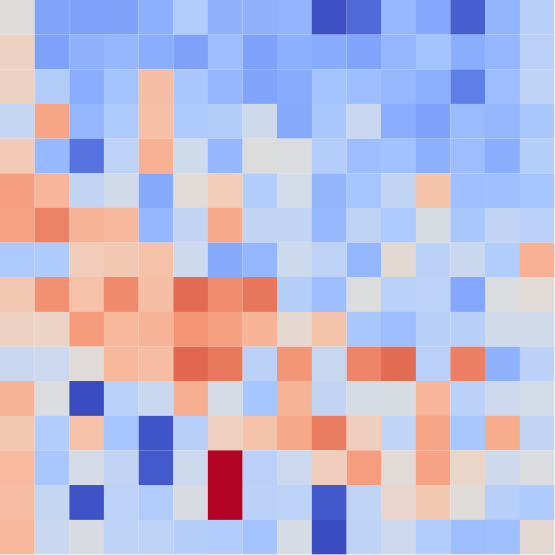}
            \put(35, -15){\footnotesize 0.36}
        \end{overpic} &
        \begin{overpic}[width=0.19\columnwidth]{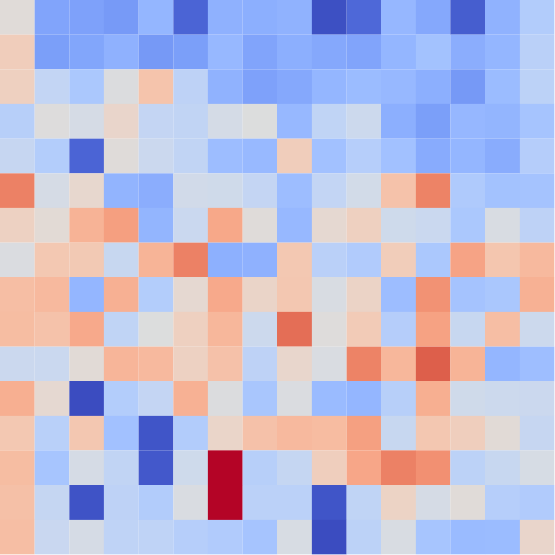}
            \put(35, -15){\footnotesize 0.36}
        \end{overpic} &
        \begin{overpic}[width=0.19\columnwidth]{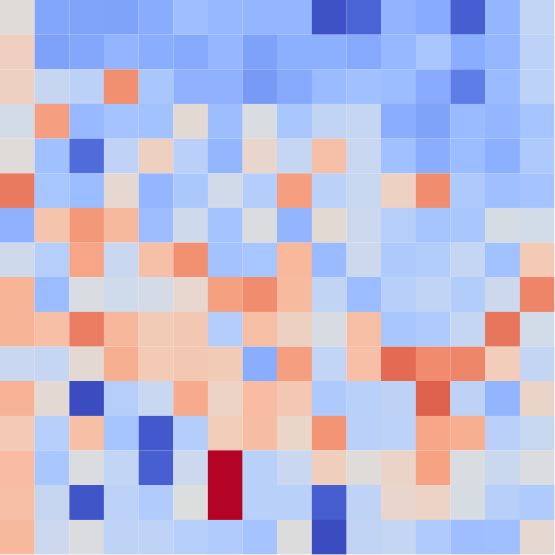}
            \put(35, -15){\footnotesize 0.35}
        \end{overpic} &
        \begin{overpic}[width=0.19\columnwidth]{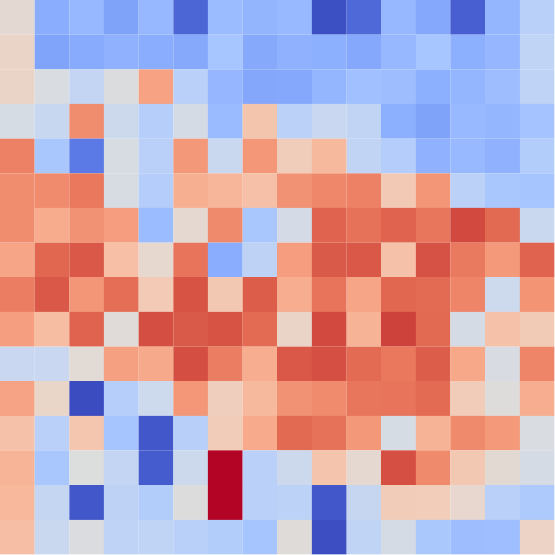}
            \put(30, -15){\footnotesize 0.47 \color{ForestGreen} \cmark}
        \end{overpic} &
        \begin{overpic}[width=0.01035\columnwidth]{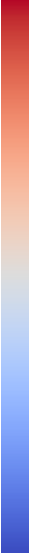}
            \put(7, -1){\tiny $-0.1$}
            \put(7, 93){\tiny $+1$}
        \end{overpic}
    \end{tabular}
    \caption{
    Example of symmetry-aware refinement.
    We adjust the inaccurate initial pose (top left corner) by comparing the input crop (bottom left corner) with the rendered images of the four object symmetries (first row). Sym. 0 (Id) is the identity transformation, \ie without applying any symmetric transformation.
    By computing the patch-level cosine similarities for each candidate (second row, red-white-blue colors), and averaging them to obtain a global score (third row), we can select the best candidate as the one with the highest score ({\color{ForestGreen} \cmark}).
    }
    \label{tab:symmetries}
\end{figure}

%%%%%%%%%%%%%%%%%%%%%%%%%%%%%%%%%%%%%%%%%%%%%%%%%%%%%%%%%%%%%%%%%%%%%%%
\noindent\textbf{Symmetry selection.}~
We render the RGB images $\{\textbf{I}_s\}^S$ depicting the 3D model of $Q$ in the poses  $\{\textbf{T}_f \circ \textbf{R}_s\}^S$. 
We use the bounding box defined in Sec.~\ref{sec:target_object} to crop images $\{\textbf{I}_s^T\}^S$.
We then process $\{\textbf{I}_s^T\}^S$ with $\Phi$ to produce visual features $\{\textbf{V}_s\}^S$.
The best candidate is selected by computing the cosine similarity between $\{\textbf{V}_s\}^S$ and the intermediate patch-level visual features $\textbf{V}^{T}$ that we computed in Sec.~\ref{sec:target_object}.
With $\Phi$ being a Transformer, we observed that patch-level features are more informative than the class token to encode pose information, a conclusion also supported in \cite{cnos}. Moreover, as both visual features coming for $I^T$ and $I_s$ are at patch-level, we use them as they are. Instead, in Sec.~\ref{sec:target_object}, we needed to map features to points, hence we augmented patch-level features to pixel-level features.
Fig.~\ref{tab:symmetries} shows an example of symmetry-aware refinement. 
Starting from an inaccurately estimated pose, we rotate and render the object according to its four estimated symmetries.
We compute the patch-level cosine similarity between the visual feature of the input crop and the rendered images. 
Symmetry no.~3 exhibits a higher similarity score ($+30\%$), enabling us to correct the initial pose.

%%%%%%%%%%%%%%%%%%%%%%%%%%%%%%%%%%%%%%%%%%%%%%%%%%%%%%%%%%%%%%%%%%%%%%
%%%%%%%%%%%%%%%%%%%%%%%%%%%%%%%%%%%%%%%%%%%%%%%%%%%%%%%%%%%%%%%%%%%%%%
%%%%%%%%%%%%%%%%%%%%%%%%%%%%%%%%%%%%%%%%%%%%%%%%%%%%%%%%%%%%%%%%%%%%%%
\section{Experiments}

%%%%%%%%%%%%%%%%%%%%%%%%%%%%%%%%%%%%%%%%%%%%%%%%%%%%%%%%%%%%%%%%%%%%%%
%%%%%%%%%%%%%%%%%%%%%%%%%%%%%%%%%%%%%%%%%%%%%%%%%%%%%%%%%%%%%%%%%%%%%%
\subsection{Experimental setup}
We randomly sample $N=5K$ points for the query object, and $M=1K$ for the target object.
We use the masks estimated with CNOS~\cite{cnos} and SAM6D~\cite{lin2023sam} to localize the object within the input image.
As geometric encoder we use GeDi~\cite{gedi}, a geometric feature extractor for point clouds with strong generalization capability across different applications domains.
To capture geometric details at different scales, we extract 32-dimensional GeDi features from local neighbours occupying 30\% and 40\% of the object's diameter, and concatenate them to obtain the final 64-dimensional geometric features.
As vision encoder we use the ViT-giant version of DINOv2~\cite{dinov2}.
DINOv2 is a vision foundation model trained through self-supervision on large-scale web data.
DINOv2 processes images of size 224$\times$224 and outputs 1536-dimensional features.
We apply PCA to reduce their dimensionality to $V = 64$, for efficiency and to match geometric features' dimensionality.

%%%%%%%%%%%%%%%%%%%%%%%%%%%%%%%%%%%%%%%%%%%%%%%%%%%%%%%%%%%%%%%%%%%%%%
%%%%%%%%%%%%%%%%%%%%%%%%%%%%%%%%%%%%%%%%%%%%%%%%%%%%%%%%%%%%%%%%%%%%%%
\subsection{Datasets}

We evaluate \ourmethod on the seven core datasets of the BOP Benchmark~\cite{sundermeyer2023bop}: 
LM-O~\cite{lmo}, 
T-LESS~\cite{tless}, 
TUD-L~\cite{tudl}, 
IC-BIN~\cite{icbin}, 
ITODD~\cite{itodd}, 
HB~\cite{hb}, and 
YCB-V~\cite{ycbv}.
Each dataset provides both 3D models of the objects and test RGBD images.
%%%%%%%%%%%%%%%%%%%%%%%%%%%%%%%%%%%%%%%%%%%%%%%%
\noindent\textbf{Object types.}~T-LESS and ITODD contain industrial objects, such as electrical and mechanical items, with planar sides, sharp edges, and hollow parts.
The other datasets contain ordinary objects, such as food packages, toy models, and stationery items, with finer geometric details and smooth boundaries.
Industrial objects are provided as CAD models,
while the 3D models of ordinary objects are reconstructed from images acquired with RGBD sensors.
%%%%%%%%%%%%%%%%%%%%%%%%%%%%%%%%%%%%%%%%%%%%%%%%
\noindent\textbf{Number of instances.}~LM-O, TUD-L, HB, and YCB-V contain at most one instance per object, while the other datasets contain multiple instances. ITODD contains 4 instances per object on average, except a few images that contain 84 instances per object. 
T-LESS and IC-BIN contains 1.3 and 9 instances per object on average, respectively.
%%%%%%%%%%%%%%%%%%%%%%%%%%%%%%%%%%%%%%%%%%%%%%%%
\noindent\textbf{Photometric information.}~T-LESS and ITODD objects are texture-less.
LM-O, TUD-L, and HB objects have mostly uniform colors.
IC-BIN and YCB-V objects have rich textures.
%%%%%%%%%%%%%%%%%%%%%%%%%%%%%%%%%%%%%%%%%%%%%%%%
\noindent \textbf{Noise levels.}~TUD-L contains a single object for each scene with difficult light conditions.
LM-O scenes are highly cluttered (other objects are present) and feature occlusions.
IC-BIN scenes contain several instances of the same object, thus featuring different levels of occlusions and a high localization ambiguity.
The other datasets contain mildly occluded objects.

%%%%%%%%%%%%%%%%%%%%%%%%%%%%%%%%%%%%%%%%%%%%%%%%%%%%%%%%%%%%%%%%%%%%%%
%%%%%%%%%%%%%%%%%%%%%%%%%%%%%%%%%%%%%%%%%%%%%%%%%%%%%%%%%%%%%%%%%%%%%%
\subsection{Metrics}\label{sec:metrics}

We evaluate the accuracy of pose estimation using the default metrics of the BOP Benchmark~\cite{sundermeyer2023bop}, which are VSD, MSSD, and MSPD.
These metrics measure different errors between two 3D models obtained by transforming the object 3D model with the predicted and ground-truth poses.
VSD measures the discrepancy between the depth maps obtained by rendering them.
MSSD and MSPD measure the maximum Euclidean distance and the maximum reprojection error between their points, respectively.
Both MSSD and MSPD take into consideration symmetries, by assuming the minimum value reached across all symmetrically equivalent ground-truth poses.
For each dataset, we report the average of the three metrics, namely Average Recall (AR).

%%%%%%%%%%%%%%%%%%%%%%%%%%%%%%%%%%%%%%%%%%%%%%%%
%%%%%%%%%%%%%%%%%%%%%%%%%%%%%%%%%%%%%%%%%%%%%%%%
\subsection{Quantitative results}

\renewcommand{\arraystretch}{0.9}
\begin{table*}[t]
    \centering
    \caption{
    Results on the BOP Benchmark datasets.
    We report the AR score on each of the seven core datasets of the BOP Benchmark and the mean AR across datasets. For a fair comparison, we report results with zero-shot localization prior, while we do not report any result with supervised or undefined prior.
    \textbf{Bold} font indicates best AR.
    Our results are \colorbox{cosmiclatte}{highlighted}.
    Keys. 
    Training free: task-specific training free;
    Prior: type of instance localization prior;
    Refin.: pose refinement; 
    Mean: Mean AR;
    `-': not available.
    }
    \label{tab:main_results}
    \vspace{-3mm}
    \resizebox{\textwidth}{!}{%
    \begin{tabular}{clcccc|ccccccc|c}
        \toprule
        & \multirow{2}{*}{Method}  & Training & \multirow{2}{*}{Input} & \multirow{2}{*}{Prior} & \multirow{2}{*}{Refin.} & \multicolumn{7}{c|}{BOP Dataset} & \multirow{2}{*}{Mean} \\
        & & free & & & & LM-O & T-LESS & TUD-L & IC-BIN & ITODD & HB & YCB-V & \\
        \toprule
        {\color{gray} \scriptsize 1} & MegaPose~\cite{labbe2022megapose} & & RGB & \multirow{6}{*}{CNOS} & & 22.9 & 17.7 & 25.8 & 15.2 & 10.8 & 25.1 & 28.1 & 20.8\\
        {\color{gray} \scriptsize 2} & ZS6D~\cite{ausserlechner2023zs6d} & \color{ForestGreen}{\cmark} & RGB &  & & 29.8 & 21.0 & - & - & - & - & 32.4 & - \\
        {\color{gray} \scriptsize 3} & GigaPose~\cite{nguyen2023gigapose} & & RGB & & & 29.9 & 27.3 & 30.2 & 23.1 & 18.8 & 34.8 & 29.0 & 27.6 \\
        {\color{gray} \scriptsize 4} & FoundPose~\cite{ornek2023foundpose} & \color{ForestGreen}{\cmark} & RGB & & & 39.7 & 33.8 & 46.9 & 23.9 & 20.4 & 50.8 & 45.2 & 37.3 \\
        {\color{gray} \scriptsize 5} & SAM6D~\cite{lin2023sam} & & RGBD & & & 57.0 & 38.2 & 69.8 & 41.5 & 41.4 & 66.9 & 73.2 & 55.4 \\
        \rowcolor{cosmiclatte}
        {\color{gray} \scriptsize 6} & \ourmethod (ours) & \color{ForestGreen}{\cmark} & RGBD & & & \textbf{64.7} & \textbf{49.3} & \textbf{86.1} & \textbf{44.3} & \textbf{49.2} & \textbf{75.7} & \textbf{78.7} & \textbf{64.0}  \\
        \cmidrule{2-14}
        
        {\color{gray} \scriptsize 7} & MegaPose~\cite{labbe2022megapose} & & RGB & \multirow{7}{*}{CNOS} & \color{ForestGreen}{\cmark}  & 56.0 & 50.8 & 68.7 & 41.9 & 34.6 & 70.6 & 62.0 & 54.9 \\
        {\color{gray} \scriptsize 8} & GigaPose~\cite{nguyen2023gigapose} & & RGB & & \color{ForestGreen}{\cmark} & 59.9 & \textbf{57.0} & 63.5 & 46.7 & 39.7 & 72.2 & 66.3 & 57.9 \\
        {\color{gray} \scriptsize 9} & FoundPose~\cite{ornek2023foundpose} & \color{ForestGreen}{\cmark} & RGB & & \color{ForestGreen}{\cmark} & 61.0 & \textbf{57.0} & 69.3 & 47.9 & 40.7 & 72.3 & 69.0 & 59.6 \\
        {\color{gray} \scriptsize 10} & ZeroPose~\cite{zeropose} & & RGBD & & \color{ForestGreen}{\cmark} & 53.8 & 40.0 & 83.5 & 39.2 & 52.1 & 65.3 & 65.3 & 57.0 \\
        {\color{gray} \scriptsize 11} & MegaPose~\cite{labbe2022megapose} & & RGBD & & \color{ForestGreen}{\cmark} & 62.6 & 48.7 & 85.1 & 46.7 & 46.8 & 73.0 & 76.4 & 62.8 \\
        {\color{gray} \scriptsize 12} & SAM6D~\cite{lin2023sam} & & RGBD & & \color{ForestGreen}{\cmark} & 63.5 & 46.3 & 80.0 & 46.5 & 54.3 & 71.1 & 80.0 & 63.2 \\
        \rowcolor{cosmiclatte} {\color{gray} \scriptsize 13} & \ourmethod (ours) & \color{ForestGreen}{\cmark} & RGBD & & \color{ForestGreen}{\cmark} & \textbf{69.0} & 52.0 & \textbf{93.6} & \textbf{49.9} & \textbf{56.1} & \textbf{79.0} & \textbf{85.3} & \textbf{69.3} \\
        \cmidrule{1-14}

        {\color{gray} \scriptsize 14} & SAM6D~\cite{lin2023sam} & & & & & 62.7 & 42.0 & 77.7 & \textbf{50.4} & 45.5 & 68.9 & 74.3 & 60.2 \\
        \rowcolor{cosmiclatte} {\color{gray} \scriptsize 15} & \ourmethod (ours) & \color{ForestGreen}{\cmark}  & \multirow{-2}{*}{RGBD} & \multirow{-2}{*}{SAM6D} & & \textbf{67.6} & \textbf{50.0} & \textbf{88.1} & 48.7 & \textbf{52.0} & \textbf{76.1} & \textbf{77.4} & \textbf{65.7}  \\
        \cmidrule{2-14}

        {\color{gray} \scriptsize 16} & SAM6D~\cite{lin2023sam} & & & & \color{ForestGreen}{\cmark} & 68.7 & 49.8 & 87.4 & \textbf{56.1} & 57.7 & 75.4 & 82.8 & 68.2 \\
        \rowcolor{cosmiclatte} {\color{gray} \scriptsize 17} & \ourmethod (ours) &\color{ForestGreen}{\cmark}  & \multirow{-2}{*}{RGBD} & \multirow{-2}{*}{SAM6D} & \color{ForestGreen}{\cmark}  & \textbf{71.6} & \textbf{53.1} & \textbf{94.9} & 54.5 & \textbf{58.6} & \textbf{79.6} & \textbf{84.0} & \textbf{70.9} \\
        \bottomrule
    \end{tabular}
}
\end{table*}

Tab.~\ref{tab:main_results} reports results and comparative analyses with previous and concurrent published works on the seven core datasets of the BOP Benchmark. All the reported methods use a localization prior estimated in a zero-shot fashion, \ie we only report methods that operate in the same setting for a fair comparison. The table is divided into two sections: the top part lists methods with CNOS localization priors, while the bottom part assumes SAM6D segmentation priors.
\ourmethod consistently outperforms all the works by a considerable margin. 
Specifically, when comparing with methods using CNOS masks and not using any pose refinement, \ourmethod achieves +8.6\% over the second-best method SAM6D~\cite{lin2023sam} (row 5).
When comparing against methods using CNOS masks and pose refinement, \ourmethod outperforms the second-best method SAM6D (row 12) by +6.1\%.
We can observe a significant improvement in TUD-L, where \ourmethod achieves a +8.5\% with respect to MegaPose~\cite{labbe2022megapose} (row 11) and a +13.6\% with respect to SAM6D (row 12).
TUD-L stands as a dataset renowned for its exceptionally challenging lighting conditions.
In this context, \ourmethod stands out by demonstrating the remarkable efficacy of leveraging geometric features to their fullest extent.
Interestingly, \ourmethod without any refinement is even better than the top competitor with refinement SAM6D (row 6 vs row 12).
On average, our refinement step contributes to an increase of +5.2\%. 
In the second part of the table, we compare \ourmethod with SAM6D using their own estimated  masks (SAM6D masks).
Also in this setting, we establish state-of-the-art results for zero-shot object 6D pose estimation, both with and without any refinement. 
Finally, \ourmethod using CNOS masks outperforms SAM6D with SAM6D masks by +1\% (row 13 vs row 16), which is remarkable because SAM6D masks are more accurate than CNOS ones.

%%%%%%%%%%%%%%%%%%%%%%%%%%%%%%%%%%%%%%%%%%%%%%%%
%%%%%%%%%%%%%%%%%%%%%%%%%%%%%%%%%%%%%%%%%%%%%%%%
\subsection{Qualitative results}

\begin{figure*}[t]
\centering
    \begin{tabular}{@{}c@{\,}c@{\,}c@{\,}c@{\,}c@{\,}c@{\,}c@{\,}c}
    \raggedright
        \begin{overpic}[width=0.122\textwidth]
        {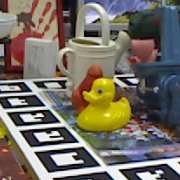}
            \put(-13, 27){\rotatebox{90}{\tiny Input}}
        \end{overpic} &
        \begin{overpic}[width=0.122\textwidth]
        {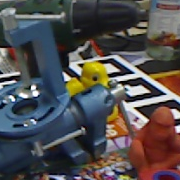}
        \end{overpic} &
        \begin{overpic}[width=0.122\textwidth]
        {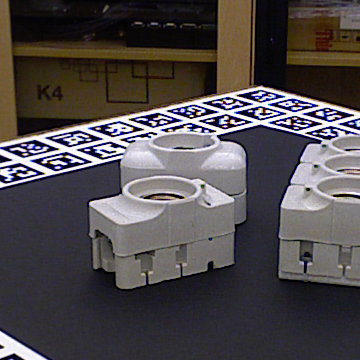}
        \end{overpic} &
        \begin{overpic}[width=0.122\textwidth]
        {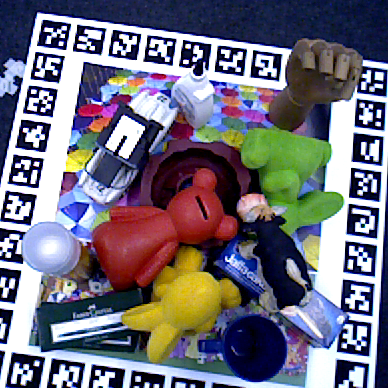}
        \end{overpic} &
        \begin{overpic}[width=0.122\textwidth]{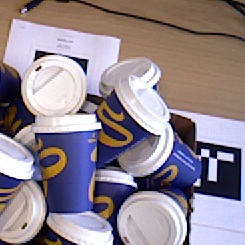}
        \end{overpic} &
        \begin{overpic}[width=0.122\textwidth]
        {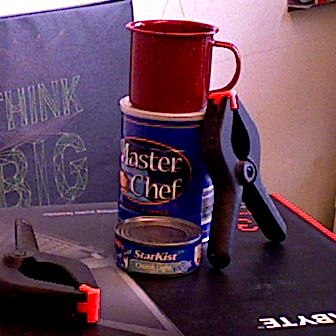}
        \end{overpic} &
        \begin{overpic}[width=0.122\textwidth]{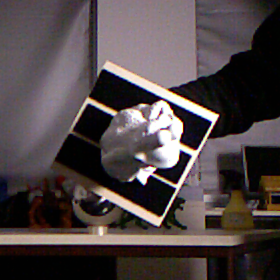}
        \end{overpic} &
        \begin{overpic}[width=0.122\textwidth]
        {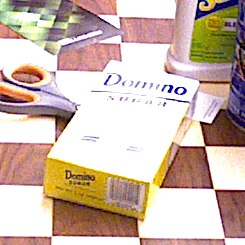}
        \end{overpic}

        \\

        \begin{overpic}[width=0.122\textwidth]{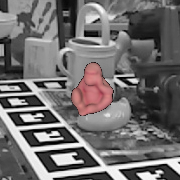}
            \put(-2.5, 4){\colorbox{white}{\scriptsize (a)}}
            \put(-13, 25){\rotatebox{90}{\tiny Output}}
        \end{overpic} &  
        \begin{overpic}[width=0.122\textwidth]{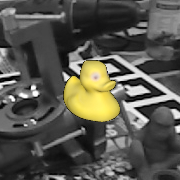}
            \put(-2.5, 4){\colorbox{white}{\scriptsize (b)}}
        \end{overpic} &
        \begin{overpic}[width=0.122\textwidth]{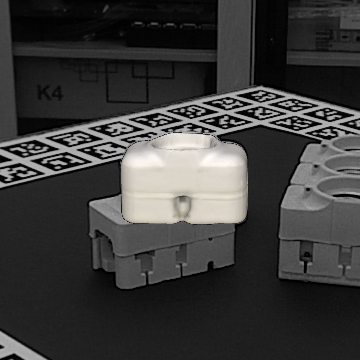}
            \put(-2.5, 4){\colorbox{white}{\scriptsize (c)}}
        \end{overpic} &
        \begin{overpic}[width=0.122\textwidth]{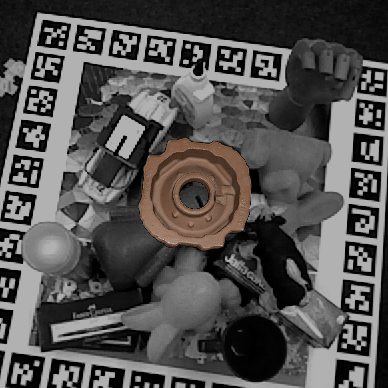}
            \put(-2.5, 4){\colorbox{white}{\scriptsize (d)}}
        \end{overpic} &
        \begin{overpic}[width=0.122\textwidth]{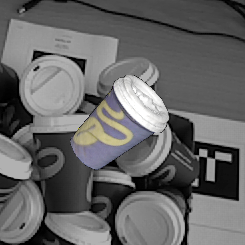}
            \put(-2.5, 4){\colorbox{white}{\scriptsize (e)}}
        \end{overpic} &  
        \begin{overpic}[width=0.122\textwidth]{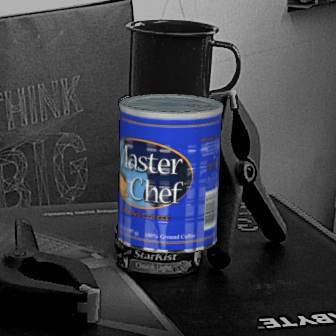}
            \put(-2.5, 4){\colorbox{white}{\scriptsize (f)}}
        \end{overpic} &
        \begin{overpic}[width=0.122\textwidth]{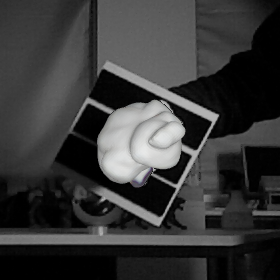}
            \put(-2.5, 4){\colorbox{white}{\scriptsize (g)}}
        \end{overpic} &
        \begin{overpic}[width=0.122\textwidth]        
        {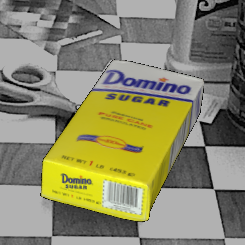}
            \put(-2.5, 4){\colorbox{white}{\scriptsize (h)}}
        \end{overpic} 
    \end{tabular}
    \vspace{-3mm}
    \caption{
    Qualitative results of challenging cases.
    The top row shows input images, while the bottom row shows \ourmethod's predictions by overlaying the object 3D model transformed according to the predicted pose.
    We use a grayscale version of the input image for a better contrast.
    Left to right: 
    (a,b) highly-occluded texture-less toy models, 
    (c,d) highly-occluded texture-less industrial items, 
    (e,f) common objects with cylindrical shapes and informative textures, 
    (g,h) difficult light conditions (strong shadows and a very bright capture).
    Despite these challenges, \ourmethod can always find the correct pose.
    }
    \label{fig:qualitative}
\end{figure*}

Fig.~\ref{fig:qualitative} shows examples of qualitative results, which include several challenges such as occlusions, clutter, variations in background and illumination conditions, poor object textures, severe object symmetries, and similarity with other objects or different instances of the same object.
Fig.~\ref{fig:qualitative}(a,b) show highly-occluded texture-less objects (the toy models of a red ape and a yellow duck); 
Fig.~\ref{fig:qualitative}(c,d) show highly-occluded texture-less symmetric objects (a white electrical outlet and a bronze gear).
To estimate the poses of (a,b,c,d), \ourmethod relies mostly on geometric information, since these objects have non-informative uniform textures.
Fig.~\ref{fig:qualitative}(e,f) show objects with cylindrical shapes (a coffee cup and a food can), where we can observe the presence of multiple instances of the same coffee cup, each with different levels of occlusions.
To estimate the poses of (e,f), \ourmethod relies mostly on visual information, since cylindrical shapes present an infinite number of symmetries, and the correct pose can be established only by relying on texture information.
Lastly, Fig.~\ref{fig:qualitative}(g,h) show images captured under challenging light conditions. 
The uniformly-colored white toy frog in (g) is affected by shadows, while (h) shows a bright capture of a textured sugar box.

\begin{figure}[t]
\centering
    \begin{tabular}{@{}c@{\,}c@{\,}c@{\,}c@{\,}c@{\,}c@{\,}c@{\,}c}
    \raggedright
        \begin{overpic}[width=0.121\columnwidth]
        {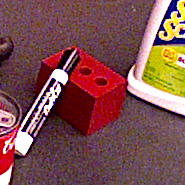}
            %\put(-13, 12){\rotatebox{90}{\scriptsize Input images}}
            \put(-2.5, 4){\colorbox{white}{\scriptsize (a)}}
        \end{overpic} &
        \begin{overpic}[width=0.121\columnwidth]
        {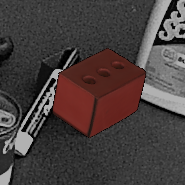}
            %\put(-13, 16){\rotatebox{90}{\scriptsize Predictions}}
        \end{overpic} &
        \begin{overpic}[width=0.121\columnwidth]
        {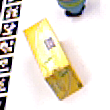}
            \put(-2.5, 4){\colorbox{white}{\scriptsize (b)}}
        \end{overpic} &
        \begin{overpic}[width=0.121\columnwidth]
        {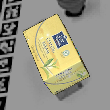}
        \end{overpic} &
        \begin{overpic}[width=0.121\columnwidth]
        {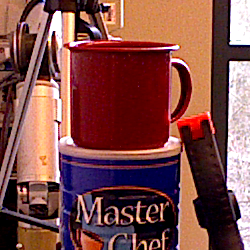}
            \put(-2.5, 4){\colorbox{white}{\scriptsize (c)}}
        \end{overpic} &
        \begin{overpic}[width=0.121\columnwidth]
        {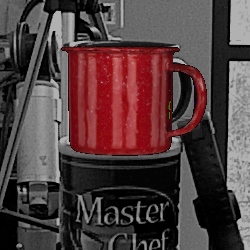}
        \end{overpic} &
        \begin{overpic}[width=0.121\columnwidth]
        {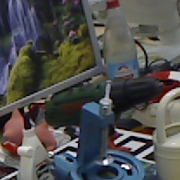}
            \put(-2.5, 4){\colorbox{white}{\scriptsize (d)}}
        \end{overpic} &
        \begin{overpic}[width=0.121\columnwidth]
        {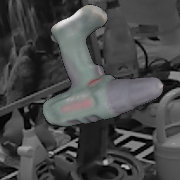}
        \end{overpic}
    \end{tabular}

    \vspace{-3mm}
    \caption{
    Failure cases. \ourmethod's predictions follow the relative input images.
    Left to right:
    (a) the foam block is aligned only to the closest corner, in a pose that is almost perpendicular to the correct one,
    (b) the tea box pose is predicted as if it was laying on its side rather than on its bottom,
    (c) the main body of the mug is nearly aligned, but the handle is not,
    (d) the pose of the drill handle (occluded) is incorrectly estimated.
    }
    \label{fig:failures}
\end{figure}

Fig.~\ref{fig:failures} shows example of failure cases.
In Fig.~\ref{fig:failures}(a), the red foam block is aligned only to the closest corner, with a pose nearly perpendicular to the correct one. 
This can be due to the part of the object that is occluded by the marker.
In Fig.~\ref{fig:failures}(b), the yellow tea box pose is predicted laying on its side rather than on its bottom.
Because of the object's simple geometry and poor texture, we believe that aligning most of the corners of the box regardless of its texture can be acceptable for \ourmethod.
In Fig.~\ref{fig:failures}(c), the main body of the red mug is fairly aligned, except the handle.
This may be due to the limited surface area of the handle that causes \ourmethod to prioritize aligning the rest of the object.
In Fig.~\ref{fig:failures}(d), the pose of the green drill handle is incorrectly estimated.
The handle is fully occluded in the input image, so that predicting a correct pose is challenging without additional prior information. 
The visible portion (main body) is correctly aligned. 

%%%%%%%%%%%%%%%%%%%%%%%%%%%%%%%%%%%%%%%%%%%%%%%%
%%%%%%%%%%%%%%%%%%%%%%%%%%%%%%%%%%%%%%%%%%%%%%%%
\subsection{Ablation study}\label{sec:ablation-study}

\ourmethod integrates geometric and visual features, making the choice of specific encoders both critical and carefully considered. 
As presented in Tab.~\ref{tab:ablations}, we conduct a thorough ablation study on LM-O~\cite{lmo} to ensure the optimal selection and to assess each component of \ourmethod.  
Unless otherwise specified, we do not apply any pose refinement.
Tab.~\ref{tab:ablation_refinement} shows the ablation study on the refinement step.
In both cases, results are reported in terms of AR. We use as localization prior CNOS segmentation masks across all the experiments.

%%%%%%%%%%%%%%%%%%%%%%%%%%%%%%%%%%%%%%%%%%%%%%%%
\noindent\textbf{Feature types.}~We assess the fusion of geometric and visual features.
In rows 1--3 of Tab.~\ref{tab:ablations}, we evaluate different geometric feature extraction methods: 
hand-crafted approach FPFH~\cite{fpfh}, and 
learning-based techniques FCGF~\cite{fcgf} and 
GeDi~\cite{gedi}.
In rows 9--11 of Tab.~\ref{tab:ablations}, we evaluate different vision encoders: 
simple RGB values, 
and the foundation models CLIP~\cite{clip} and DINOv2~\cite{dinov2}.
We observed that:
(i) GeDi is highly effective across scenes (+34.5 in AR w.r.t.~FCGF);
(ii) DINOv2 outperforms CLIP by about +6 in AR despite they are both based on ViT-L;
(iii) geometric features outperforms visual ones by a significant margin.

%%%%%%%%%%%%%%%%%%%%%%%%%%%%%%%%%%%%%%%%%%%%%%%%
\noindent\textbf{Multi-scale geometric feature fusion.}~GeDi~\cite{gedi} uses a single radius to define the extent of local descriptors.
Instead, we use multiple radii to encode information at multiple scales.
In rows 4--8 of Tab.~\ref{tab:ablations}, we compare the original single-scale version of GeDi with the proposed multi-scale fusion of GeDi features for different radii.
Radii are expressed as a ratio of the object diameter, \ie~0.2 corresponds to 20\% of the object diameter.
The best AR is achieved when using multiple scales.
Another crucial advantage of using multiple scales lies in its generalizability.
We experimentally observed that the best-performing single-scale radius is dataset dependent, whereas using multiple scales allow us to use the same configuration consistently across all the seven core datasets.

\begin{table}[t!]

\centering
\tabcolsep 9pt
\caption{
Ablation study on the LM-O~\cite{lmo} dataset. We individually asses different components of \ourmethod. MS-GeDi is our proposed multi-scale GeDi.
Keys.
Prior: type of instance localization prior;
m: segmenation mask;
bb: bounding box;
`-': not available.
}
\label{tab:ablations}
\vspace{-3.5mm}

\resizebox{\columnwidth}{!}{%
\begin{tabular}{ccccccccc}
    \toprule
    & &\multicolumn{2}{c}{Geometric encoder} & \multicolumn{2}{c}{Vision encoder} & \multirow{2}{*}{Prior}  & \multirow{2}{*}{ICP}& \multirow{2}{*}{AR} \\
    & & Method & Radius & Method & Backbone & \\
    \midrule
    \multirow{9}{*}{\color{gray}\rotatebox{90}{Geometry}} & {\color{gray} \small 1}  & FPFH & 0.3 & - & - & m & & 20.7 \\
    &{\color{gray} \small 2}  & FCGF & -   & - &  - & m & & 20.8\\
    &{\color{gray} \small 3}  & GeDi & 0.3 & - &  - & m & & \textbf{55.3} \\
    \cmidrule{2-9} 
    &{\color{gray} \small 4}  & GeDi & 0.2 & - &  - & m & & 52.6 \\
    &{\color{gray} \small 5}  & GeDi & 0.3 & - &  - & m & & 55.3 \\
    &{\color{gray} \small 6}  & GeDi & 0.4 & - &  - & m & & 55.0 \\
    &{\color{gray} \small 7}  & MS-GeDi & (0.2, 0.3) & - &  - & m & & 56.2 \\
    &{\color{gray} \small 8}  & MS-GeDi & (0.3, 0.4) & - &  - & m & & \textbf{56.5} \\  
    \midrule
    \multirow{7}{*}{\color{gray}\rotatebox{90}{Vision}} &{\color{gray} \small 9}  & - & - & RGB & - &  m & & 15.1 \\ 
    &{\color{gray} \small 10}  & - & - & CLIP &  ViT-L & m & & 30.5 \\
    &{\color{gray} \small 11}  & - & - & DINOv2 & ViT-L & m & & \textbf{36.8} \\
    \cmidrule{2-9} 
    &{\color{gray} \small 12}  & - & - & DINOv2 & ViT-S & m & & 31.9 \\
    &{\color{gray} \small 13}  & - & - & DINOv2 & ViT-B & m & & 33.7 \\
    &{\color{gray} \small 14}  & - & - & DINOv2 & ViT-L & m & & 36.8 \\
    & {\color{gray} \small 15}  & - & - & DINOv2 & ViT-G & m & & \textbf{39.6} \\
    \midrule
    \multirow{4}{*}{\color{gray}\rotatebox{90}{Prior}}  &{\color{gray} \small 16}  & MS-GeDi & (0.3, 0.4) & DINOv2 & ViT-G & bb & & 64.7 \\
    &{\color{gray} \small 17}  & MS-GeDi & (0.3, 0.4) & DINOv2 & ViT-G & bb & \color{ForestGreen}{\cmark} & 68.6 \\
    &{\color{gray} \small 18}  & MS-GeDi & (0.3, 0.4) & DINOv2 & ViT-G & m & & 64.7 \\
    %\rowcolor{cosmiclatte}
    &{\color{gray} \small 19}  & MS-GeDi & (0.3, 0.4) & DINOv2 & ViT-G & m & \color{ForestGreen}{\cmark} & \textbf{69.0} \\
    \bottomrule
\end{tabular}
}
\end{table}

%%%%%%%%%%%%%%%%%%%%%%%%%%%%%%%%%%%%%%%%%%%%%%%%
\noindent\textbf{Visual features backbone.}~In rows 12--15 of Tab.~\ref{tab:ablations}, we assess different DINOv2 ViT backbones: small (S), base (B), large (L), and giant (G) models. 
The latter achieves the highest AR (+2.8 in AR w.r.t.~ViT-L, the CLIP backbone).

%%%%%%%%%%%%%%%%%%%%%%%%%%%%%%%%%%%%%%%%%%%%%%%%
\noindent\textbf{Localization prior type.}~In rows 16--19 of Tab.~\ref{tab:ablations}, we test a bounding box localization prior rather than a segmentation mask prior.
Without using any pose refinement, bounding boxes and segmentation masks produce the same results.
However, once when refine the coarse pose using ICP, \ourmethod performs betters with the segmentation masks as localization prior instead of the bounding boxes.
This suggests that our coarse pose estimation is robust to background variations, while the ICP-based refinement step is sensitive to background outliers.

%%%%%%%%%%%%%%%%%%%%%%%%%%%%%%%%%%%%%%%%%%%%%%%%
\noindent\textbf{Pose refinement.}
In Tab.~\ref{tab:ablation_refinement}, we evaluate the effectiveness of the pose refinement by comparing the coarse estimation (row 1) with the pose obtained using an ICP-based refinement (row 2), and the one obtained adding also our Symmetry-Aware Refinement (SAR for short, row 3).
SAR gains the most improvement over the basic ICP on datasets containing geometrically symmetric objects with discriminant texture, particularly on IC-BIN (+2.6 in AR), HB (+0.6 in AR) and YCB-V(+0.4 in AR).
Instead, on datasets like LM-O, T-LESS, TUD-L and ITODD, SAR does not provide any advantage compared to solely relying on ICP.

\begin{table}[t!]
    \centering
    \tabcolsep 5pt
    \caption{
        Ablation study on the pose refinement module. 
        We report the AR scores achieved without any refinement, with only an ICP-based refinement and with the complete ICP+SAR refinement on the seven core datasets of the BOP Benchmark. 
        Keys. Refin.: pose refinement; 
Mean: Mean AR.
    }
    \label{tab:ablation_refinement} 
    \vspace{-3.5mm}

    \resizebox{\columnwidth}{!}{%
    \begin{tabular}{llcccccccc}
        \toprule
        & Refin. & LM-O & T-LESS & TUD-L & IC-BIN & ITODD & HB & YCB-V & Mean \\
        \toprule
        {\color{gray} \small 1} & None  & 64.7 & 49.3  & 86.1 & 44.3 &  49.2 & 75.7 & 78.7 & 64.0 \\
        {\color{gray} \small 2} & +ICP & \textbf{69.0} &  \textbf{52.0} & \textbf{93.6} & 47.3 & \textbf{56.1} & 78.4 & 84.9 & 68.8 \\
        {\color{gray} \small 3} & +SAR & \textbf{69.0} & \textbf{52.0} & \textbf{93.6} & \textbf{49.9} & \textbf{56.1} & \textbf{79.0} & \textbf{85.3} & \textbf{69.3} \\
        \bottomrule
    \end{tabular}
    }
\end{table}

\section{Conclusions}

We presented \ourmethod, a novel approach to zero-shot object 6D pose estimation that leverages the strengths of pre-trained geometric and vision foundation models without the need for training on task-specific data. 
By leveraging 3D geometric features and 2D visual features to create discriminative 3D point-level descriptors, \ourmethod outperforms all competitors on the BOP Benchmark, which includes seven datasets, setting a new state-of-the-art bar for the field.
However, \ourmethod faces certain \textbf{limitations}, which include the large size of foundation models that may restrict their deployment on edge devices, limiting real-world applicability in certain scenarios.
One can mitigate this by using distillation techniques.
Another \textbf{future research} direction includes improving our 3D registration process, possibly by advancing beyond simple RANSAC algorithms.

\section*{Acknowledgements}
This work was supported by the European Union's
Horizon Europe research and innovation programme under grant agreement No 101058589 (AI-PRISM), and by the PNRR project FAIR -- Future AI Research (PE00000013), under the NRRP MUR program funded by the NextGenerationEU.

\clearpage
\bibliographystyle{eccv24toolkit/splncs04}
\bibliography{main}

\pagestyle{headings}
\mainmatter

\title{{\color{NavyBlue}\ourmethod}: Training-{\color{NavyBlue}free ze}ro-shot 6D pose estimation with geometric and vision foundation models}
\titlerunning{FreeZe: Training-free zero-shot 6D pose estimation}

\author{Andrea Caraffa\inst{1} \and
Davide Boscaini\inst{1} \and
Amir Hamza\inst{1,2} \and
Fabio Poiesi \inst{1}}

\authorrunning{A.~Caraffa et al.}

\institute{Fondazione Bruno Kessler, Trento, Italy 
\email{\{acaraffa,dboscaini,ahamza,poiesi\}fbk.eu}
\and
University of Trento, Italy}
\title{Supplementary material for\\ {\color{NavyBlue}\ourmethod}: Training-{\color{NavyBlue}free ze}ro-shot 6D pose estimation with geometric and vision foundation models}

\maketitle

\section{Introduction}\label{sec:supp_intro}

We present the supplementary material in support of our main paper.
The content is organized as follows: 
\begin{itemize}[noitemsep,nolistsep,leftmargin=*]
    \item In Sec.~\ref{sec:supp_feat}, we evaluate the effectiveness of entangling geometric and visual features compared to using the two components independently.
    \item In Sec.~\ref{sec:supp_sar}, we highlight the role of the proposed Symmetry-Aware Refinement~(SAR) in resolving ambiguous poses of geometrically symmetric objects.
    \item In Sec.~\ref{sec:supp_visib}, we investigate how object occlusions affect pose estimation accuracy.
    \item In Sec.~\ref{sec:supp_timings}, we provide an analysis of the computational time.
    \item In Sec.~\ref{sec:supp_bop_qual}, we provide additional qualitative results on the seven core datasets of the BOP Benchmark~\cite{sundermeyer2023bop}.
\end{itemize}

\section{Effectiveness of geometric and visual feature entanglement}\label{sec:supp_feat}

\begin{figure*}[t!]

    \raggedright
    \begin{minipage}{0.32\textwidth}
        \centering
        \footnotesize (a)
    \end{minipage}
    \begin{minipage}{0.32\textwidth}
        \centering
        \footnotesize (b)
    \end{minipage}
    \begin{minipage}{0.32\textwidth}
        \centering
        \footnotesize (c)
    \end{minipage}

    \vspace{1.mm}
    \begin{tabular}{@{}c@{\,}c@{\,}c}
    \raggedright
    %
    % first row
    %
    \begin{overpic}[width=0.33\textwidth]{supp/images/feats_preds/000701_0_input.png}
        \put(-10, 14){\rotatebox{90}{\footnotesize Input images}}
    \end{overpic} &
    \begin{overpic}[width=0.33\textwidth]{supp/images/feats_preds/001062_0_input.png}
    \end{overpic} &
    \begin{overpic}[width=0.33\textwidth]{supp/images/feats_preds/000058_0_input.png}
    \end{overpic} \\
    %
    % second row
    %
    \begin{overpic}[width=0.33\textwidth]{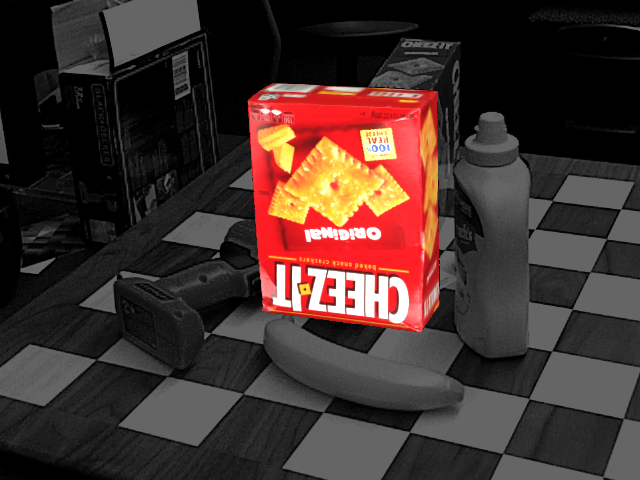}
        \put(-10, 4){\rotatebox{90}{\footnotesize Geometric encoder}}
    \end{overpic} &
    \begin{overpic}[width=0.33\textwidth]{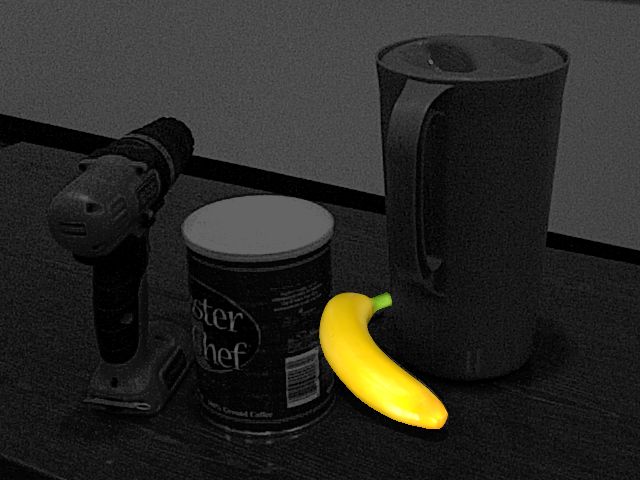}
    \end{overpic} &
    \begin{overpic}[width=0.33\textwidth]{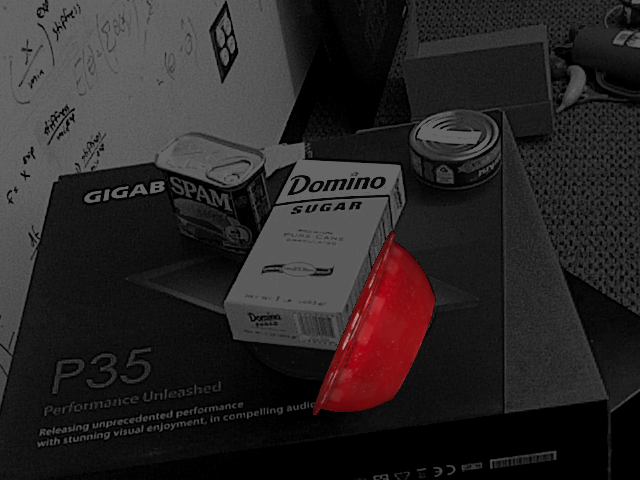}
    \end{overpic} \\
    %
    % third row
    %
    \begin{overpic}[width=0.33\textwidth]{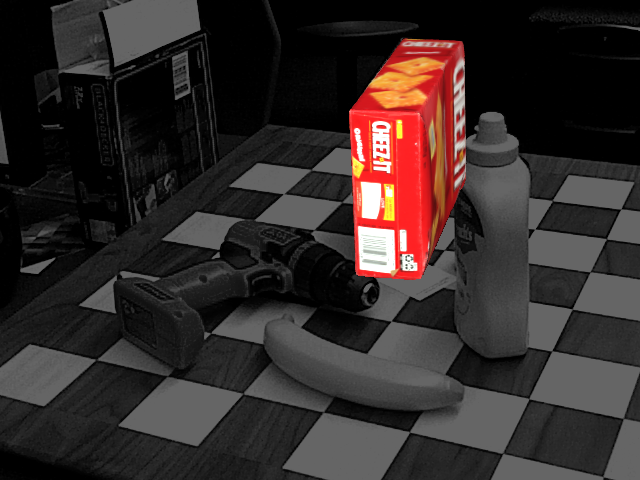}
        \put(-10, 12){\rotatebox{90}{\footnotesize Vision encoder}}
    \end{overpic} &
    \begin{overpic}[width=0.33\textwidth]{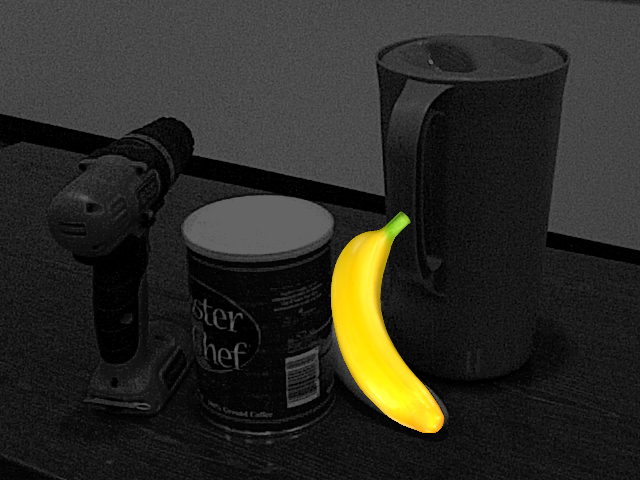}
    \end{overpic} &
    \begin{overpic}[width=0.33\textwidth]{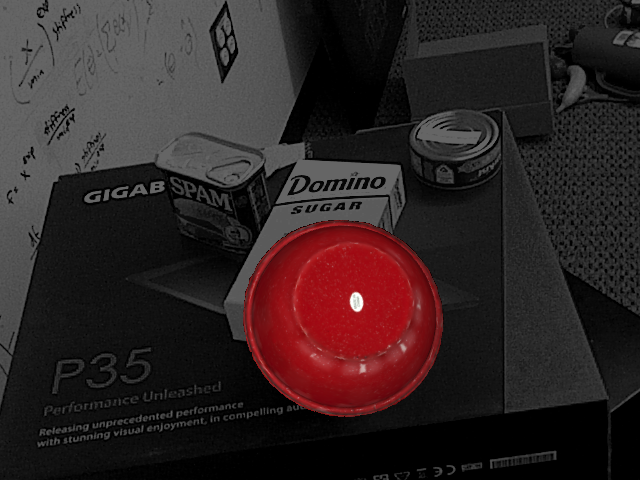}
    \end{overpic} \\
    %
    % fourth row
    %
    \begin{overpic}[width=0.33\textwidth]{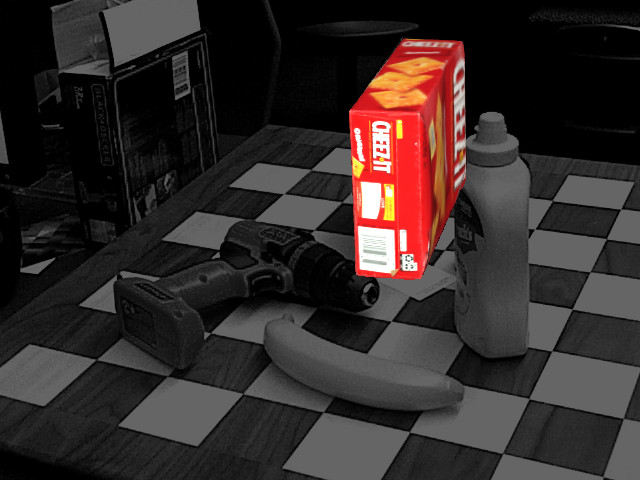}
        \put(-10, 25){\rotatebox{90}{\footnotesize \ourmethod}}
    \end{overpic} &
    \begin{overpic}[width=0.33\textwidth]{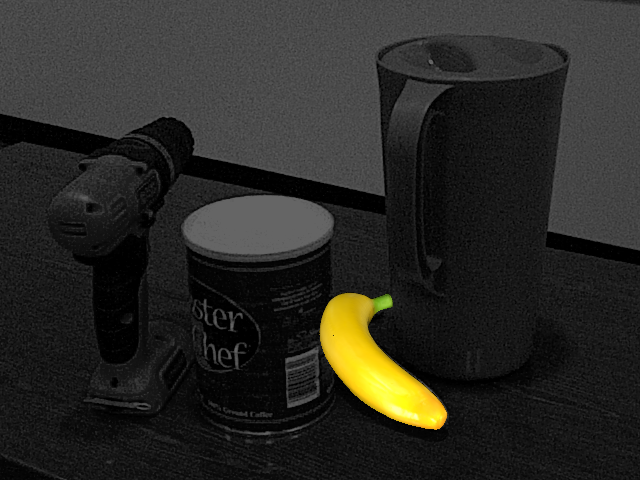}
    \end{overpic} &
    \begin{overpic}[width=0.33\textwidth]{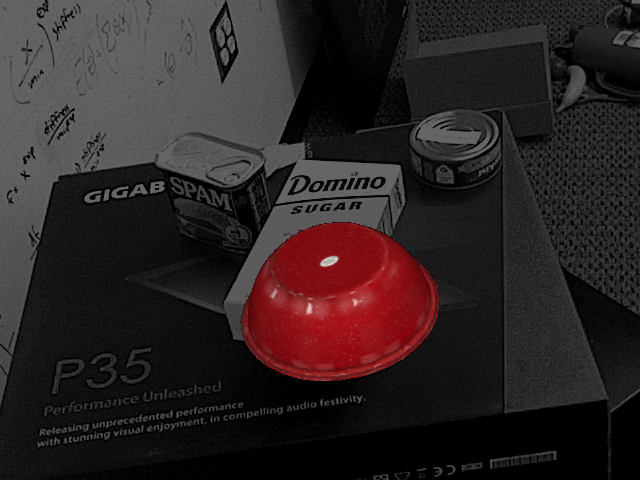}
    \end{overpic} \\
    \end{tabular}

    \vspace{-3mm}
    \caption{
    Qualitative results illustrating the effectiveness of geometric and visual feature entanglement in \ourmethod.
    Columns show three images from YCB-V~\cite{ycbv}.
    Rows show, from top to bottom, the input image, the pose predicted using exclusively geometric features, the pose predicted using exclusively visual features, and the pose predicted using our fused features.
    Backgrounds are converted to grayscale to enhance contrast.
    }
    \label{fig:feats_preds}
\end{figure*}

We present qualitative results to illustrate the contribution of geometric and visual features within \ourmethod.
Fig.~\ref{fig:feats_preds} showcases examples of poses predicted using exclusively geometric features~(second row), exclusively visual features~(third row), and their fusion~(fourth row).
In column~(a), geometric features alone fail to accurately estimate the pose of the red cracker box, while visual features succeed in doing so.
Conversely, in column~(b), visual features alone struggle, while geometric features accurately predict the pose of the yellow banana.
Finally, in column~(c), both geometric and visual features fail to correctly estimate the pose of the red bowl when used individually.
Through feature fusion, \ourmethod capitalizes on the strengths of both feature types, enabling precise pose predictions for both~(a) and (b).
Remarkably, \ourmethod achieves accurate pose estimation even when both feature types independently falter, as illustrated in column (c).

Our experimental observations indicate that geometric features extracted from objects with geometric symmetries, such as boxes and cylinders, can lead to two types of pose estimation errors.
In the first case, the predicted pose might be inaccurate, as depicted in column~(a).
In the second case, although the object is positioned correctly, it may have an incorrect rotation about one of its symmetry axes.
The proposed Symmetry-Aware Refinement (SAR) module can only address errors of the second type but fails to solve errors of the first type.
Therefore, the fusion of visual features with geometric ones is crucial for facilitating accurate pose predictions.

\begin{figure*}[th!]
    \centering
    \begin{overpic}[width=0.9\textwidth]{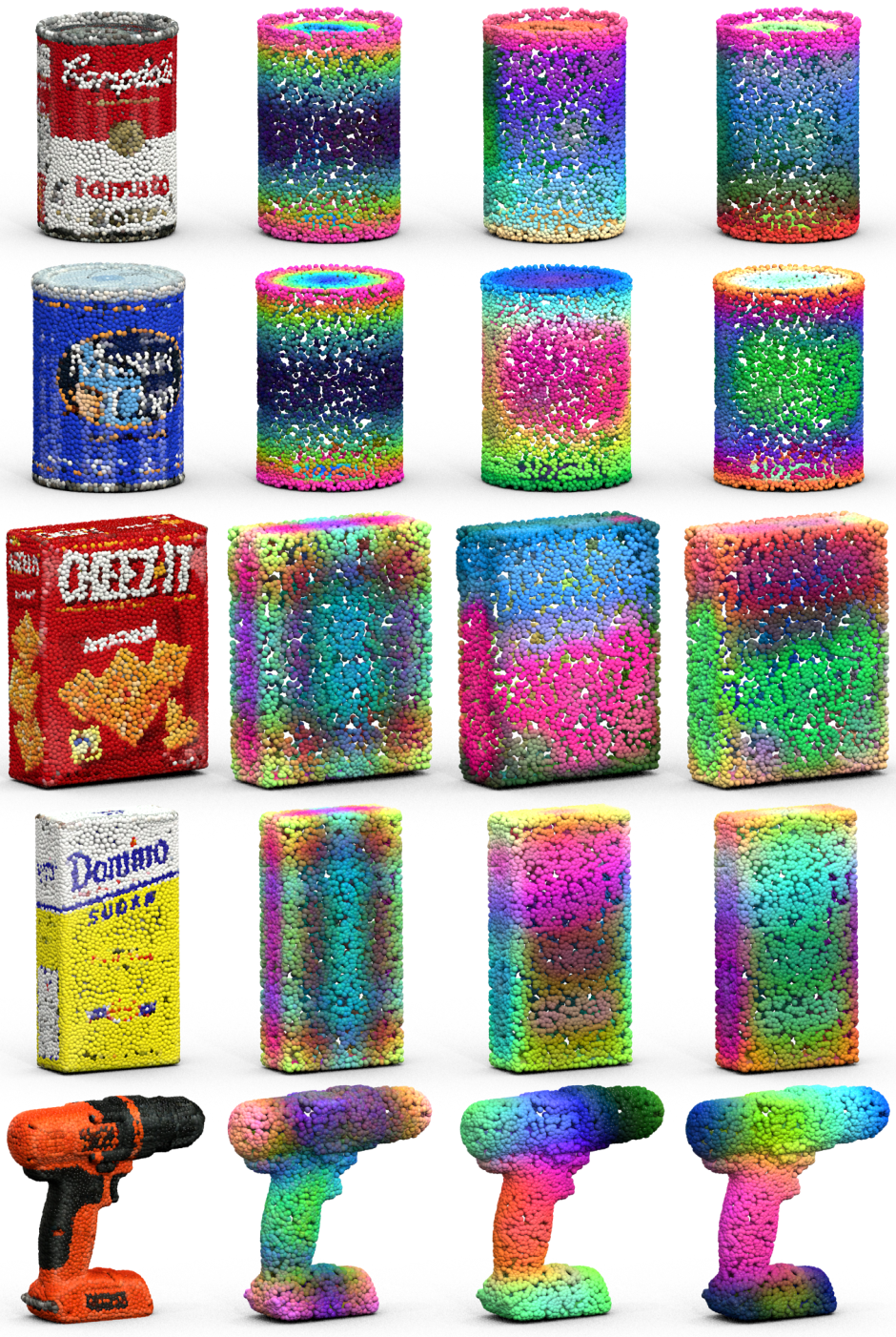}   \end{overpic}

    \vspace{-2mm}
    \begin{minipage}{0.24\textwidth}
        \centering
        \footnotesize Input
    \end{minipage}
    \begin{minipage}{0.24\textwidth}
        \centering
        \footnotesize Geometric features
    \end{minipage}
    \begin{minipage}{0.24\textwidth}
        \centering
        \footnotesize Visual features
    \end{minipage}
    \begin{minipage}{0.24\textwidth}
        \centering
        \footnotesize Our fused features
    \end{minipage}

    \vspace{-2mm}
    \caption{
    Visualization of the different types of features considered by \ourmethod.
    Rows show different objects from YCB-V~\cite{ycbv}.
    Columns show, from left to right, the 3D model of the object, the geometric features, the visual features, and our fused features.
    }
    \label{fig:supp_feat_pca}
\end{figure*}

Fig.~\ref{fig:supp_feat_pca} visualizes geometric, visual, and our fused features in the RGB space by reducing their dimensionality via Principal Component Analysis.
Interestingly, different instances of cans and boxes have similar geometric features, while this is not true for our fused features thanks to the integration with visual features. 

\section{Impact of Symmetry-Aware Refinement}\label{sec:supp_sar}

\begin{figure*}[th!]
\centering

    \raggedright
    \begin{minipage}{0.32\textwidth}
        \centering
        \footnotesize Input
    \end{minipage}
    \begin{minipage}{0.32\textwidth}
        \centering
        \footnotesize Before SAR
    \end{minipage}
    \begin{minipage}{0.32\textwidth}
        \centering
        \footnotesize After SAR
    \end{minipage}

    \vspace{1.6 mm}
    \begin{tabular}{@{}c@{\,}c@{\,}c}
    \raggedright
    %
    % first row
    %
    \begin{overpic}[width=0.33\textwidth]{supp/images/sar/20_input.png}
        \put(43.2,33.75){\color{green}\setlength{\fboxrule}{2.5 pt}\framebox(15,12){}}  % 41.5, (18, 12)
    \end{overpic} &
    \begin{overpic}[width=0.33\textwidth]{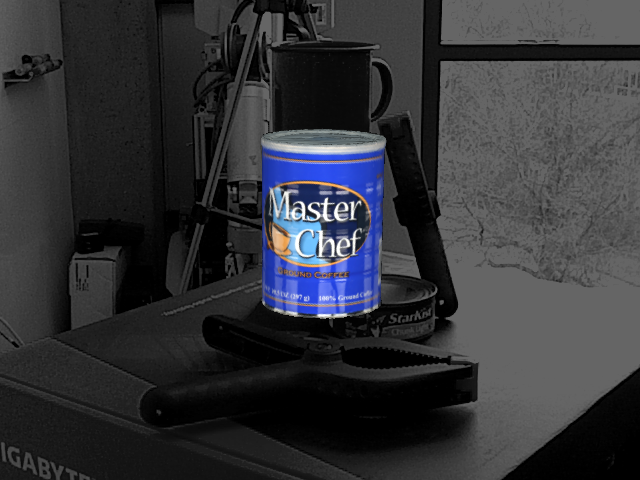}
         \put(43.2,33.75){\color{green}\setlength{\fboxrule}{2.5 pt}\framebox(15,12){}}
        \put(-0.3, 2.2){\colorbox{white}{\footnotesize AR=75.3}}
    \end{overpic} &
    \begin{overpic}[width=0.33\textwidth]{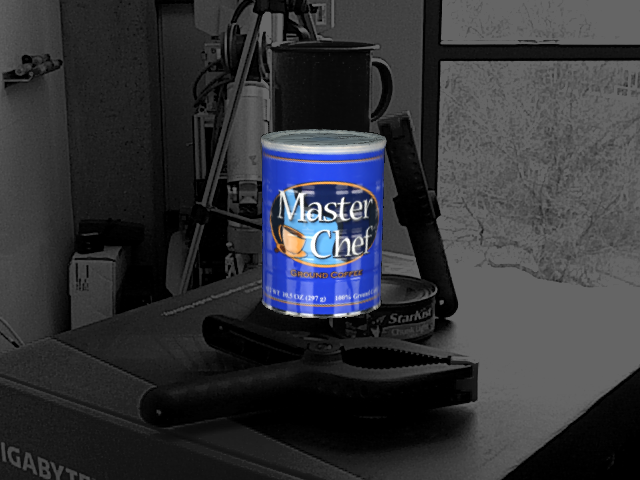}
        \put(43.2,33.75){\color{green}\setlength{\fboxrule}{2.5 pt}\framebox(15,12){}}
        \put(-0.3, 2.2){\colorbox{white}{\footnotesize AR=85.3}}
    \end{overpic} \\
    %
    % second row
    %
    \begin{overpic}[width=0.33\textwidth]{supp/images/sar/782_input.png}
        \put(69.1,33.5){\color{green}\setlength{\fboxrule}{2.5 pt}\framebox(5.75,17.05){}}
        \linethickness{2.1pt}
    \end{overpic} &
    \begin{overpic}[width=0.33\textwidth]{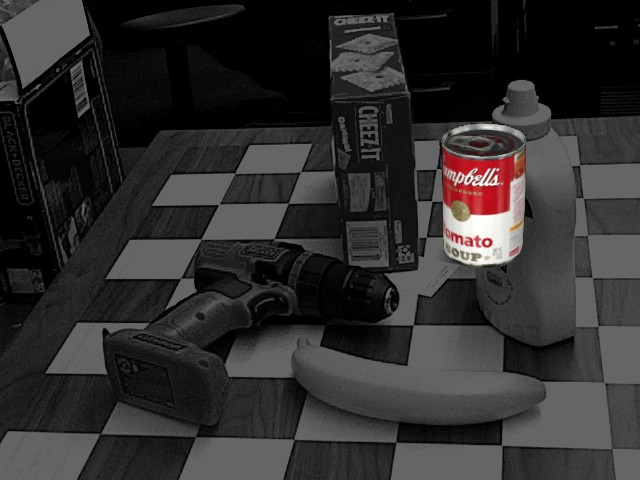}
         \put(69.1,33.5){\color{green}\setlength{\fboxrule}{2.5 pt}\framebox(5.75,17.05){}}
        \put(-0.3, 2.2){\colorbox{white}{\footnotesize AR=77.3}}
        \linethickness{2.1pt}
    \end{overpic} &
    \begin{overpic}[width=0.33\textwidth]{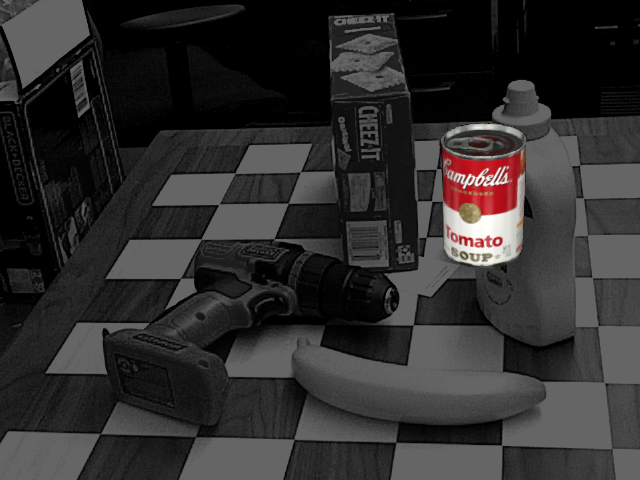}
        \put(69.1,33.5){\color{green}\setlength{\fboxrule}{2.5 pt}\framebox(5.75,17.05){}}
        \put(-0.3, 2.2){\colorbox{white}{\footnotesize AR=87.3}}
        \linethickness{2.1pt}
    \end{overpic} \\
    %
    % third row
    %
    \begin{overpic}[width=0.33\textwidth]{supp/images/sar/453_input.png}
        \put(35.75,41){\color{green}\setlength{\fboxrule}{2.5 pt}\framebox(6.2,7.5){}}
    \end{overpic} &
    \begin{overpic}[width=0.33\textwidth]{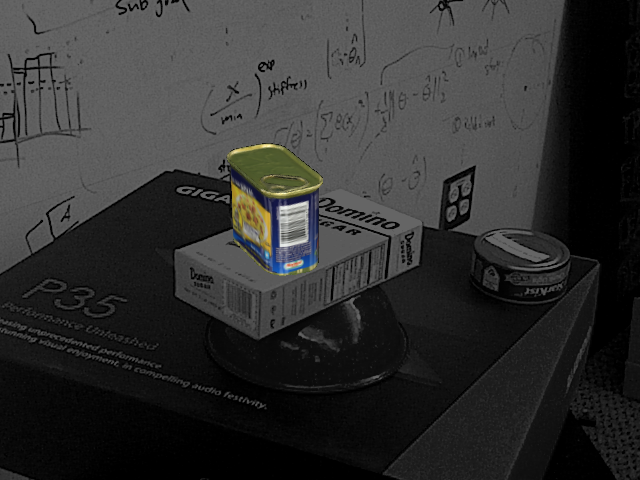}
        \put(35.75,41){\color{green}\setlength{\fboxrule}{2.5 pt}\framebox(6.2,7.5){}}
        \put(-0.3, 2.2){\colorbox{white}{\footnotesize AR=32.3}}
        \linethickness{2.1pt}
    \end{overpic} &
    \begin{overpic}[width=0.33\textwidth]{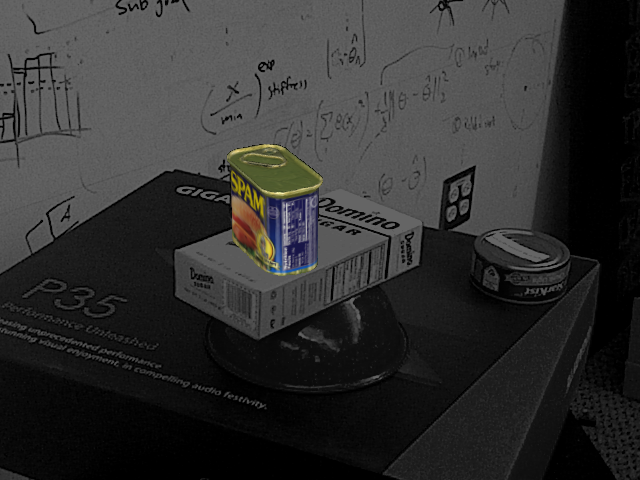}
        \put(35.75,41){\color{green}\setlength{\fboxrule}{2.5 pt}\framebox(6.2,7.5){}}
        \put(-0.3, 2.2){\colorbox{white}{\footnotesize AR=95.6}}
        \linethickness{2.1pt}
    \end{overpic} \\
    %
    % fourth row
    %
    \begin{overpic}[width=0.33\textwidth]{supp/images/sar/739_input.png}
        \put(74.5,22.85){\color{green}\setlength{\fboxrule}{2.5 pt}\framebox(6.5,23){}}
        \linethickness{2.1pt}
    \end{overpic} &
    \begin{overpic}[width=0.33\textwidth]{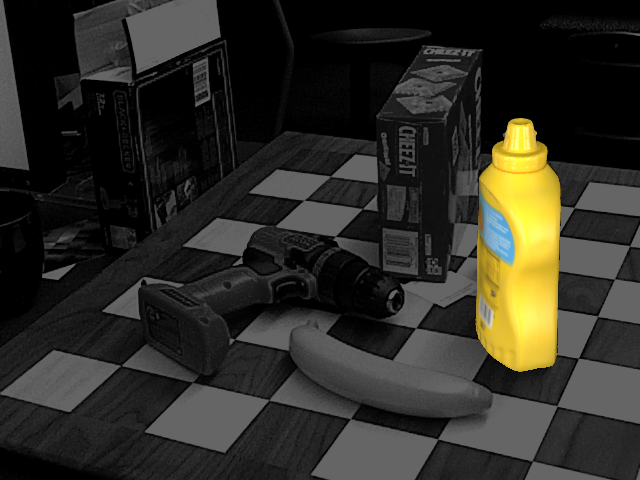}
        \put(74.5,22.85){\color{green}\setlength{\fboxrule}{2.5 pt}\framebox(6.5,23){}}
        \put(-0.3, 2.2){\colorbox{white}{\footnotesize AR=30.0}}
        \linethickness{2.1pt}
   \end{overpic} &
        \begin{overpic}[width=0.33\textwidth]{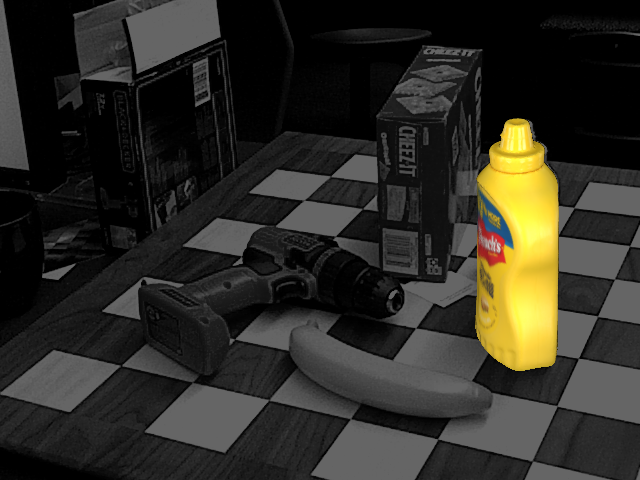}
        \put(74.5,22.85){\color{green}\setlength{\fboxrule}{2.5 pt}\framebox(6.5,23){}}
        \put(-0.3, 2.2){\colorbox{white}{\footnotesize AR=93.0}}
        \linethickness{2.1pt}
    \end{overpic} \\
    %
    % fifth row
    %
    \begin{overpic}[width=0.33\textwidth]{supp/images/sar/1864_input.png}
        \put(21,56){\color{green}\setlength{\fboxrule}{2.5 pt}\framebox(22,12){}}  % (17.25,32.25), (32.5,41)
        \linethickness{2.1pt}
    \end{overpic} &
    \begin{overpic}[width=0.33\textwidth]{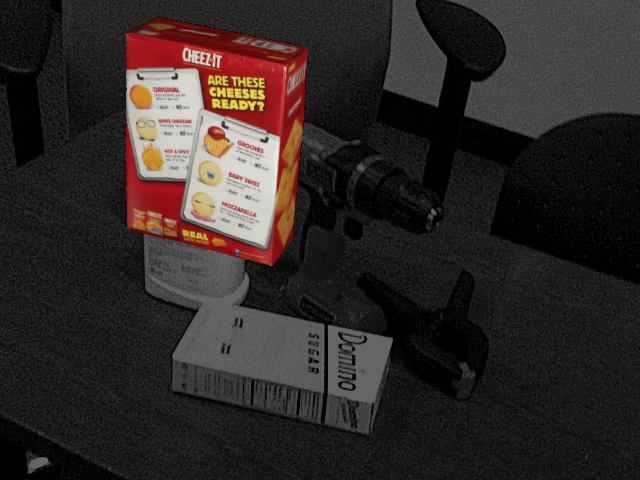}
        \put(21,56){\color{green}\setlength{\fboxrule}{2.5 pt}\framebox(22,12){}}
        \put(-0.3, 2.2){\colorbox{white}{\footnotesize AR=33.3}}
    \end{overpic} &
    \begin{overpic}[width=0.33\textwidth]{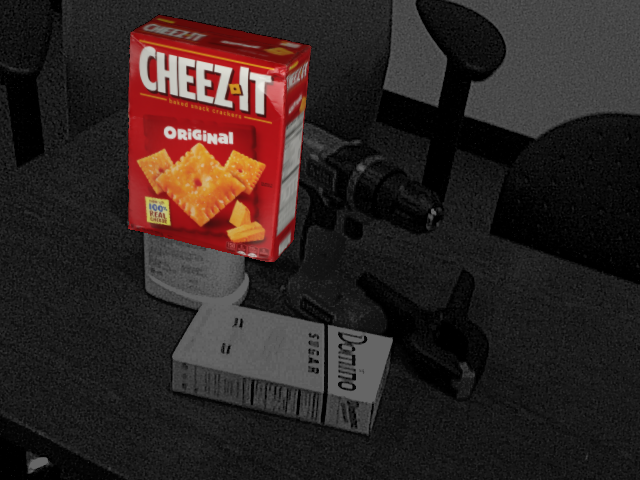}
        \put(21,56){\color{green}\setlength{\fboxrule}{2.5 pt}\framebox(22,12){}}
        \put(-0.3, 2.2){\colorbox{white}{\footnotesize AR=100.0}}
    \end{overpic} \\
    \end{tabular}

    \vspace{-3mm}
    \caption{
    Qualitative results for the proposed Symmetry-Aware Refinement (SAR) module on five images from YCB-V~\cite{ycbv}.
    Rows show different examples.
    Columns show, from left to right, the input data, the coarse pose prediction, and the SAR-refined pose prediction.
    Backgrounds are converted to grayscale to enhance contrast. The overlaid green boxes highlight regions where pose correction  by SAR is more visible.
    }
    \label{fig:sar}
\end{figure*}

We evaluate the effectiveness of our proposed Symmetry-Aware Refinement (SAR) module in correcting the ambiguous poses of geometrically symmetric objects. 
In Fig.~\ref{fig:sar}, we illustrate the pose correction for various objects from YCB-V~\cite{ycbv}. 
The first three rows depict cans of different types and levels of occlusion. 
SAR is capable of correcting minor errors in the estimated poses. 
The last two rows present cases where poses are incorrectly flipped by approximately 180 degrees. 
Such errors are infrequent, as the majority of poses are accurately estimated during the initial coarse pose estimation phase. 
In such instances, SAR successfully corrects the poses, despite significant rotational errors. 
Most of the poses of geometrically symmetric objects are correctly predicted thanks to the fusion with visual features, while about 15\% of them are corrected using SAR.

\section{Impact of object occlusions}\label{sec:supp_visib}

\begin{figure}[t!]    
    \begin{subfigure}{0.5\textwidth}
    \centering
    \includegraphics[width=1.0\linewidth]{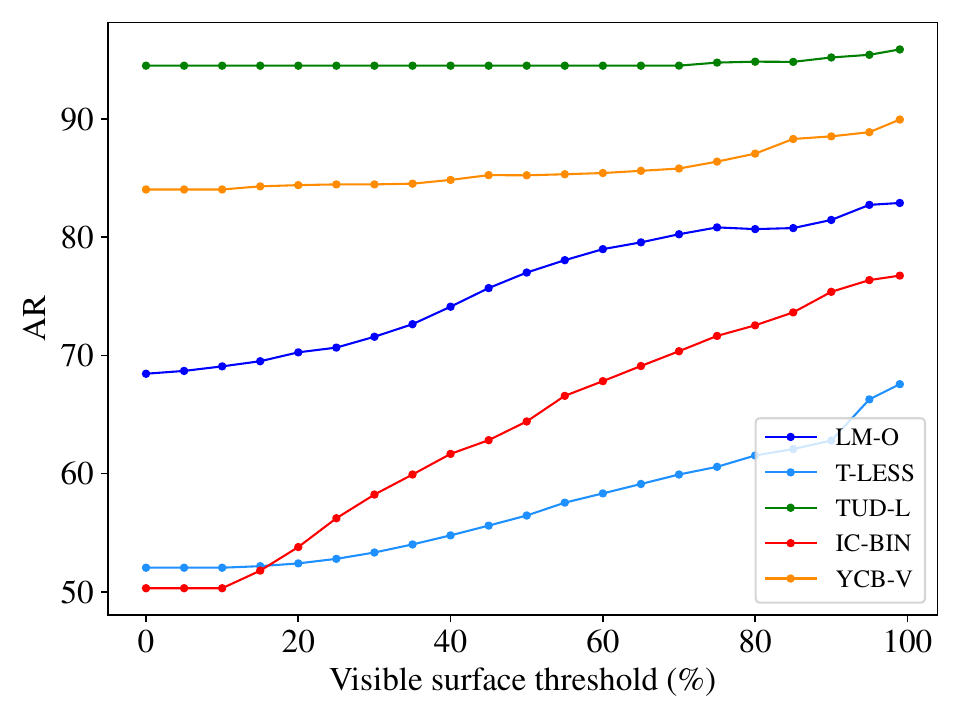}
    \end{subfigure}
    \begin{subfigure}{0.5\textwidth}
    \centering
    \includegraphics[width=1.0\linewidth]{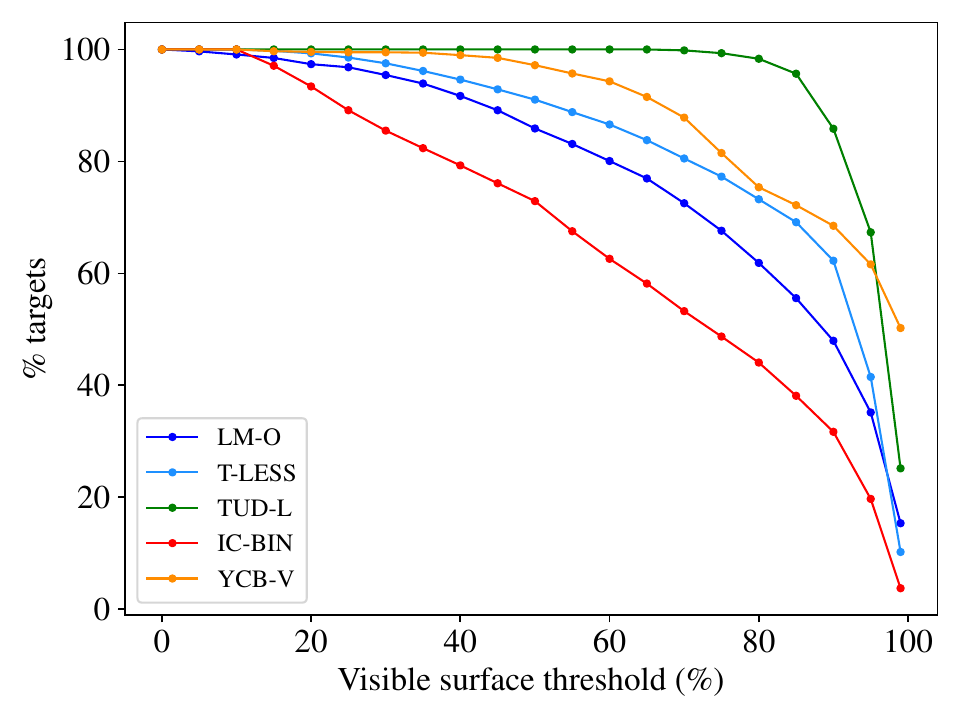}
    \end{subfigure}
  
    \vspace{-2mm}
    \caption{
    Left: Average Recall (AR) for target objects selected according to different visible surface thresholds. The visible surface is computed as the IoU of the ground-truth mask and ground-truth visibility mask. The AR is reported on five core datasets of the BOP Benchmark.
    Right: Percentage of target objects as a function of different visible surface thresholds. The visible surface is computed as the IoU between the ground-truth mask and the ground-truth visibility mask. The targets percentage is reported on five core datasets of the BOP Benchmark.
    }
    \label{fig:visib}
\end{figure}

The BOP Benchmark~\cite{sundermeyer2023bop} includes challenging occlusion scenarios.
We assess how this aspect affects \ourmethod's performance. 
In Fig.~\ref{fig:visib}(left), we compute the AR on LM-O~\cite{lmo}, T-LESS~\cite{tless}, TUD-L~\cite{tudl}, IC-BIN~\cite{icbin}, and YCB-V~\cite{ycbv} considering only target objects with visible surface greater than a given threshold.
More precisely, we define the visible surface threshold as the minimum value below which we no longer consider the target in the computation of the AR. 
In Fig.~\ref{fig:visib}(right) we also show the number of valid targets as percentage over the whole dataset for different visible surface thresholds. 
Given a target object in an image, we compute its visible surface as the Intersection over Union (IoU) of its ground-truth mask and visibility mask. 
A visible surface equal to 100\% means the target is totally visible, while a visible surface equal to 0\% means the target it totally occluded.
As the minimum visible surface increases so does the AR, however, the performances on the datasets exhibit different trends.
In TUD-L, occlusions are less relevant, i.e.~nearly all targets have a visible surface larger than 80\%. 
LM-O and IC-BIN are the most challenging datasets due to occlusions; only about 60\% and 40\% of the targets, respectively, have a visible surface larger than 80\% (see Fig.~\ref{fig:visib}(right)). Across all the datasets \ourmethod has lower AR on IC-BIN and T-LESS.
We can deduce that the low AR on IC-BIN is due its severe occlusions.  \ourmethod's performance on T-LESS, instead, is less related to occlusions since it presents additional challenges as the objects lack informative textures.

\section{Analysis of computational time}\label{sec:supp_timings}

\begin{table}[t]
\centering
\tabcolsep 10pt
\caption{
Comparison of computational times on TUD-L~\cite{tudl}.
We compared \ourmethod against MegaPose~\cite{labbe2022megapose} and the concurrent work SAM6D~\cite{lin2023sam}.
To ensure a fair comparison, we reproduced all experiments on the same hardware.
}
\label{tab:computational}

\vspace{-3mm}
\resizebox{0.8\columnwidth}{!}{%
\begin{tabular}{lcccccc}
    \toprule
    \multirow{2}{*}{Method} & \multicolumn{2}{c}{Input data} & \multicolumn{2}{c}{Localization prior} & \multirow{2}{*}{AR} & \multirow{2}{*}{Time (s)} \\
    & RGB & D & CNOS & SAM6D & & \\
    \midrule
    MegaPose~\cite{labbe2022megapose} & \checkmark & &  \checkmark & & 65.9 & 4.59 \\
    MegaPose~\cite{labbe2022megapose} & \checkmark & & & \checkmark & 68.4 & 6.06 \\
    SAM6D~\cite{lin2023sam} & \checkmark & \checkmark & \checkmark & & 82.0 & 0.98 \\
    SAM6D~\cite{lin2023sam} & \checkmark & \checkmark & & \checkmark & 90.3 & 3.94 \\
    \ourmethod (ours) & \checkmark & \checkmark & \checkmark & & 93.6 & 2.72 \\
    \ourmethod (ours) & \checkmark & \checkmark & & \checkmark & 94.9 & 4.79 \\
    \bottomrule
\end{tabular}
}
\end{table}

In Tab.~\ref{tab:computational}, we analyze the average execution time of \ourmethod on TUD-L~\cite{tudl}.
Our experiments are conducted using a NVIDIA Tesla A40 GPU and an Intel(R) Xeon(R) Silver 4316 CPU operating at 2.30GHz, utilizing 16 cores. 
We also reproduced the experiments of MegaPose~\cite{labbe2022megapose}\footnote{\href{https://github.com/agimus-project/happypose}{github.com/agimus-project/happypose}, accessed Mar.~2024.} and SAM6D~\cite{lin2023sam}\footnote{\href{https://github.com/JiehongLin/SAM-6D}{github.com/JiehongLin/SAM-6D}, accessed Mar. 2024} on our hardware to ensure a fair comparison of the computational times. For a given 3D model and an image, \ourmethod estimates the pose in 2.72 seconds using CNOS~\cite{cnos} masks. When using SAM6D masks, \ourmethod takes 4.79 seconds. In both cases the timing for predicting the masks are included.
Compared to MegaPose, using the same CNOS mask priors, \ourmethod achieves a significant speedup of 41\% and a gain of +27.7 in terms of Average Recall (AR).
When using SAM6D masks, \ourmethod still achieves a speedup of about 21\% and a gain of +26.5 in AR.
It is important to note that the tested version of MegaPose relies solely on RGB images, while \ourmethod also incorporates depth information, surpassing MegaPose in both accuracy and speed.
\ourmethod achieves significantly higher accuracy (+4.6 AR) than SAM6D when utilizing SAM6D masks, despite being 0.85 seconds slower.
When employing CNOS masks, \ourmethod surpasses SAM6D by +11.6 AR, albeit being  1.78 seconds slower.
Finally, \ourmethod with CNOS masks surpasses SAM6D with its own masks, gaining a +3.3 AR, and being 1.22 seconds faster.

%%%%%%%%%%%%%%%%%%%%%%%%%%%%%%%%%%%%%%%%%%%%%%%%%%%%%%%%%%%%%%%%%%%%%%%
%%%%%%%%%%%%%%%%%%%%%%%%%%%%%%%%%%%%%%%%%%%%%%%%%%%%%%%%%%%%%%%%%%%%%%%
\section{Additional qualitative results}\label{sec:supp_bop_qual}

We present qualitative results for each of the seven core datasets of the BOP Benchmark~\cite{sundermeyer2023bop}.
Each figure has the same structure.
Rows show the methods: MegaPose~\cite{labbe2022megapose} (second), SAM6D~\cite{lin2023sam} (third) and \ourmethod (fourth). 
Columns show different examples.
To aid readers in analyzing the results, incorrect predictions are highlighted with {\color{red}\textbf{red}} arrows, and missing predictions are emphasized using {\color{Dandelion}\textbf{yellow}} arrows.
Additionally, for better contrast against the RGB-colored objects, the backgrounds in the second to fifth rows are converted to grayscale.
Across the seven datasets we highlight cases where \ourmethod performs equivalently, surpasses, or falls short compared to the other methods.
All the selected methods use masks estimated by CNOS~\cite{cnos}.

\begin{figure*}[t]
\centering

    \raggedright
    \begin{minipage}{0.32\textwidth}
        \centering
        \footnotesize (a)
    \end{minipage}
    \begin{minipage}{0.32\textwidth}
        \centering
        \footnotesize (b)
    \end{minipage}
    \begin{minipage}{0.32\textwidth}
        \centering
        \footnotesize (c)
    \end{minipage}

    \vspace{1.3 mm}
    \begin{tabular}{@{}c@{\,}c@{\,}c}
    \raggedright

        % first row
        \begin{overpic}[width=0.33\textwidth]{supp/images/qualitatives/lmo/000310_input.png}
            \put(-10, 14){\rotatebox{90}{\footnotesize Input images}}
        \end{overpic} &
        \begin{overpic}[width=0.33\textwidth]{supp/images/qualitatives/lmo/000956_input.png}
        \end{overpic} &
        \begin{overpic}[width=0.33\textwidth]{supp/images/qualitatives/lmo/001123_input.png}
        \end{overpic} \\

        % \put(120,200){\vector(0,-1){150}}

        % second row

        \begin{overpic}[width=0.33\textwidth]{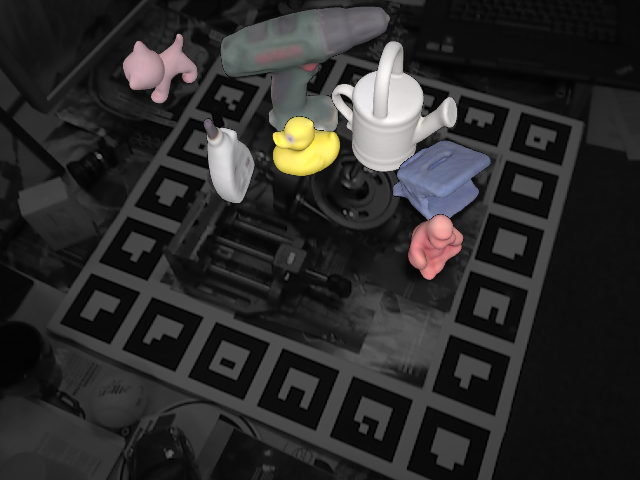}
            \put(-10, 20){\rotatebox{90}{\footnotesize MegaPose}}
            \linethickness{2.1pt}
            \put(20, 60){\color{red}\vector(1.4, -0.7){12}}
        \end{overpic} &
        \begin{overpic}[width=0.33\textwidth]{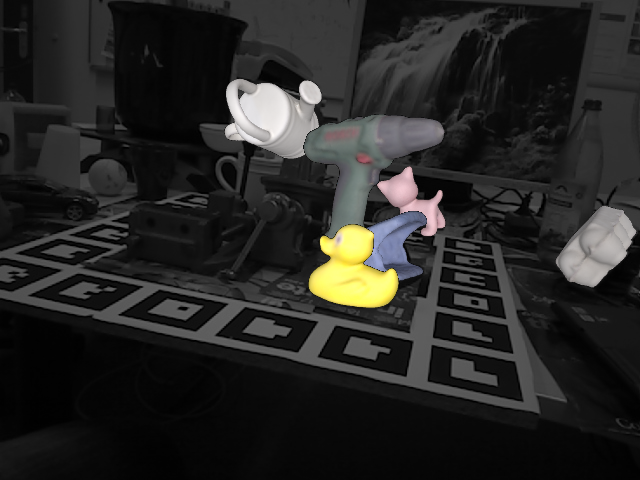}
            \linethickness{2.1pt}
            \put(22, 70){\color{red}\vector(1.4, -1){12}}
        \end{overpic} &
        \begin{overpic}[width=0.33\textwidth]{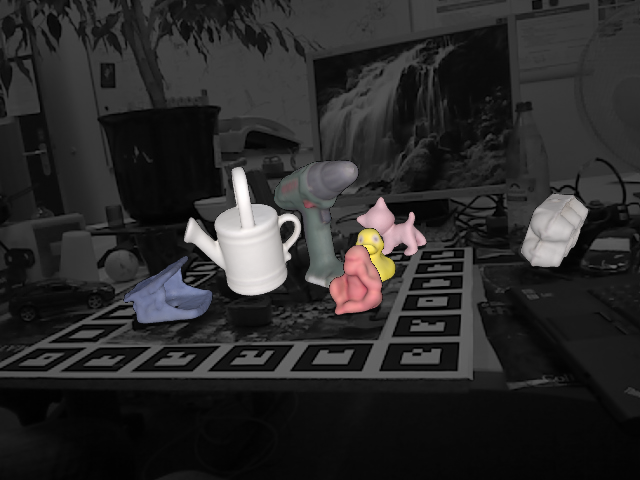}
            \linethickness{2.1pt}
            \put(22, 50){\color{red}\vector(1.4, -1){12}}
            \put(44, 16){\color{red}\vector(-1.4, 1){12}}
        \end{overpic} \\

%%% third row

        \begin{overpic}[width=0.33\textwidth]{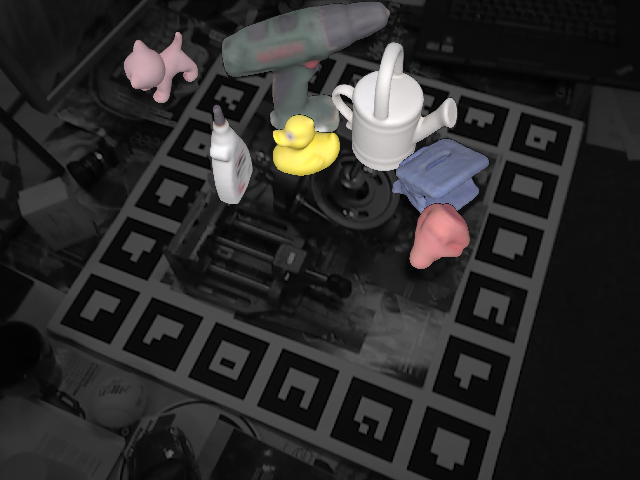}
            \put(-10, 23){\rotatebox{90}{\footnotesize SAM6D}}
            \linethickness{2.1pt}
            \put(84, 25){\color{red}\vector(-1.4, 1){12}}
        \end{overpic} &
        \begin{overpic}[width=0.33\textwidth]{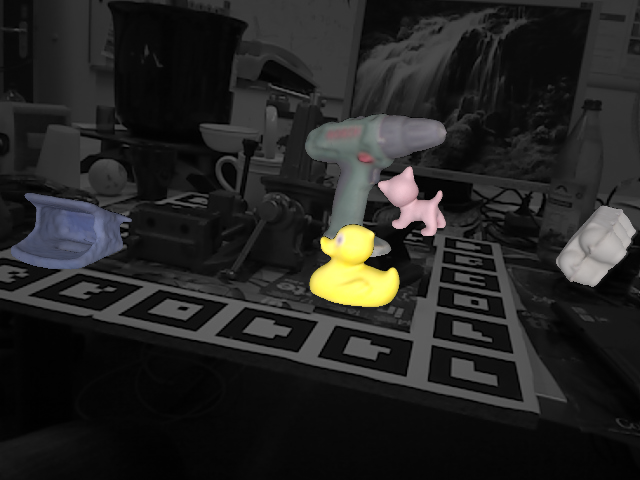}
            \linethickness{2.1pt}
            \put(30, 25){\color{red}\vector(-1.4, 1){12}}
            \linethickness{2.1pt}
            \put(22, 60){\color{Goldenrod}\vector(1.4, -1){12}}
        \end{overpic} &
        \begin{overpic}[width=0.33\textwidth]{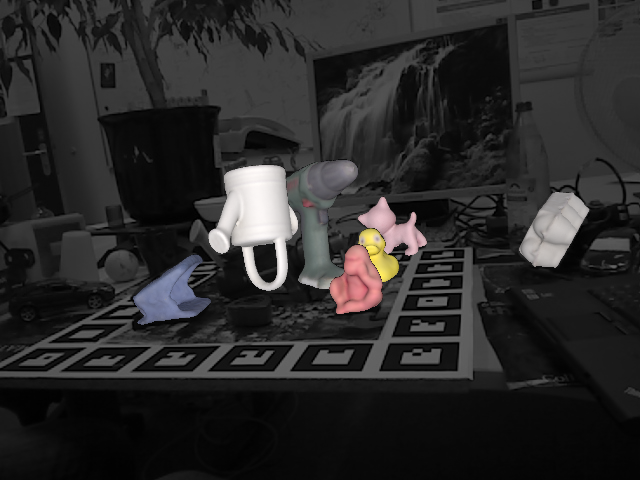}
            \linethickness{2.1pt}
            \put(22, 57){\color{red}\vector(1.4, -1){12}}
        \end{overpic} \\

        \begin{overpic}[width=0.33\textwidth]{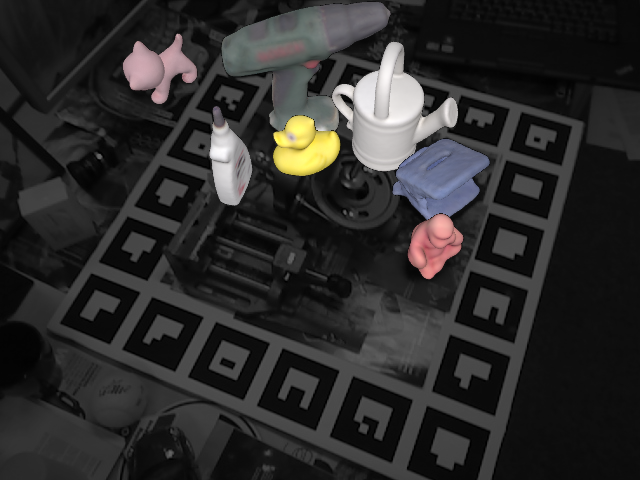}
            \put(-10, 25){\rotatebox{90}{\footnotesize \ourmethod}}
        \end{overpic} &
        \begin{overpic}[width=0.33\textwidth]{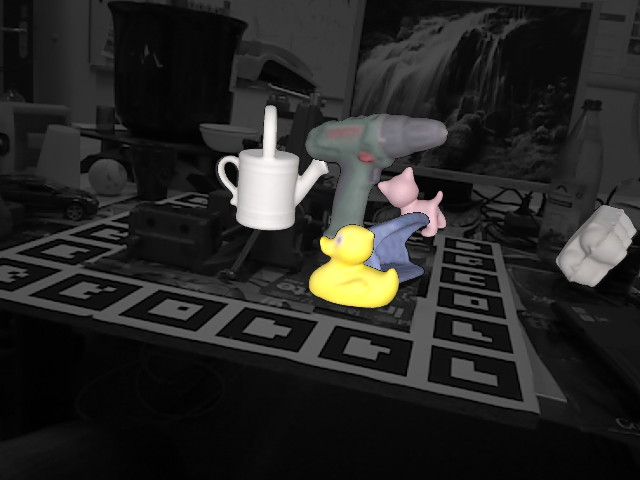}
        \end{overpic} &
        \begin{overpic}[width=0.33\textwidth]{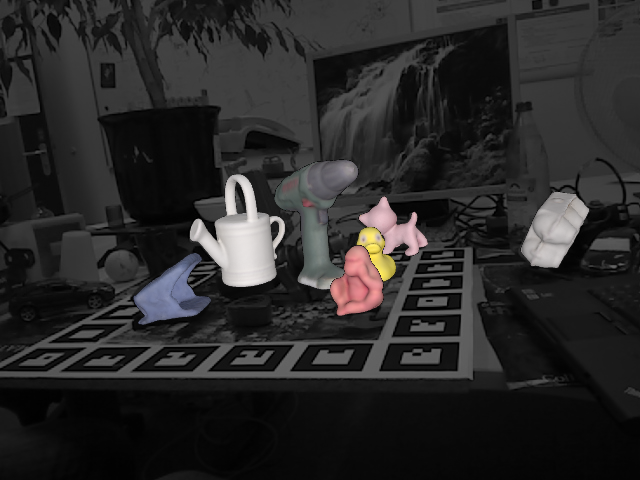}
        \end{overpic} \\

    \end{tabular}
    \vspace{-1mm}
    \caption{
        Qualitative results on LM-O~\cite{lmo}.
        Columns show different examples.
        Rows show a comparison against different methods.
        {\color{red}\textbf{Red}} ({\color{Dandelion}\textbf{yellow}}) arrows highlight wrong (missing) predictions.
        Backgrounds are converted to grayscale for a better contrast.
    }

    \label{fig:supp_qual_lmo}
\end{figure*}

%%%%%%%%%%%%%%%%%%%%%%%%%%%%%%%%%%%%%%%%%%%%%%%%%%%%%%%%%%%%%%%%%%%%%%%
\noindent\textbf{LM-O dataset.}~Fig.~\ref{fig:supp_qual_lmo} shows qualitative results on LM-O~\cite{lmo}.
In column (a), MegaPose predictions for the white-black glue flip the pose by 180 degrees. This discrepancy is noticeable either by examining the shape portion highlighted with the red arrow, or by observing the mismatched side texture compared to the input image. The challenging aspect of this object lies in its near-symmetry: it has minimal differences in shape, and a subtle brand presence in the front texture, absent in the back texture. SAM6D correctly estimates the glue's pose, however, it sharply fails to predict the red primate's pose.
In column (b), MegaPose prediction for the white watering can is quite inaccurate (red arrow). SAM6D does not predict any pose for the same object (yellow arrow). This scene is challenging due to heavy occlusion of the watering can's sprinkler.
Additionally, SAM6D fails to predict the correct pose for the blue hole-puncher.
In column (c), the white watering can's pose is incorrectly predicted by both MegaPose (less severe error) and SAM6D (upside-down), most likely because its handle is occluded. The pose for the blue hole-puncher is incorrectly predicted by MegaPose, presenting the object upside-down (evidenced by the red arrow tip pointing towards the top of the hole-puncher), despite the object being un-occluded.
\ourmethod consistently identifies the correct pose for all objects in all cases.

\begin{figure*}[t]
\centering

    \raggedright
    \begin{minipage}{0.32\textwidth}
        \centering
        \footnotesize (a)
    \end{minipage}
    \begin{minipage}{0.32\textwidth}
        \centering
        \footnotesize (b)
    \end{minipage}
    \begin{minipage}{0.32\textwidth}
        \centering
        \footnotesize (c)
    \end{minipage}

   \vspace{1.3 mm}
    \begin{tabular}{@{}c@{\,}c@{\,}c}
    \raggedright

        \begin{overpic}[width=0.33\textwidth]{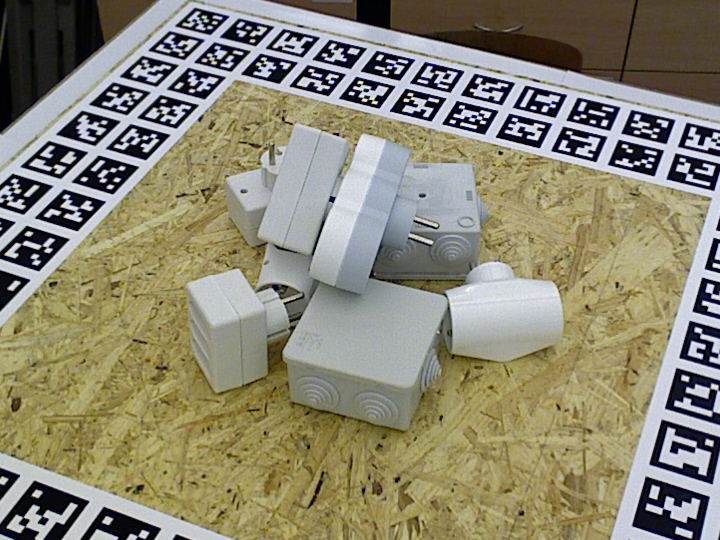}
            \put(-10, 14){\rotatebox{90}{\footnotesize Input images}}
        \end{overpic} &
        \begin{overpic}[width=0.33\textwidth]{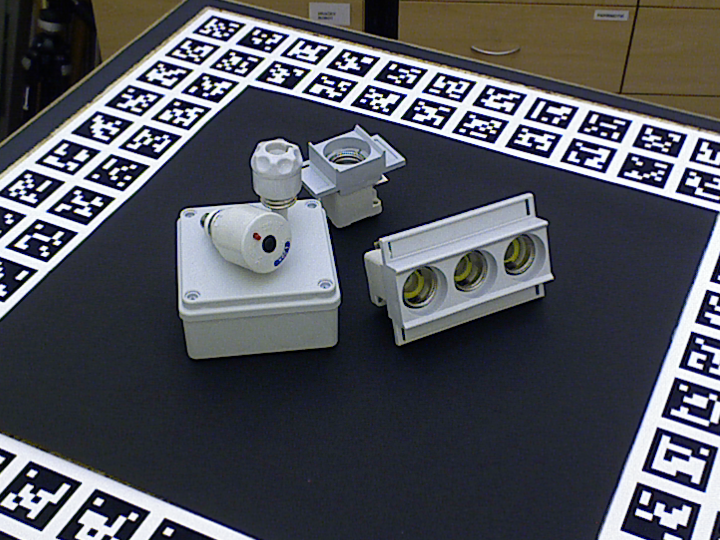}
        \end{overpic} &
        \begin{overpic}[width=0.33\textwidth]{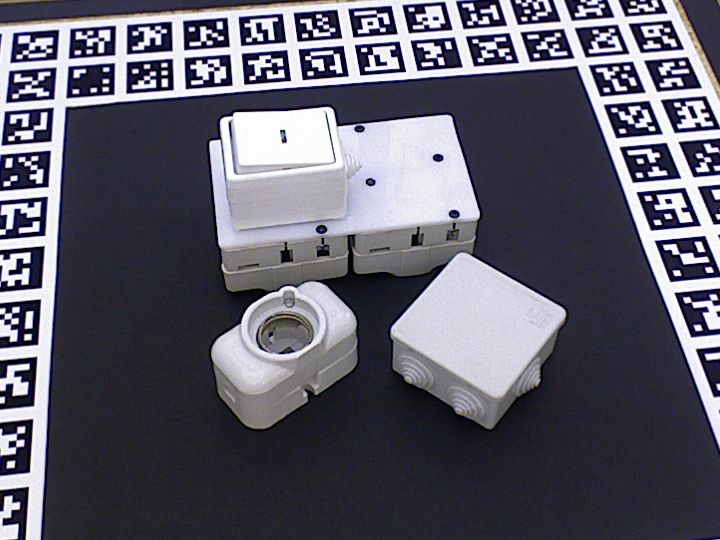}
        \end{overpic} \\

        \begin{overpic}[width=0.33\textwidth]{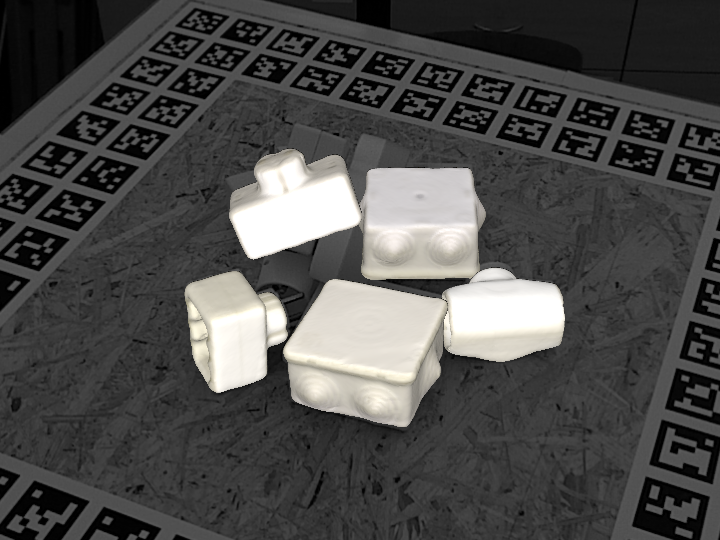}
            \put(-10, 20){\rotatebox{90}{\footnotesize MegaPose}}
            \linethickness{2.1pt}
            \put(67, 64){\color{Goldenrod}\vector(-1.4, -1){12}}
            \put(55, 66){\color{Goldenrod}\vector(-1.4, -1){12}}
            \put(19, 58){\color{red}\vector(1.4, -1){12}}
        \end{overpic} &
        \begin{overpic}[width=0.33\textwidth]{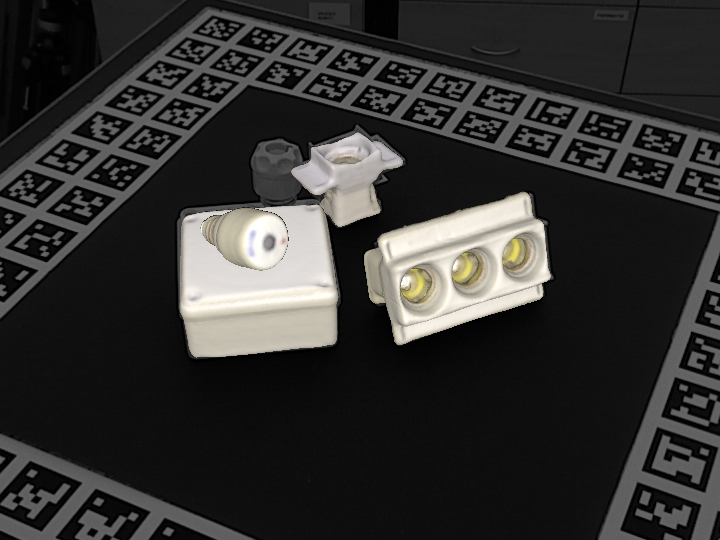}
            \linethickness{2.1pt}
            \put(22, 60){\color{Goldenrod}\vector(1.4, -0.7){12}}
        \end{overpic} &
        \begin{overpic}[width=0.33\textwidth]{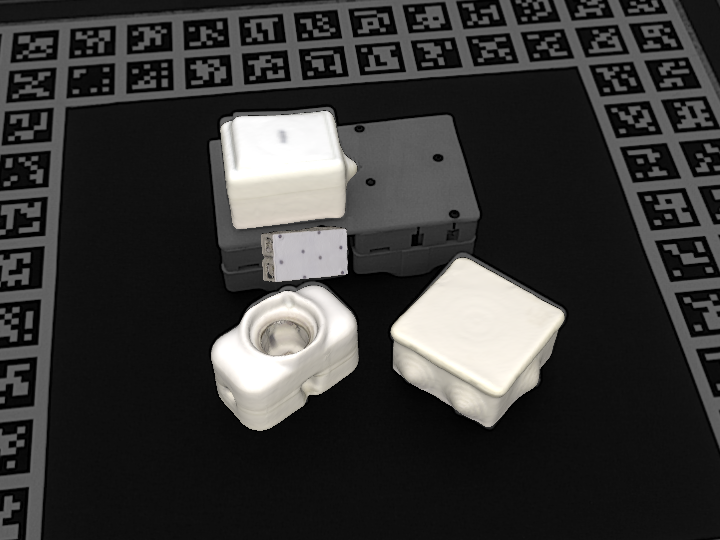}
            \linethickness{2.1pt}
            \put(62, 48){\color{red}\vector(-1.4, -0.7){12}}
        \end{overpic} \\

%%%%%%%%%%%%%%%%%%%%%%%%%%%%%%%%%%%%%%%%%%%%

        \begin{overpic}[width=0.33\textwidth]{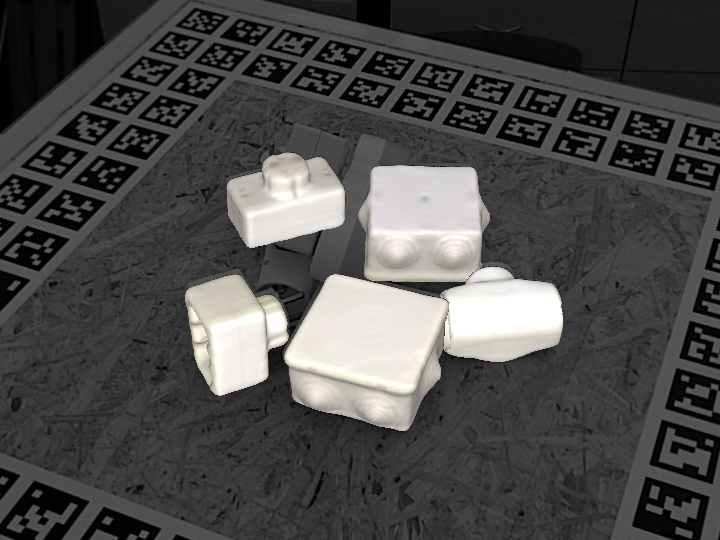}
            \put(-10, 23){\rotatebox{90}{\footnotesize SAM6D}}
            \linethickness{2.1pt}
            \put(67, 64){\color{Goldenrod}\vector(-1.4, -1){12}}
            \put(55, 66){\color{Goldenrod}\vector(-1.4, -1){12}}
        \end{overpic} &
        \begin{overpic}[width=0.33\textwidth]{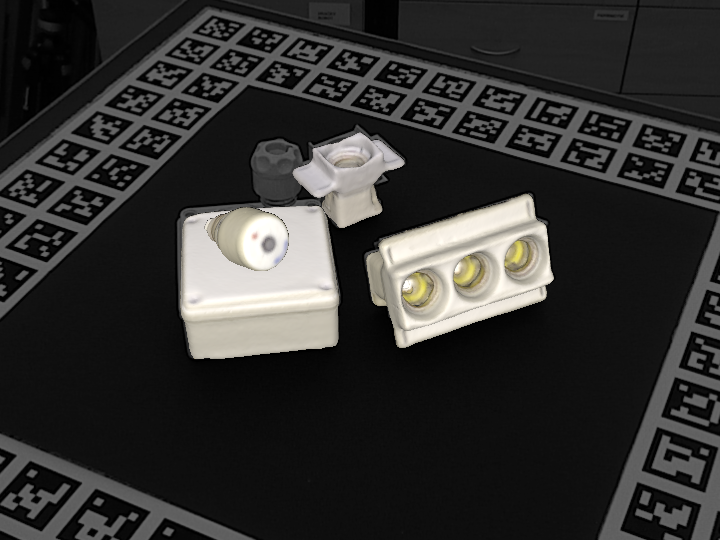}
            \linethickness{2.1pt}
            \put(22, 60){\color{Goldenrod}\vector(1.4, -0.7){12}}
        \end{overpic} &
        \begin{overpic}[width=0.33\textwidth]{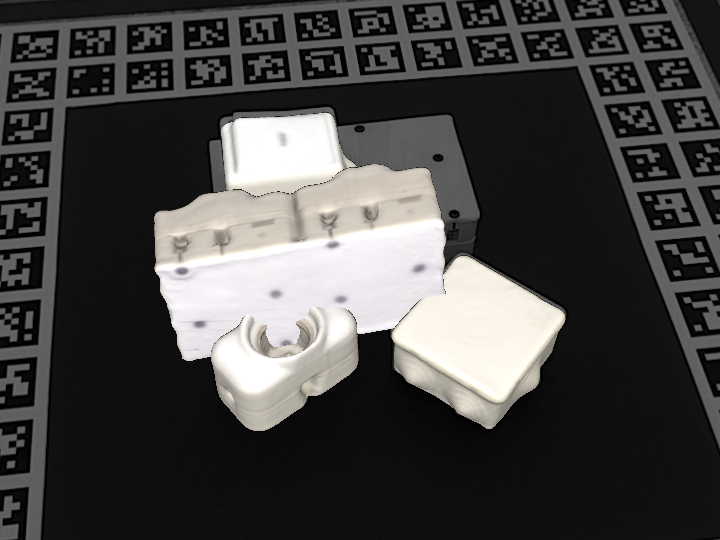}
            \linethickness{2.1pt}
            \put(9, 55){\color{red}\vector(1.4, -1){12}}
        \end{overpic} \\

        \begin{overpic}[width=0.33\textwidth]{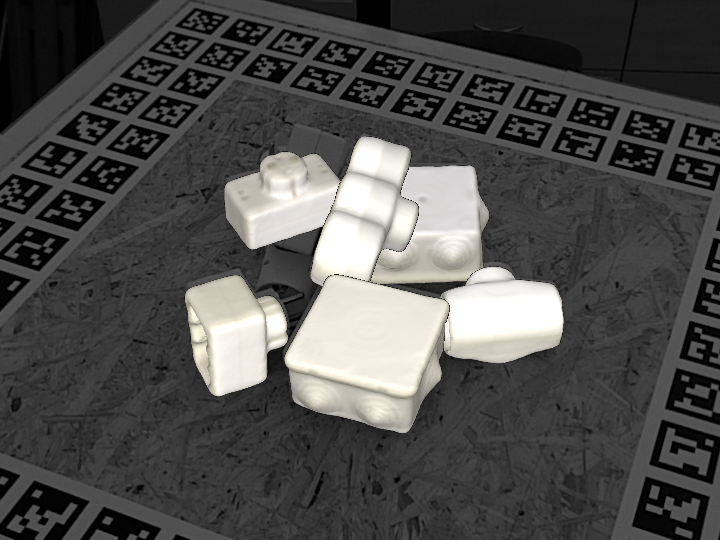}
            \put(-10, 25){\rotatebox{90}{\footnotesize \ourmethod}}
            \linethickness{2.1pt}
            \put(55, 66){\color{Goldenrod}\vector(-1.4, -1){12}}
        \end{overpic} &
        \begin{overpic}[width=0.33\textwidth]{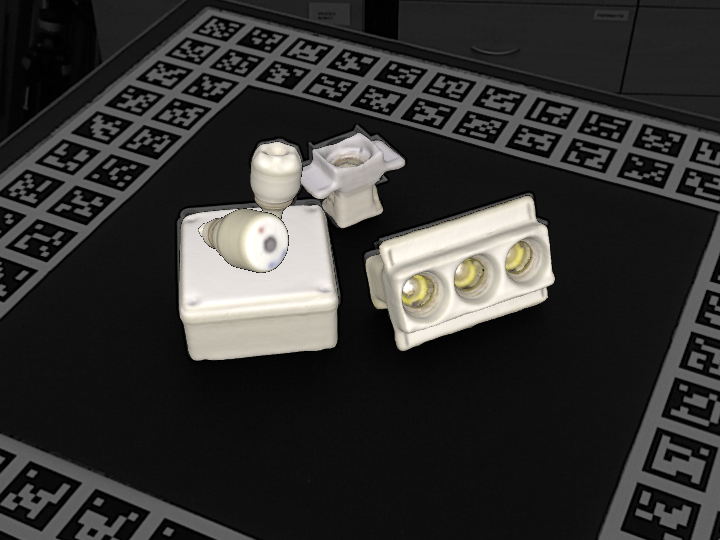}
        \end{overpic} &
        \begin{overpic}[width=0.33\textwidth]{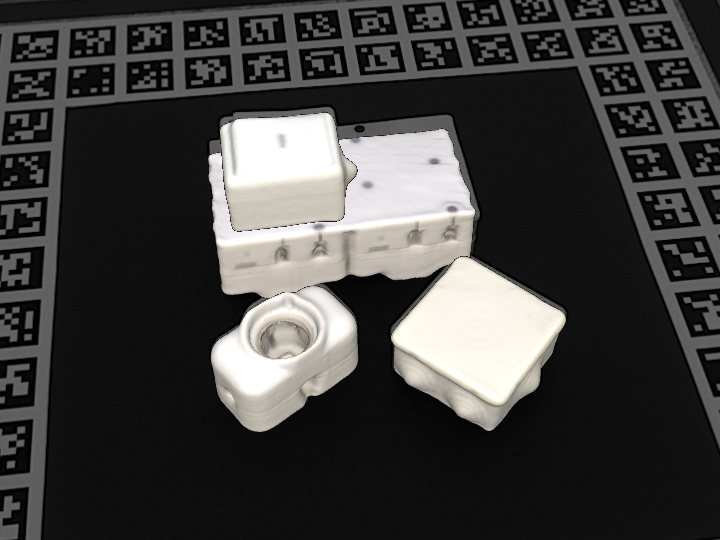}
        \end{overpic} \\

    \end{tabular}
    \vspace{-1mm}
    \caption{
        Qualitative results on T-LESS~\cite{tless}.
        Columns show different examples.
        Rows show a comparison against different methods.
        {\color{red}\textbf{Red}} ({\color{Dandelion}\textbf{yellow}}) arrows highlight wrong (missing) predictions.
        Backgrounds are converted to grayscale for a better contrast.
    }
    \label{fig:supp_qual_tless}
\end{figure*}

%%%%%%%%%%%%%%%%%%%%%%%%%%%%%%%%%%%%%%%%%%%%%%%%%%%%%%%%%%%%%%%%%%%%%%%
\noindent\textbf{T-LESS dataset.}~Fig.~\ref{fig:supp_qual_tless} shows qualitative results on T-LESS~\cite{tless}.
In column (a), no methods can correctly predict all the object poses, with \ourmethod only missing a single object pose (yellow arrow).
In column (b), all methods but \ourmethod miss one of the objects (yellow arrows), possibly because the instance localization priors provided by CNOS are too wide and include both background and other objects.
In column (c), all methods except \ourmethod inaccurately predict the pose of the largest item.
SAM6D predicts the orientation of the electrical item as if it was lying on its side instead of its bottom, whereas MegaPose's prediction of the translation component notably deviates (the scale of the object is significantly different from the correct one).
These inaccuracies likely result from the low quality of the instance localization prior, which in this case covers only a tiny portion of the object.
Despite using the same localization priors, \ourmethod predicts the correct pose for all objects in (b,c).

\begin{figure*}[t]
\centering

    \begin{minipage}{0.32\textwidth}
        \centering
        \footnotesize (a)
    \end{minipage}
    \begin{minipage}{0.32\textwidth}
        \centering
        \footnotesize (b)
    \end{minipage}
    \begin{minipage}{0.32\textwidth}
        \centering
        \footnotesize (c)
    \end{minipage}

    \vspace{1.3 mm}
    \begin{tabular}{@{}c@{\,}c@{\,}c}
    \raggedright

        % first row
        \begin{overpic}[width=0.33\textwidth]{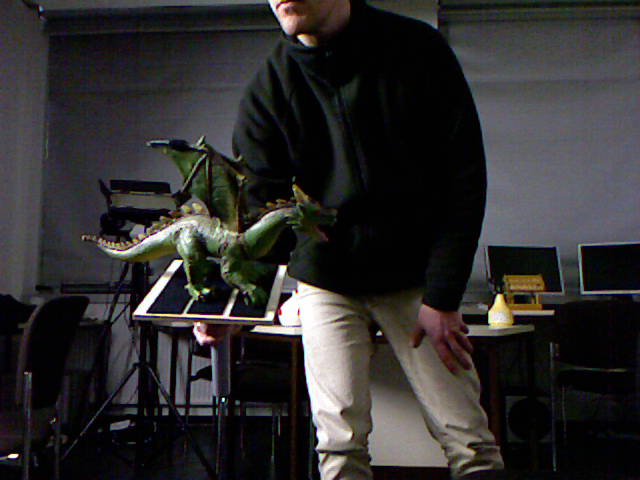}
            \put(-10, 14){\rotatebox{90}{\footnotesize Input images}}
        \end{overpic} &
        \begin{overpic}[width=0.33\textwidth]{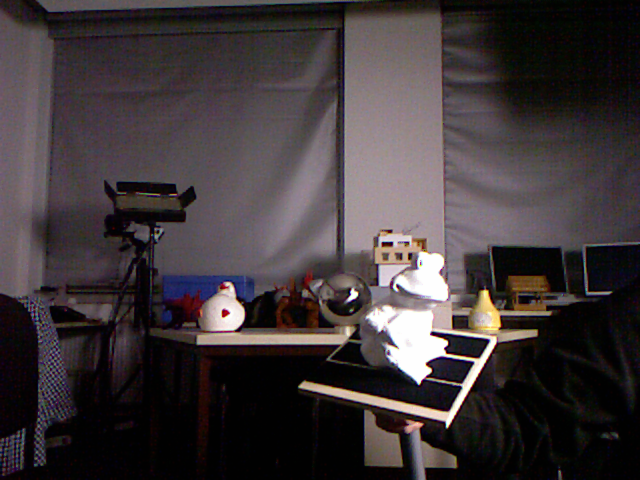}
        \end{overpic} &
        \begin{overpic}[width=0.33\textwidth]{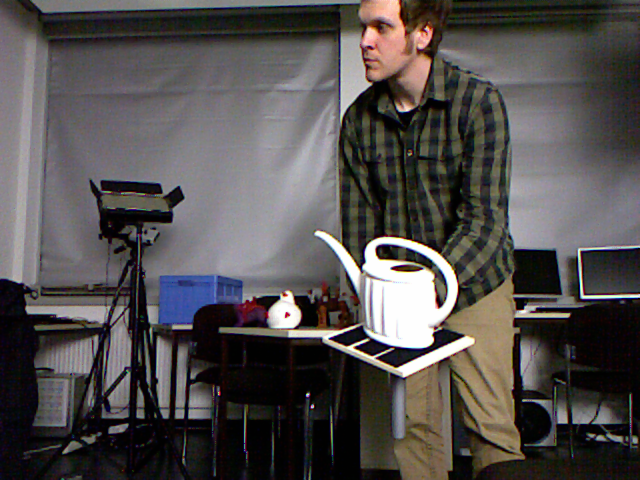}
        \end{overpic} \\

        % second row

        \begin{overpic}[width=0.33\textwidth]{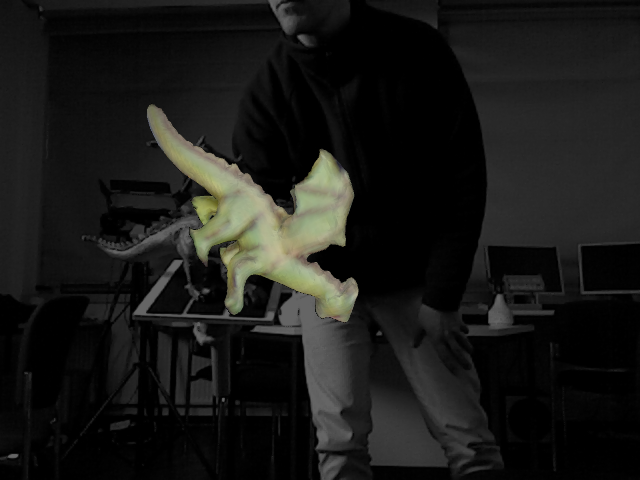}
            \linethickness{2.1pt}
            \put(68, 39){\color{red}\vector(-1.4, -1){12}}
            \put(-10, 20){\rotatebox{90}{\footnotesize MegaPose}}
            % \linethickness{2pt}
            % \put(69,39){\color{red}\vector(-1.4,-1){12}}
        \end{overpic} &
        \begin{overpic}[width=0.33\textwidth]{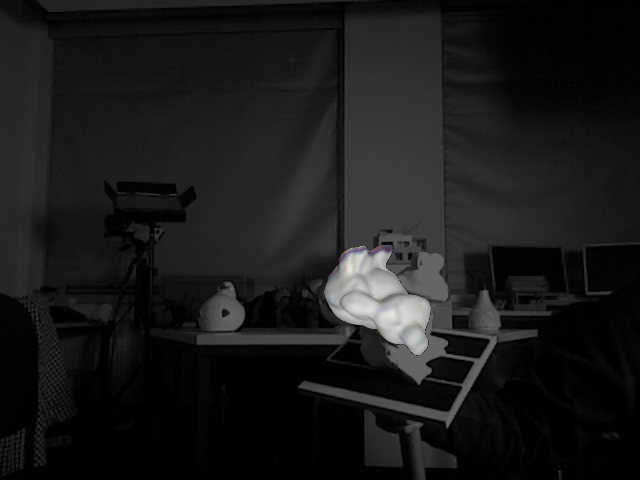}
            \linethickness{2.1pt}
            \put(38,39){\color{red}\vector(1.4,-0.8){12}}
        \end{overpic} &
        \begin{overpic}[width=0.33\textwidth]{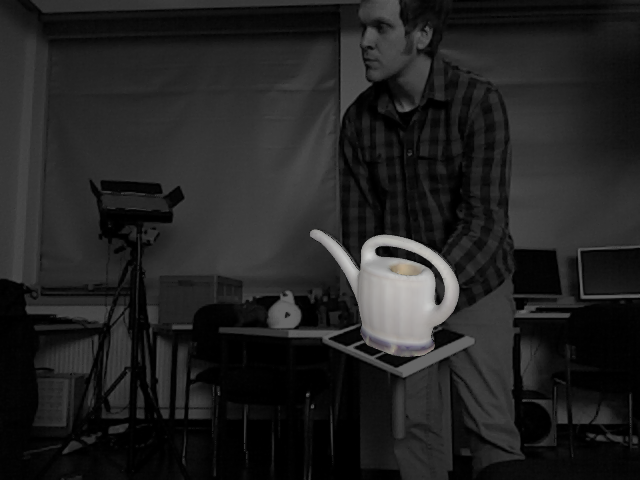}
        \end{overpic} \\
        
   % third row
   
        \begin{overpic}[width=0.33\textwidth]{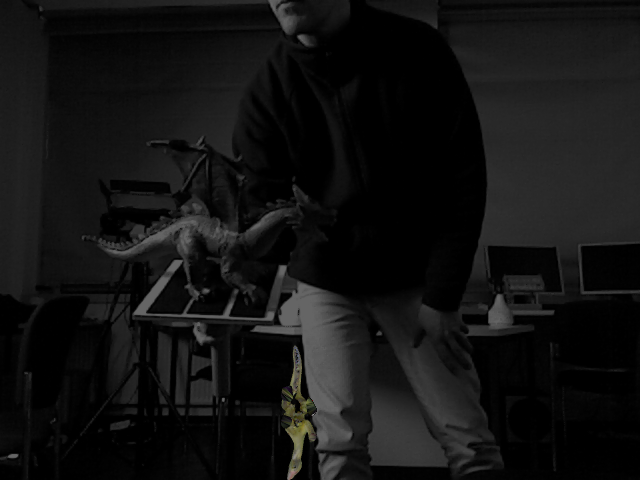}
        \linethickness{2.1pt}
        \put(62, 20){\color{red}\vector(-1.4, -1){12}}
            \put(-10, 23){\rotatebox{90}{\footnotesize SAM6D}}
            % \linethickness{2pt}
            % \put(62,23){\color{red}\vector(-1.4,-1){12}}
        \end{overpic} &
        \begin{overpic}[width=0.33\textwidth]{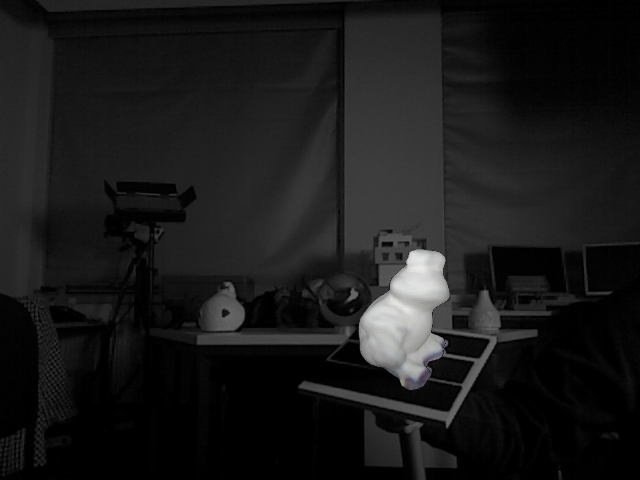}
        \end{overpic} &
        \begin{overpic}[width=0.33\textwidth]{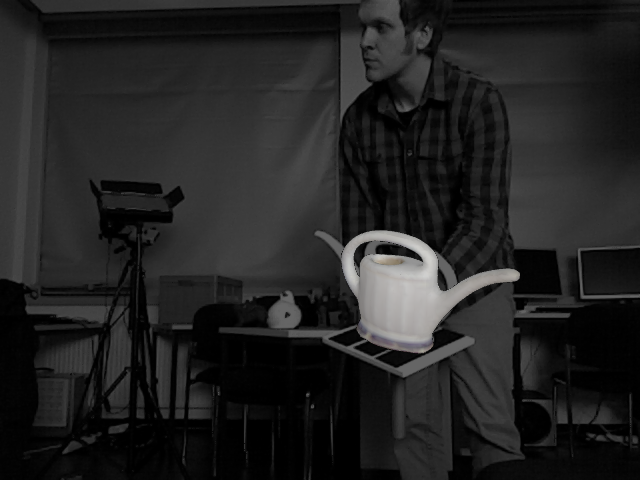}
            \linethickness{2.1pt}
            \put(94,42){\color{red}\vector(-1.4,-1){12}}
        \end{overpic} \\

        % fifth row
        \begin{overpic}[width=0.33\textwidth]{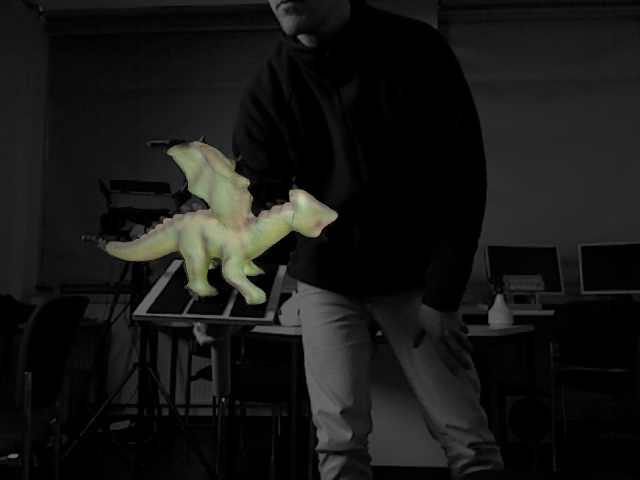}
            \put(-10, 25){\rotatebox{90}{\footnotesize \ourmethod}}
        \end{overpic} &
        \begin{overpic}[width=0.33\textwidth]{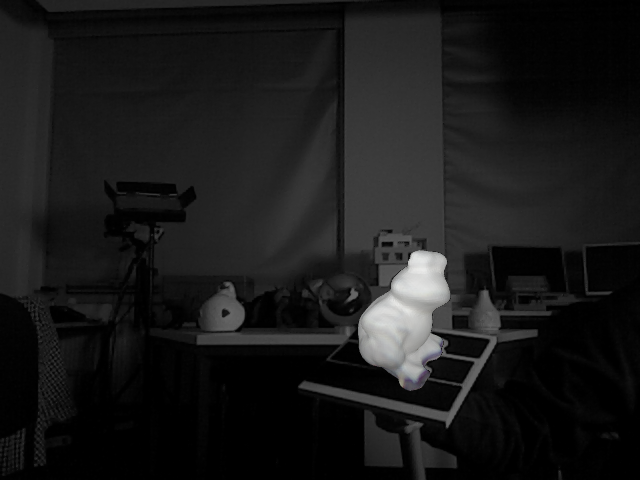}
        \end{overpic} &
        \begin{overpic}[width=0.33\textwidth]{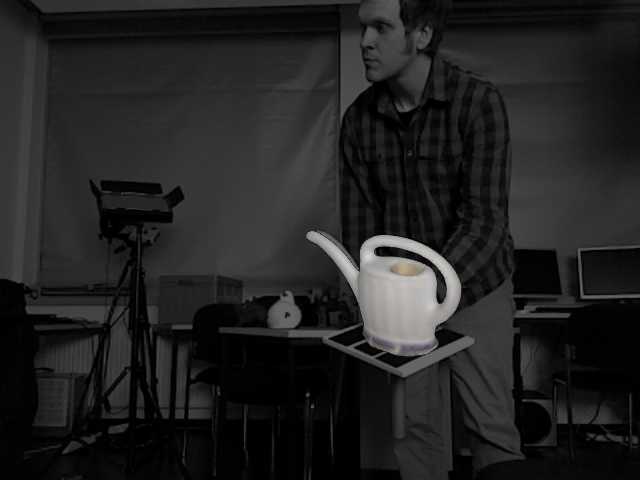}
            % \linethickness{2pt}
            % \put(65,47){\color{red}\vector(-1.4,-1){12}}
        \end{overpic} \\
        
    \end{tabular}
    \vspace{-1mm}
    \caption{
        Qualitative results on the TUD-L~\cite{tudl} dataset.
        Rows show a comparison against different methods.
        Columns show different examples.
        {\color{red}\textbf{Red}} arrows highlight wrong predictions.
        Backgrounds are converted to grayscale for a better contrast.
    }
    \label{fig:supp_qual_tudl}
\end{figure*}

%%%%%%%%%%%%%%%%%%%%%%%%%%%%%%%%%%%%%%%%%%%%%%%%%%%%%%%%%%%%%%%%%%%%%%%
\noindent\textbf{TUD-L dataset.}~Fig.~\ref{fig:supp_qual_tudl} shows qualitative results on TUD-L~\cite{tudl}.
It contains three different objects: a toy model of a green-brown dragon, a toy model of a white frog, and a white watering can.
In column (a), \ourmethod estimates are accurate, while both MegaPose and SAM6D struggle to accurately estimate the dragon's pose.
This issue arises because the CNOS localization prior is too wide, including also the person holding the object.
This can be evinced by noticing that SAM6D's prediction is close to the person's leg (red arrow) rather than their hand.
In column (b), \ourmethod outperforms MegaPose, despite the associated localization prior only covering the frog's head.
In column (c), \ourmethod performs slightly worse than MegaPose. Our predicted pose leads to a moderate misalignment of the watering can's spout. This discrepancy occurs because the localization prior lacks coverage of the watering can's spout portion.

\begin{figure*}[t]
\centering

    \begin{minipage}{0.32\textwidth}
        \centering
        \footnotesize (a)
    \end{minipage}
    \begin{minipage}{0.32\textwidth}
        \centering
        \footnotesize (b)
    \end{minipage}
    \begin{minipage}{0.32\textwidth}
        \centering
        \footnotesize (c)
    \end{minipage}

    \vspace{1.3 mm}
    \begin{tabular}{@{}c@{\,}c@{\,}c}
    \raggedright

        % first row
        \begin{overpic}[width=0.33\textwidth]
        {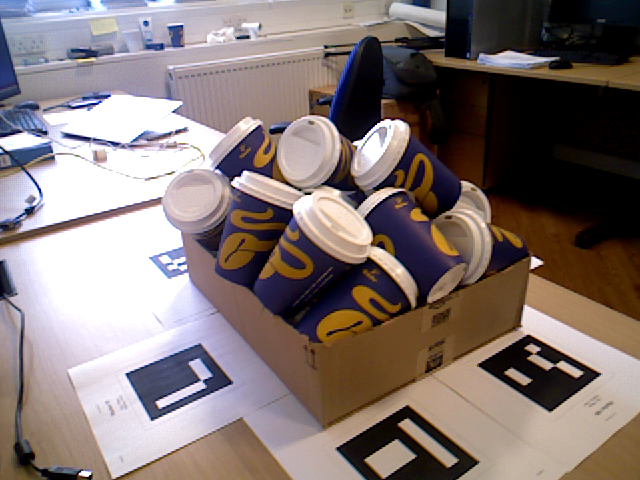}
            \put(-10, 14){\rotatebox{90}{\footnotesize Input images}}
        \end{overpic} &
        \begin{overpic}[width=0.33\textwidth]
        {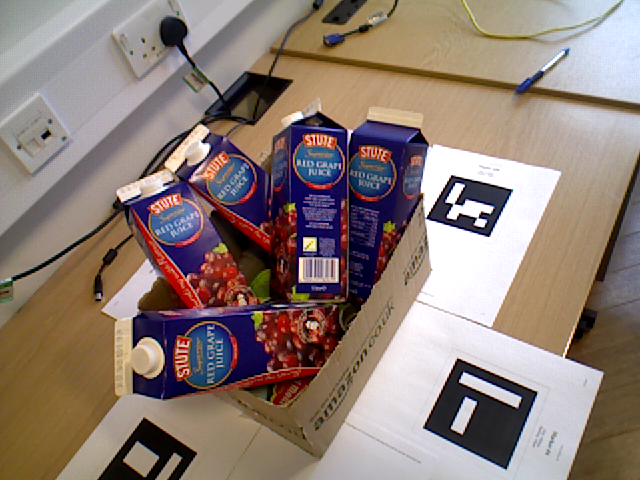}
        \end{overpic} &
        \begin{overpic}[width=0.33\textwidth]
        {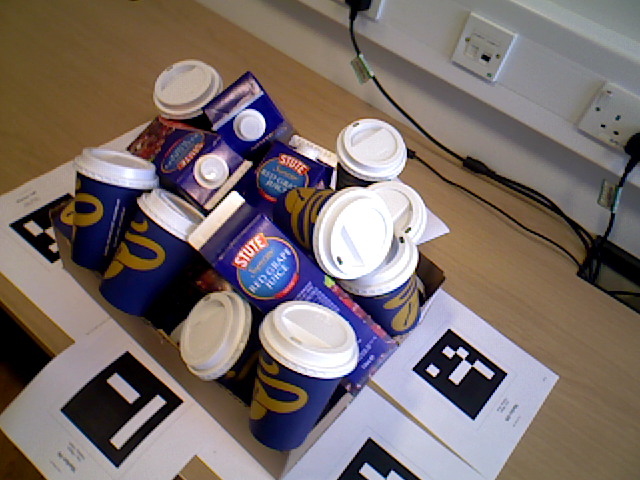}
        \end{overpic} \\

        % second row

        \begin{overpic}[width=0.33\textwidth]
        {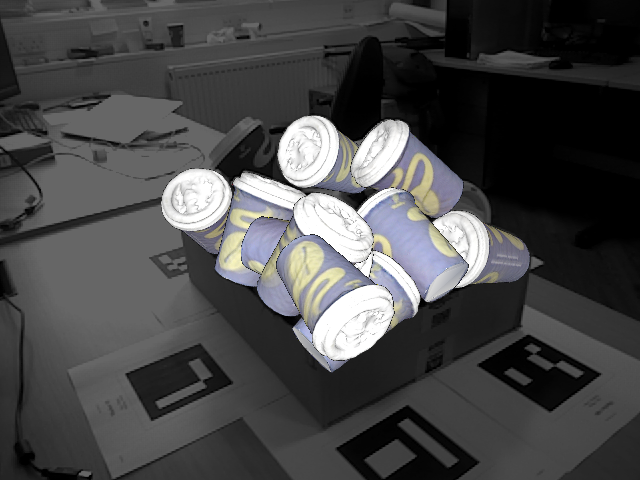}
            \linethickness{2.1pt}
            \put(23, 63){\color{Goldenrod}\vector(1.4, -1){12}}
            \put(88, 53){\color{Goldenrod}\vector(-1.4, -1){12}}
            \put(28, 25){\color{red}\vector(1.4, 1){12}}
            \put(36, 18){\color{red}\vector(1.4, 1){12}}
            \put(-10, 20){\rotatebox{90}{\footnotesize MegaPose}}
        \end{overpic} &
        \begin{overpic}[width=0.33\textwidth]
        {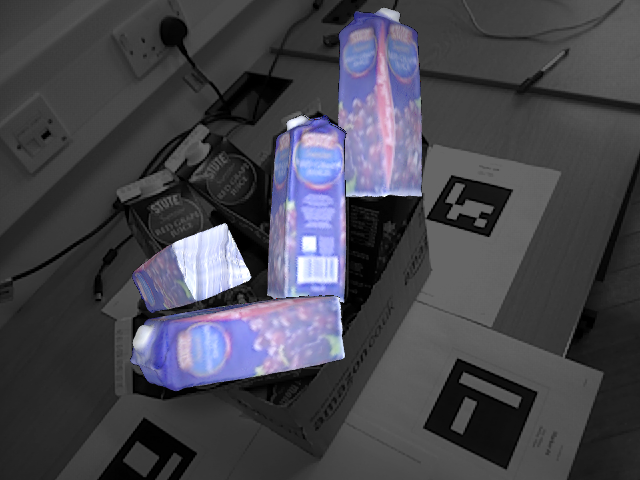}
        \linethickness{2.1pt}
        \put(10, 41){\color{red}\vector(1.4, -1){12}}
        \put(78, 53){\color{red}\vector(-1.4, 1){12}}
        \put(17, 61){\color{Goldenrod}\vector(1.4, -1){12}}
        \end{overpic} &
        \begin{overpic}[width=0.33\textwidth]
        {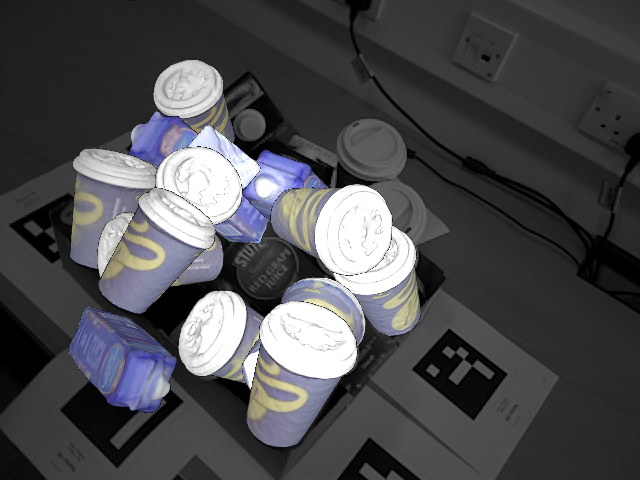}
        \linethickness{2.1pt}
        \put(9, 63){\color{red}\vector(1.4, -1){12}}
        \put(3, 7){\color{red}\vector(1.4, 1){12}}
        \put(66, 17){\color{red}\vector(-1.4, 1){12}}
        \put(55, 58){\color{red}\vector(-1.4, -1){12}}
        \put(3, 47){\color{red}\vector(1.4, -1){12}}
        \put(73, 63){\color{Goldenrod}\vector(-1.4, -1){12}}
        \put(77, 52){\color{Goldenrod}\vector(-1.4, -1){12}}
        \put(55, 67){\color{Goldenrod}\vector(-1.4, -1){12}}
        \end{overpic} \\

   % third row

        \begin{overpic}[width=0.33\textwidth]
        {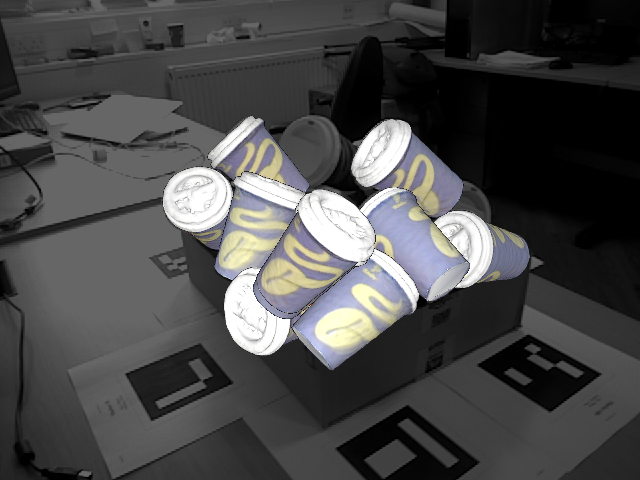}
        \linethickness{2.1pt}
        \put(64, 65){\color{Goldenrod}\vector(-1.4, -1){12}}
        \put(88, 53){\color{Goldenrod}\vector(-1.4, -1){12}}
        \put(24, 13){\color{red}\vector(1.4, 1){12}}
        \put(23, 63){\color{red}\vector(1.4, -1){12}}   
        %\put(30, 21){\color{red}\vector(1.4, 1){12}}    
        \put(60, 21){\color{red}\vector(-1.4, 1){12}}    
            \put(-10, 23){\rotatebox{90}{\footnotesize SAM6D}}
            \end{overpic} &
        \begin{overpic}[width=0.33\textwidth]
        {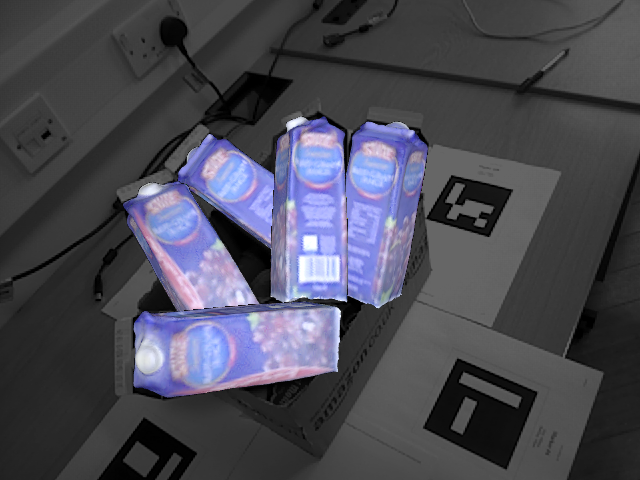}
        \linethickness{2.1pt}
        \put(16, 62){\color{red}\vector(1.4, -1){12}}
        \end{overpic} &
        \begin{overpic}[width=0.33\textwidth]
        {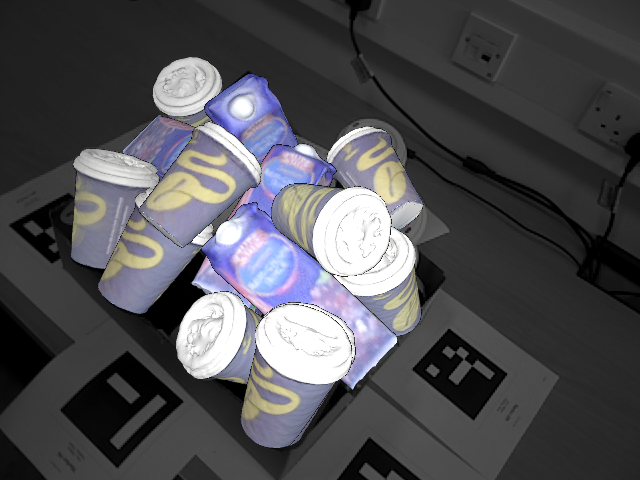}
        \linethickness{2.1pt}
        \put(16, 57){\color{red}\vector(1.4, -1){12}}
        \put(73, 63){\color{red}\vector(-1.4, -1){12}}
        \put(22, 28){\color{red}\vector(1.4, 1){12}}
        \put(77, 52){\color{Goldenrod}\vector(-1.4, -1){12}}
        \end{overpic} \\

        % fifth row
        \begin{overpic}[width=0.33\textwidth]
        {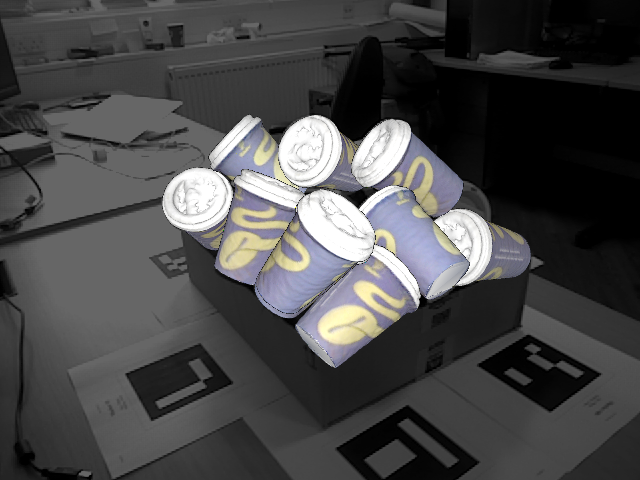}
            \linethickness{2.1pt}
            \put(-10, 25){\rotatebox{90}{\footnotesize \ourmethod}}
            \put(88, 53){\color{Goldenrod}\vector(-1.4, -1){12}}
            \put(65, 64){\color{red}\vector(-1.4, -1){12}}
        \end{overpic} &
        \begin{overpic}[width=0.33\textwidth]
        {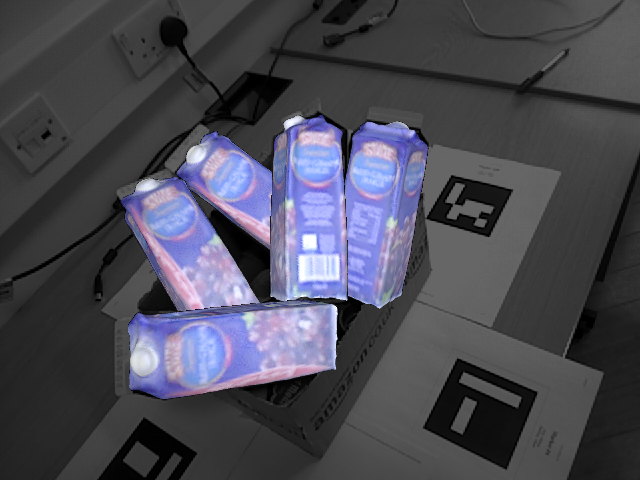}
        \end{overpic} &
        \begin{overpic}[width=0.33\textwidth]
        {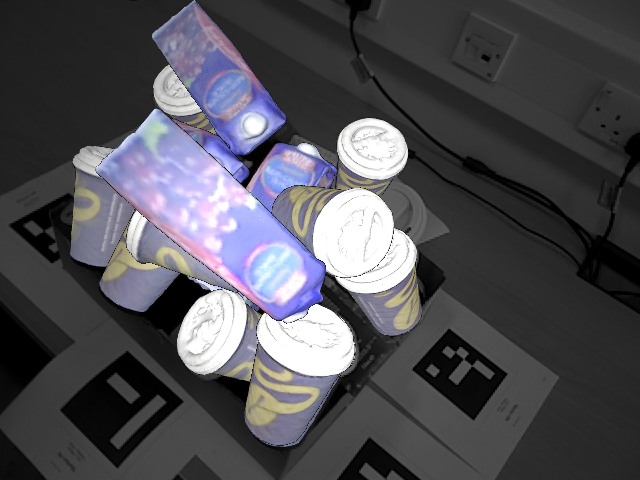}
        \linethickness{2.1pt}
        \put(52, 72){\color{red}\vector(-1.4, -1){12}}
        \put(77, 52){\color{Goldenrod}\vector(-1.4, -1){12}}
        \put(10, 62){\color{red}\vector(1.4, -1){12}}
        \put(13, 27){\color{red}\vector(1.4, 1){12}}
        
        \end{overpic} \\
        
    \end{tabular}
    \vspace{-1mm}
    \caption{
        Qualitative results on IC-BIN~\cite{icbin}.
        Columns show different examples.
        Rows show a comparison against different methods.
        {\color{red}\textbf{Red}} ({\color{Dandelion}\textbf{yellow}}) arrows highlight wrong (missing) predictions. 
        Backgrounds are converted to grayscale for a better contrast.
    }
    \label{fig:supp_qual_icbin}
\end{figure*}

%%%%%%%%%%%%%%%%%%%%%%%%%%%%%%%%%%%%%%%%%%%%%%%%%%%%%%%%%%%%%%%%%%%%%%%
\noindent\textbf{IC-BIN dataset.}~Fig.~\ref{fig:supp_qual_icbin} shows qualitative results on IC-BIN~\cite{icbin}.
It contains multiple instances of two object categories, a box-shaped juice carton and a cylindrical-shaped coffee cup, placed inside a bin.
In column (a), we present a challenging scenario involving heavily-occluded coffee cups. MegaPose and SAM6D incorrectly predict the poses of the most visible objects and miss the occluded ones. \ourmethod accurately predict the poses of all objects except missing the most occluded one (highlighted by the yellow arrow). Regarding one of the cups (indicated by the red arrow), \ourmethod correctly aligns its cylindrical shape but faces challenges with aligning its texture (gold-colored brand logo).
In column (b), we present a scenario involving five instances of the same juice carton. SAM6D makes one incorrect prediction while MegaPose makes two errors and misses one object. \ourmethod outperforms all competitors by accurately estimating the pose of all five objects.
In column (c), we consider both object types simultaneously. MegaPose and SAM6D perform poorly, with three and five errors, respectively, and multiple missing predictions. \ourmethod yields poses with three errors and one missing prediction. Interestingly, for the juice carton in the middle, \ourmethod correctly aligns the portion included in the instance localization prior, which consists only of the circular brand logo.

\begin{figure*}[t]

\centering

    \begin{minipage}{0.32\textwidth}
        \centering
        \footnotesize (a)
    \end{minipage}
    \begin{minipage}{0.32\textwidth}
        \centering
        \footnotesize (b)
    \end{minipage}
    \begin{minipage}{0.32\textwidth}
        \centering
        \footnotesize (c)
    \end{minipage}

    \vspace{1.3 mm}
    \begin{tabular}{@{}c@{\,}c@{\,}c}
    \raggedright

        % first row
        \begin{overpic}[width=0.33\textwidth]
        {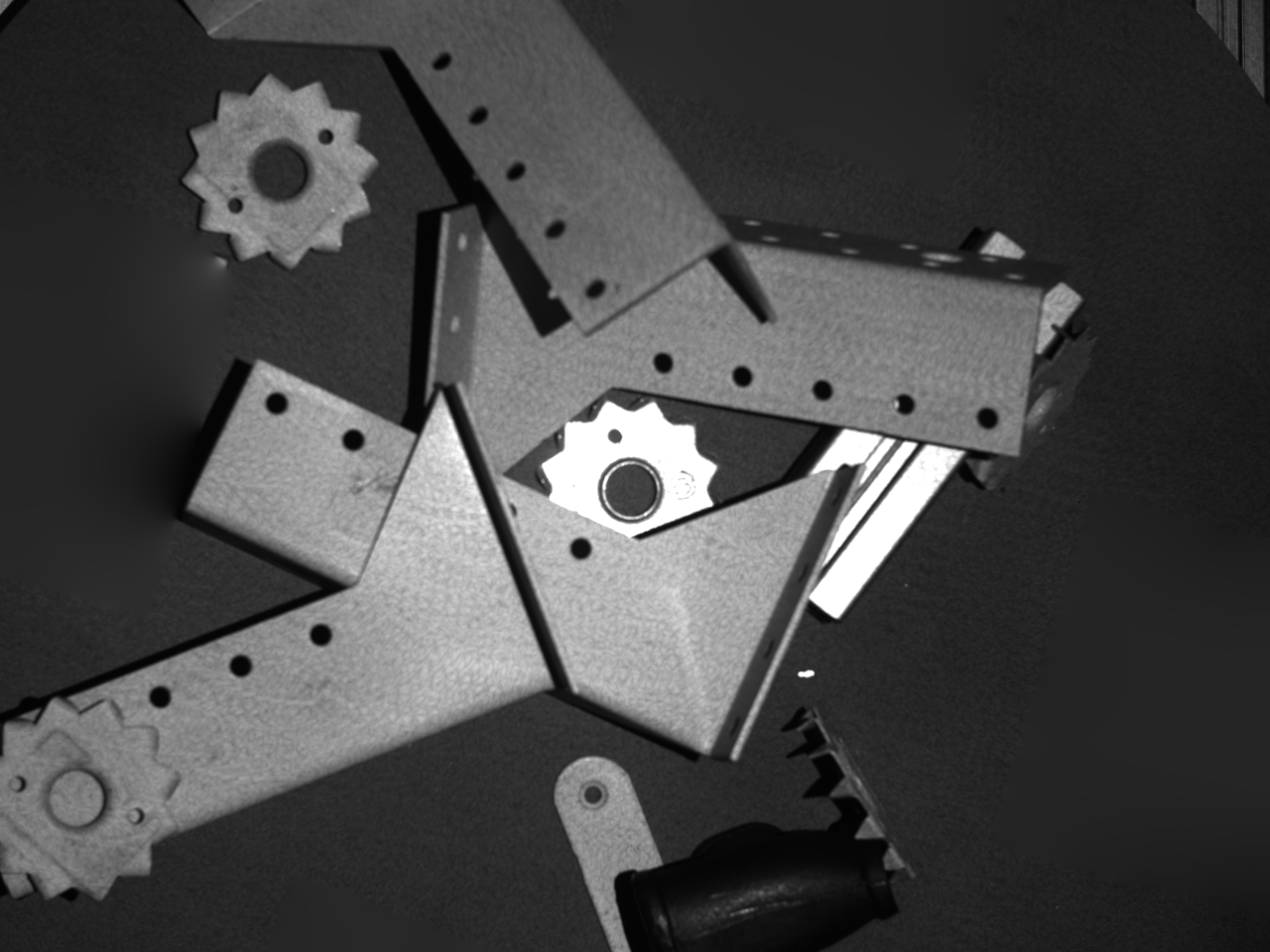}
            \put(-10, 14){\rotatebox{90}{\footnotesize Input images}}
        \end{overpic} &
        \begin{overpic}[width=0.33\textwidth]
        {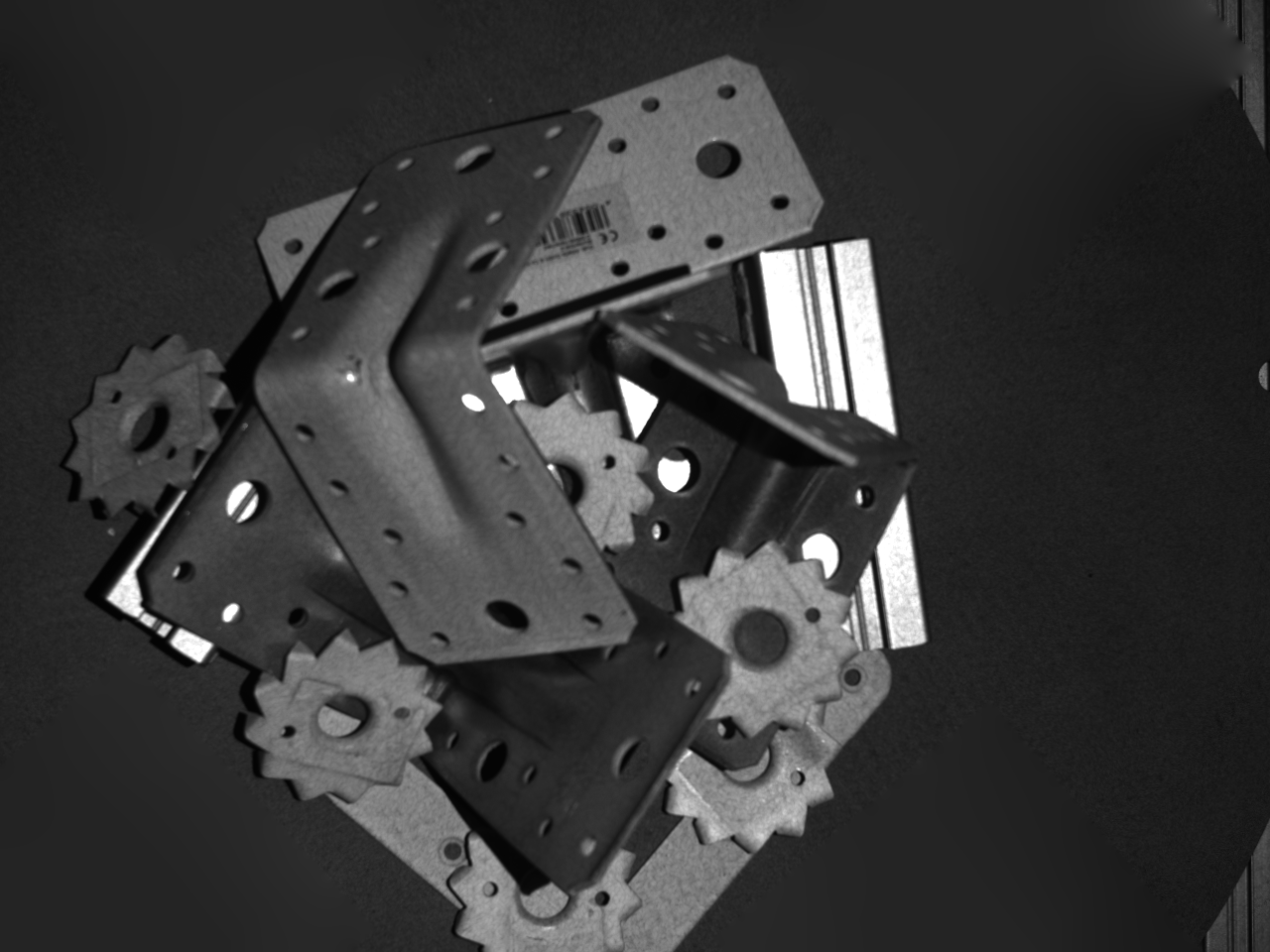}

        \end{overpic} &
        \begin{overpic}[width=0.33\textwidth]
        {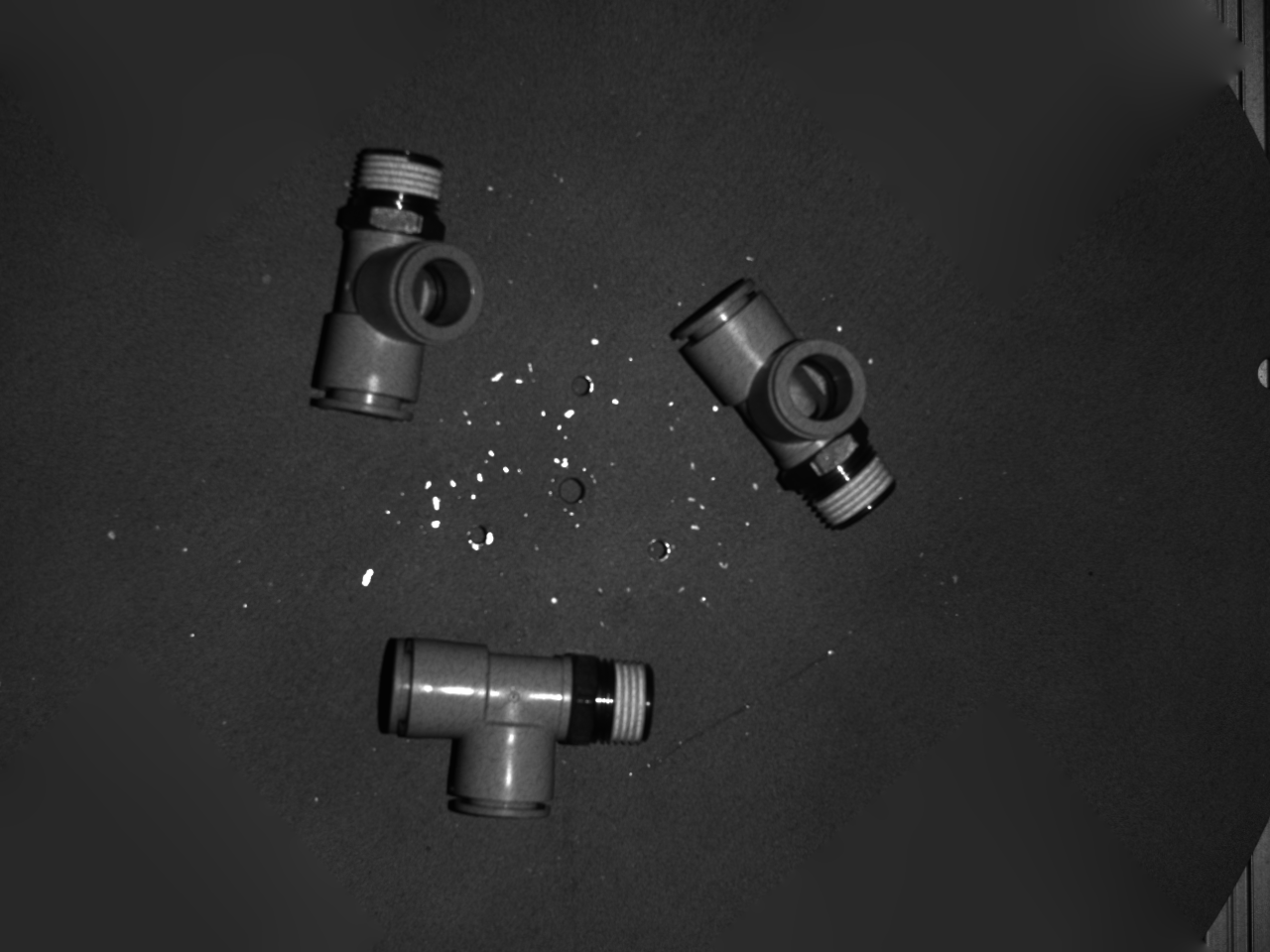}
        \end{overpic}  \\

        % second row
        
        \begin{overpic}[width=0.33\textwidth]
        {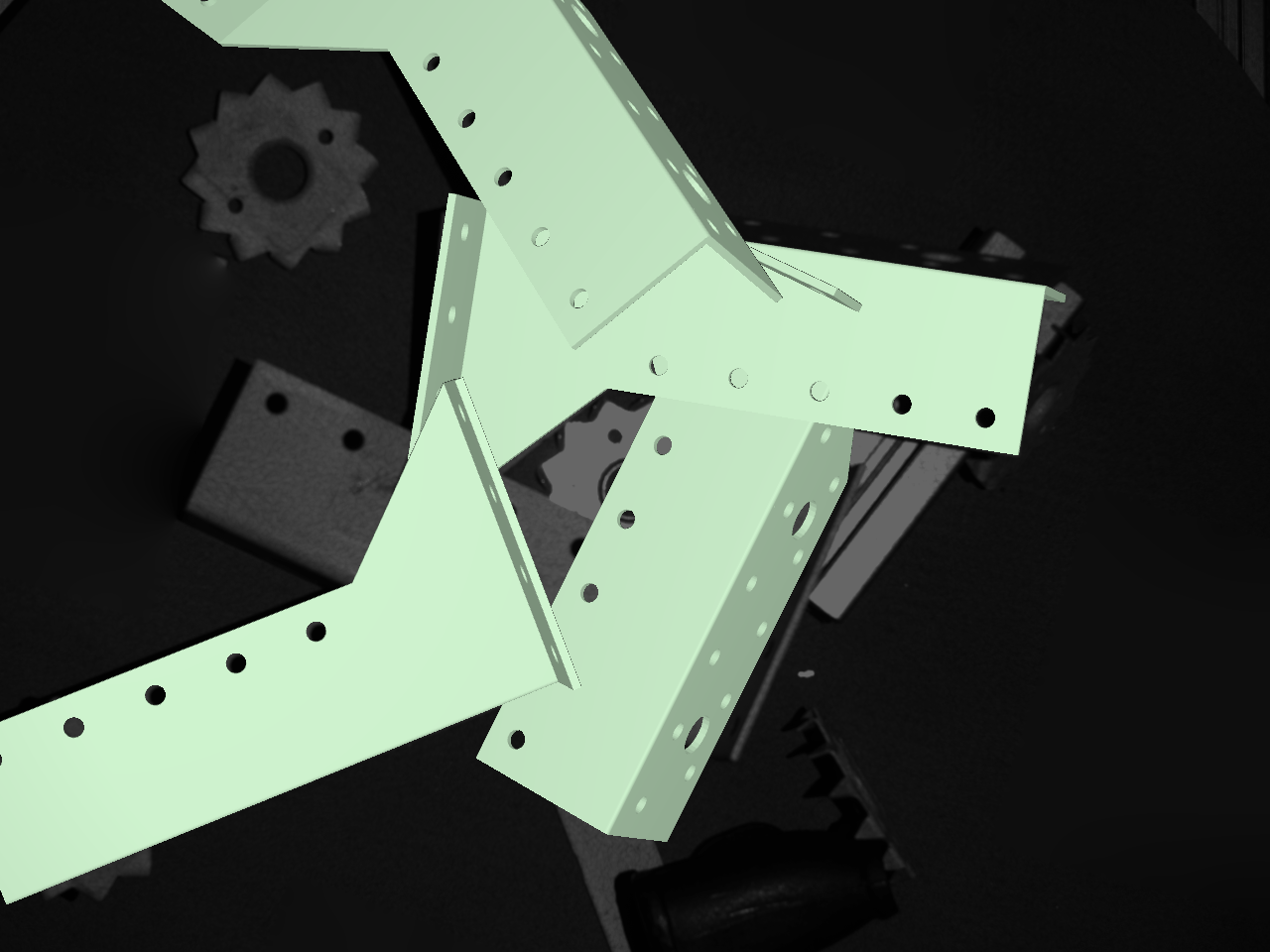}
            \linethickness{2.1pt}
            % \put(5, 51){\color{red}\vector(1.4, -1){12}}
            \put(70, 10){\color{red}\vector(-1.4, 1){12}}
            %\put(76, 62){\color{red}\vector(-1.4, -1){12}}
            \put(-10, 20){\rotatebox{90}{\footnotesize MegaPose}}
        \end{overpic} &
        \begin{overpic}[width=0.33\textwidth]
        {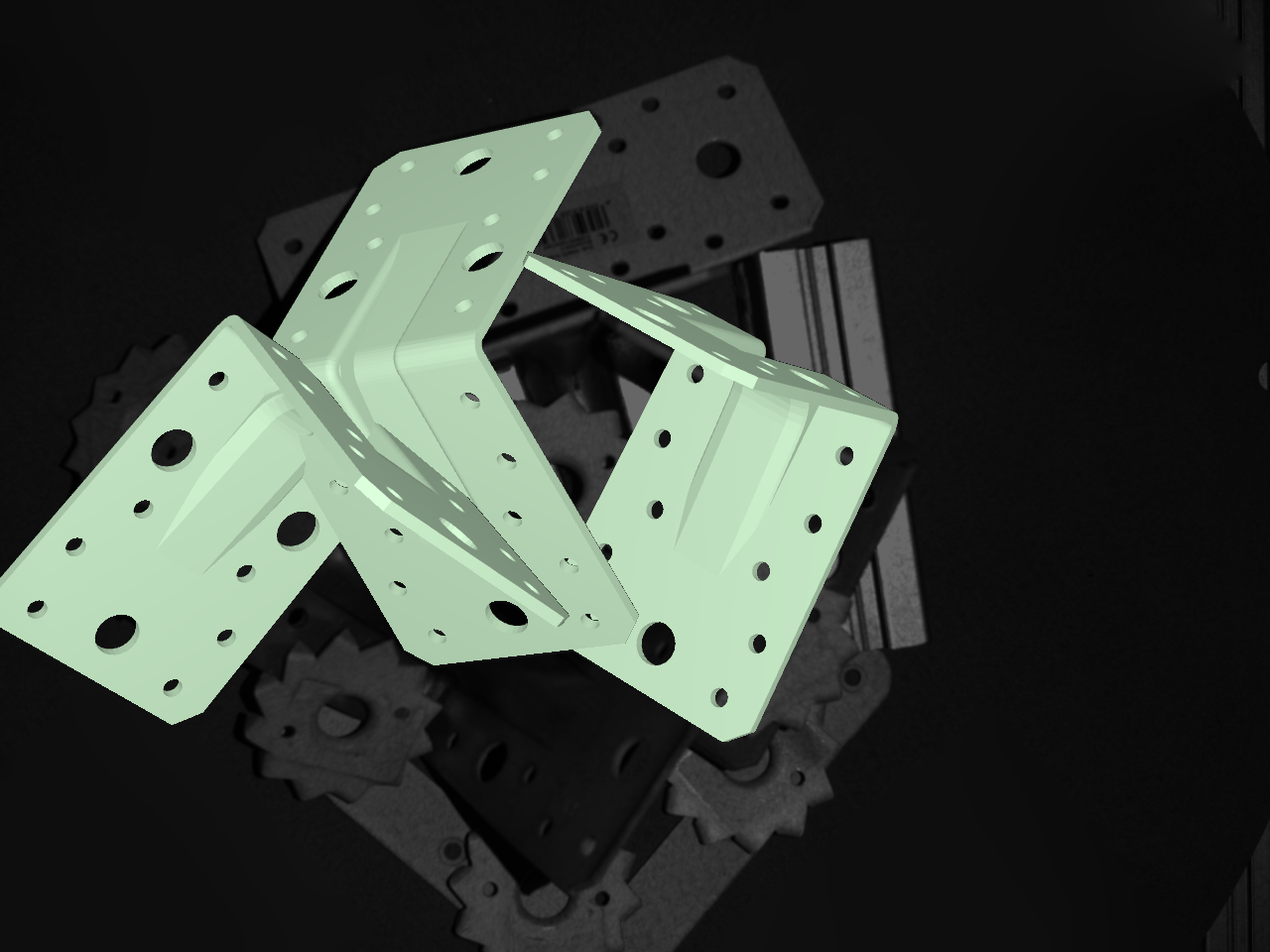}
        \linethickness{2.1pt}
        \put(1, 54){\color{red}\vector(1.4, -1){12}}
        
        \end{overpic} &
        \begin{overpic}[width=0.33\textwidth]
        {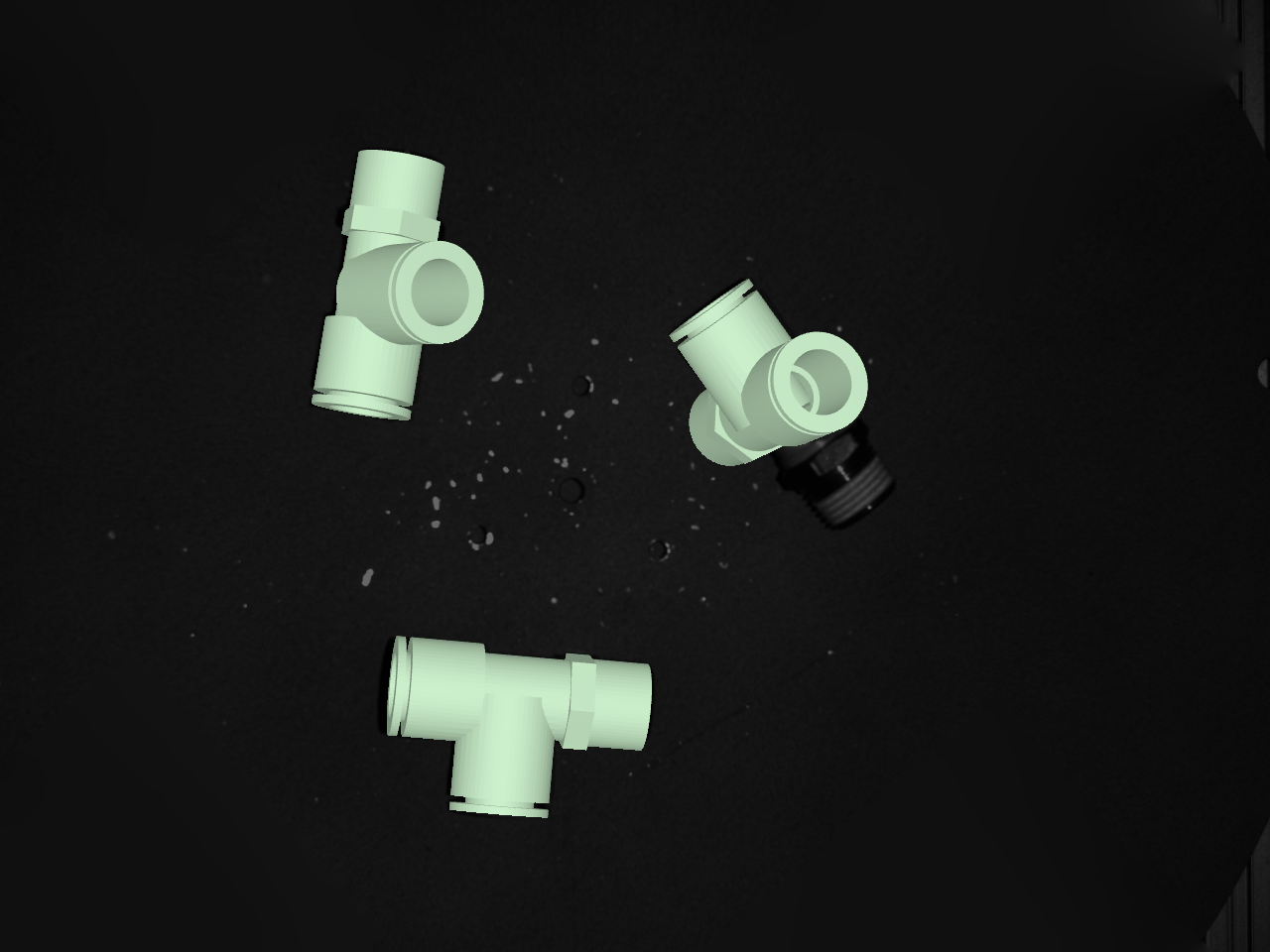}
        \linethickness{2.1pt}
        % \put(82, 25){\color{red}\vector(-1.4, 1){12}}
        \put(79, 57){\color{red}\vector(-1.4, -1){12}}
        
        \end{overpic} \\

        %%%% third row

        \begin{overpic}[width=0.33\textwidth]
        {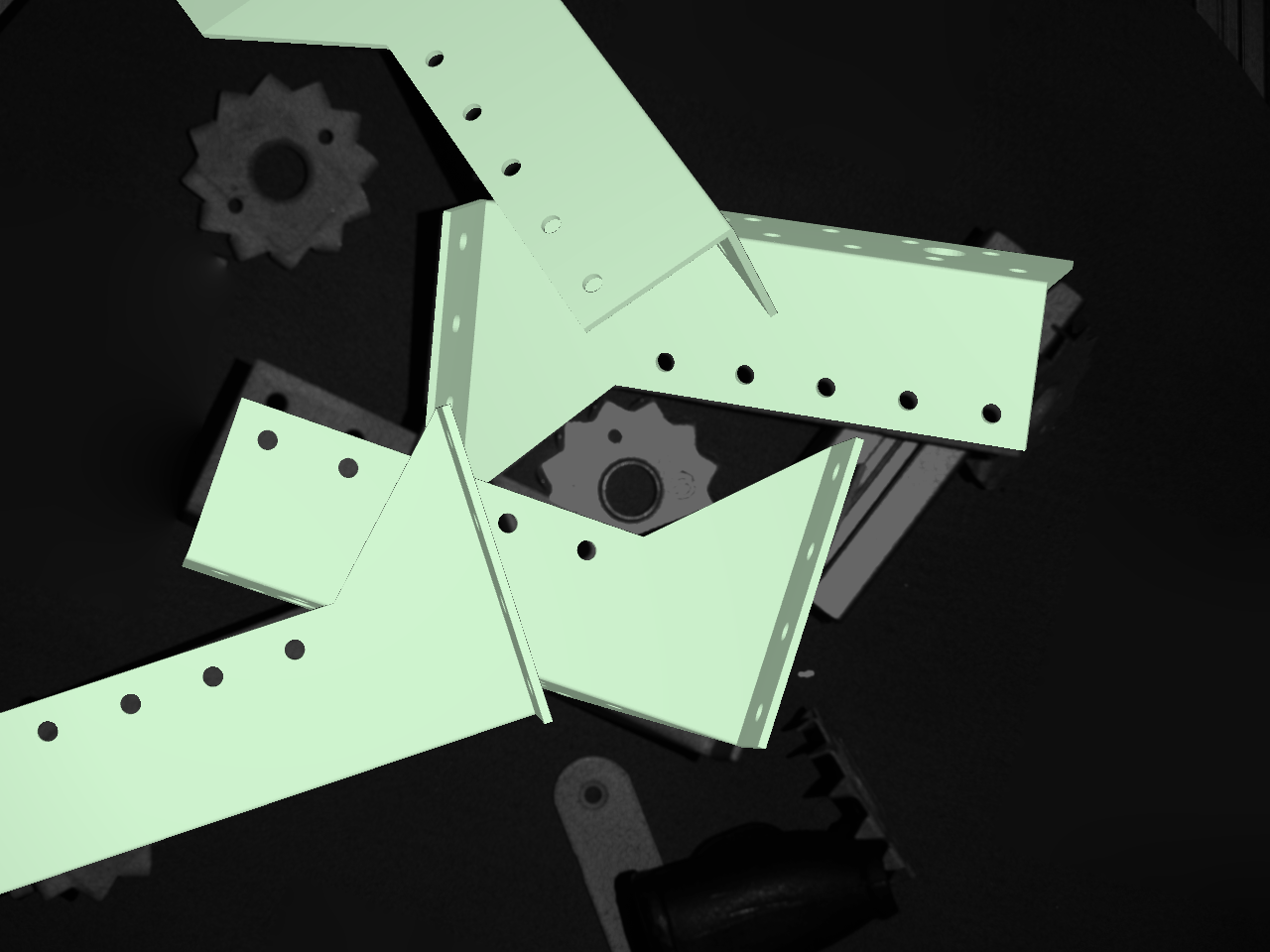}
            \linethickness{2.1pt}
            \put(7, 55){\color{red}\vector(1.4, -1){12}}

            \put(-10, 23){\rotatebox{90}{\footnotesize SAM6D}}
            \end{overpic} &
        \begin{overpic}[width=0.33\textwidth]
        {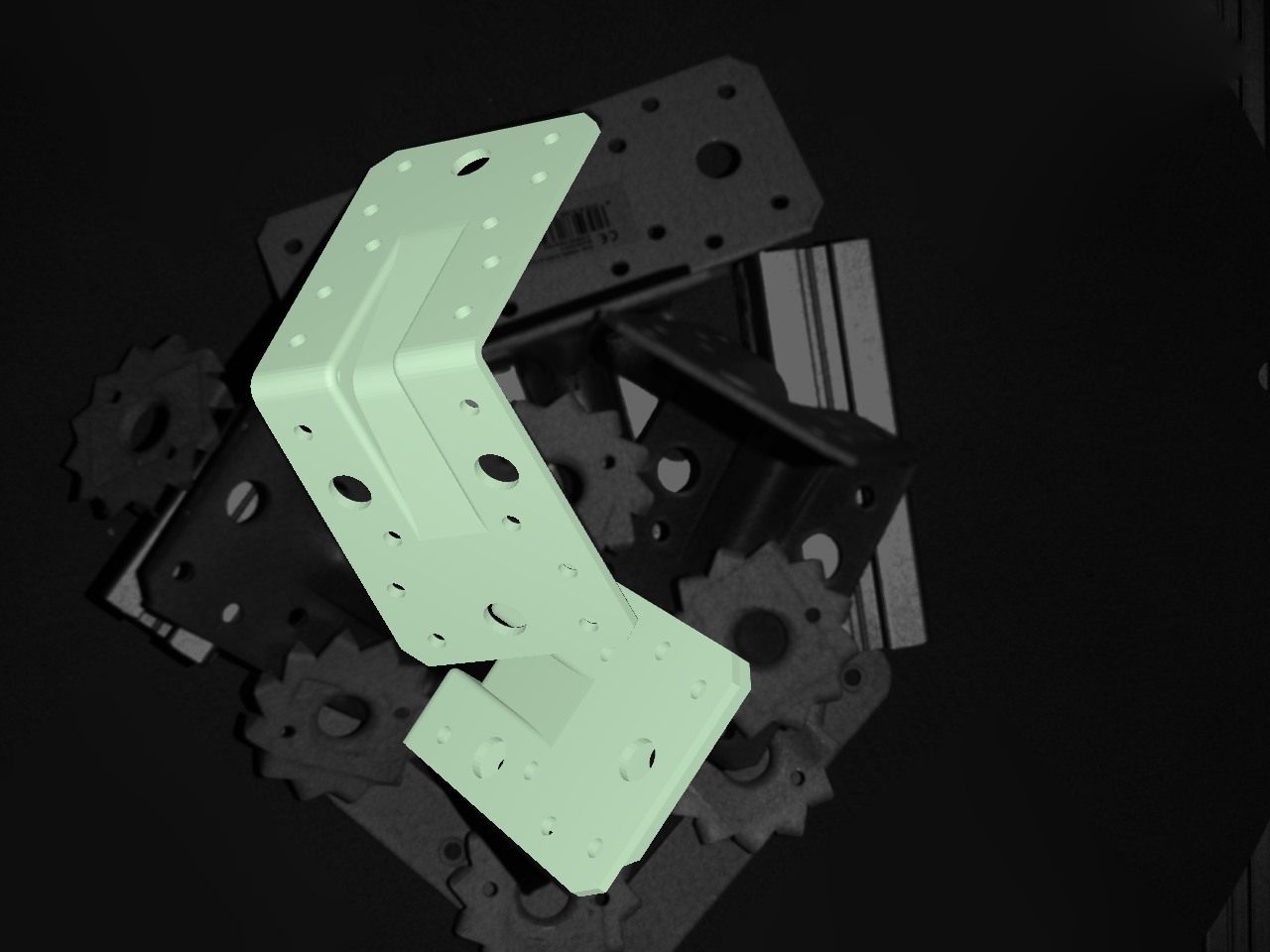}
        \linethickness{2.1pt}
        \put(66, 3){\color{red}\vector(-1.4, 1){12}}
        \put(74, 52){\color{Goldenrod}\vector(-1.4, -1){12}}
        \end{overpic} &
        \begin{overpic}[width=0.33\textwidth]
        {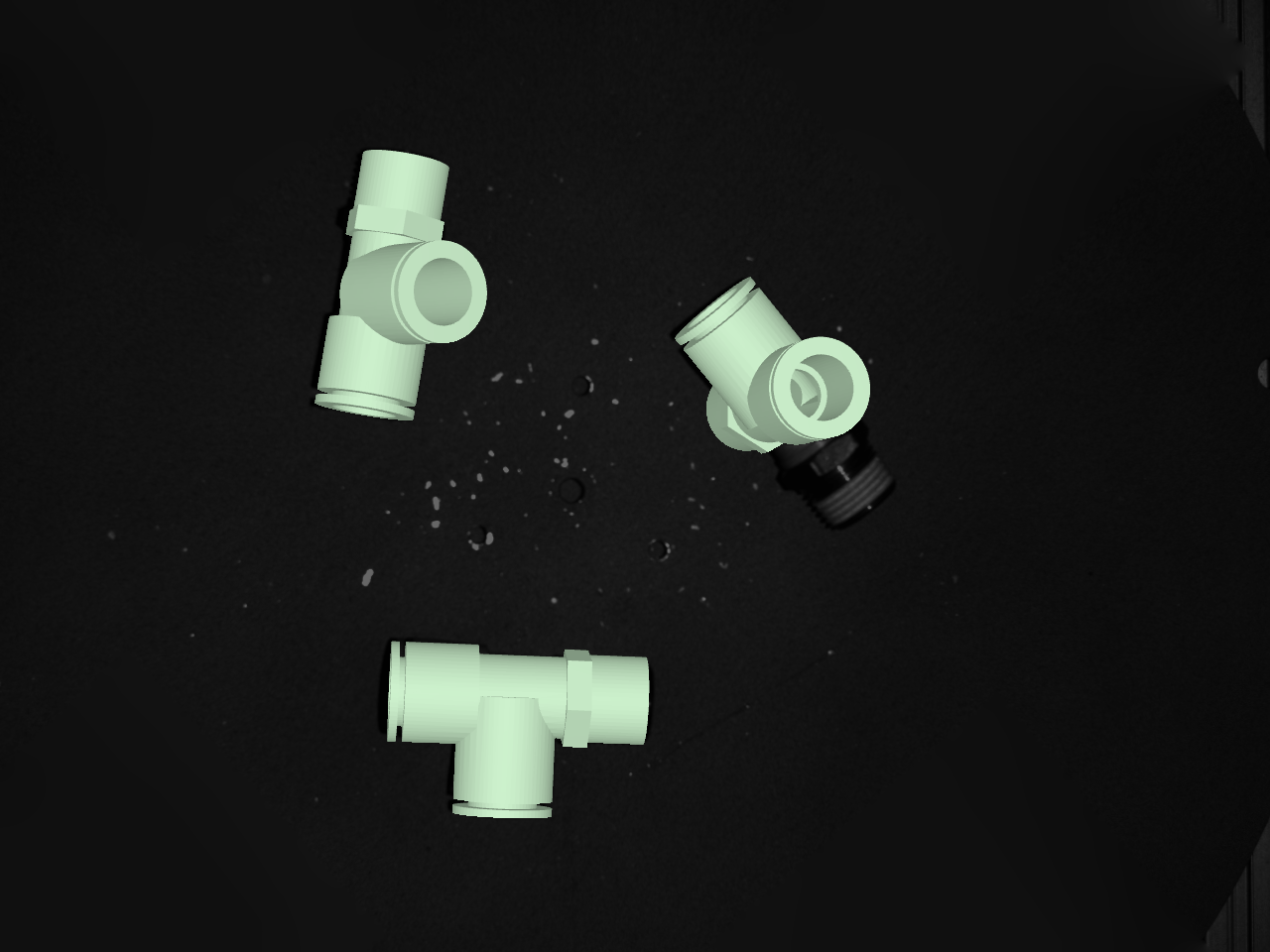}
        \linethickness{2.1pt}
        % \put(82, 25){\color{red}\vector(-1.4, 1){12}}
        \put(79, 57){\color{red}\vector(-1.4, -1){12}}
        \end{overpic} \\

        % fifth row
        \begin{overpic}[width=0.33\textwidth]
        {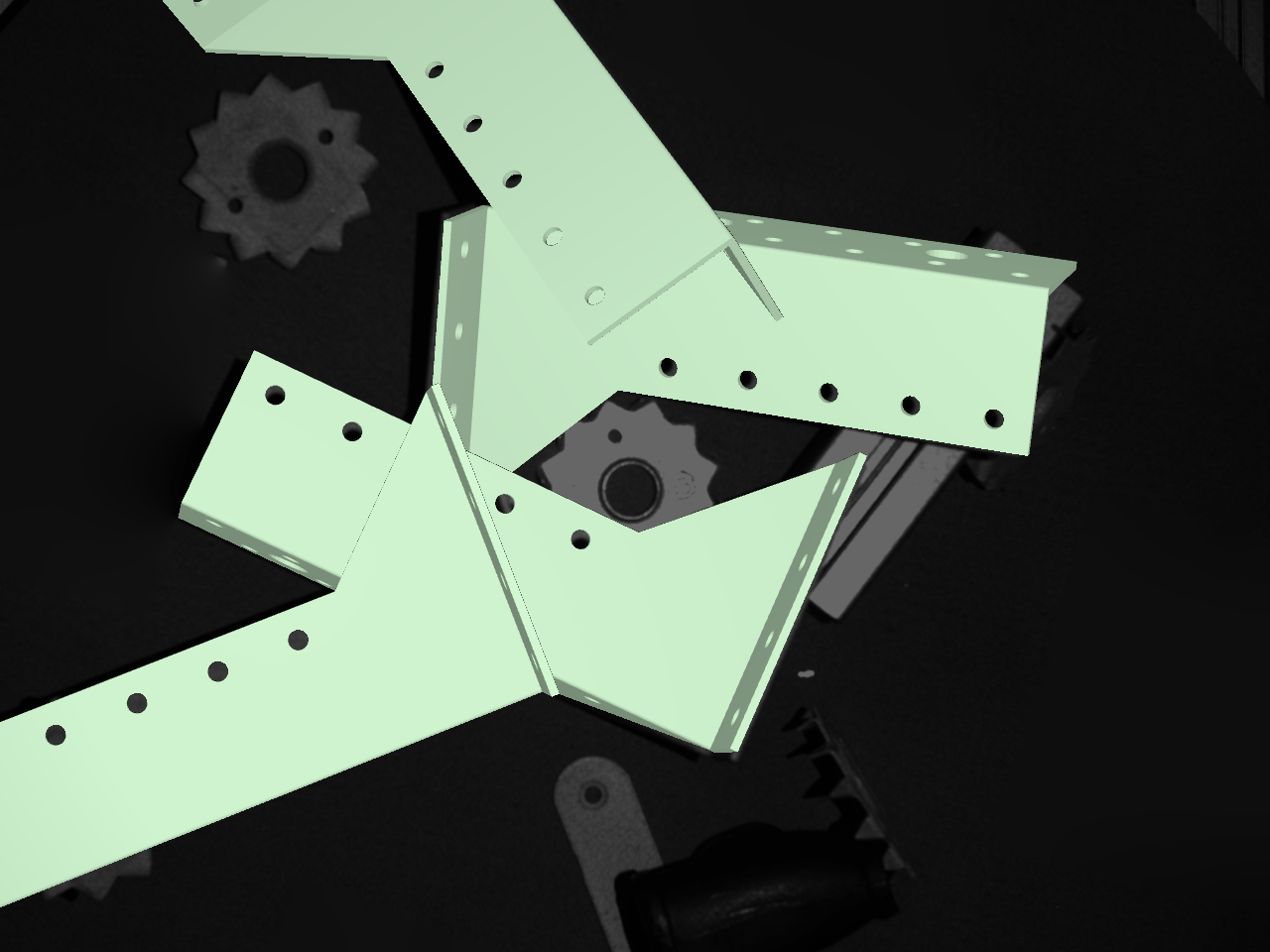}
        \put(-10, 25){\rotatebox{90}{\footnotesize \ourmethod}}
        \end{overpic} &
        \begin{overpic}[width=0.33\textwidth]
        {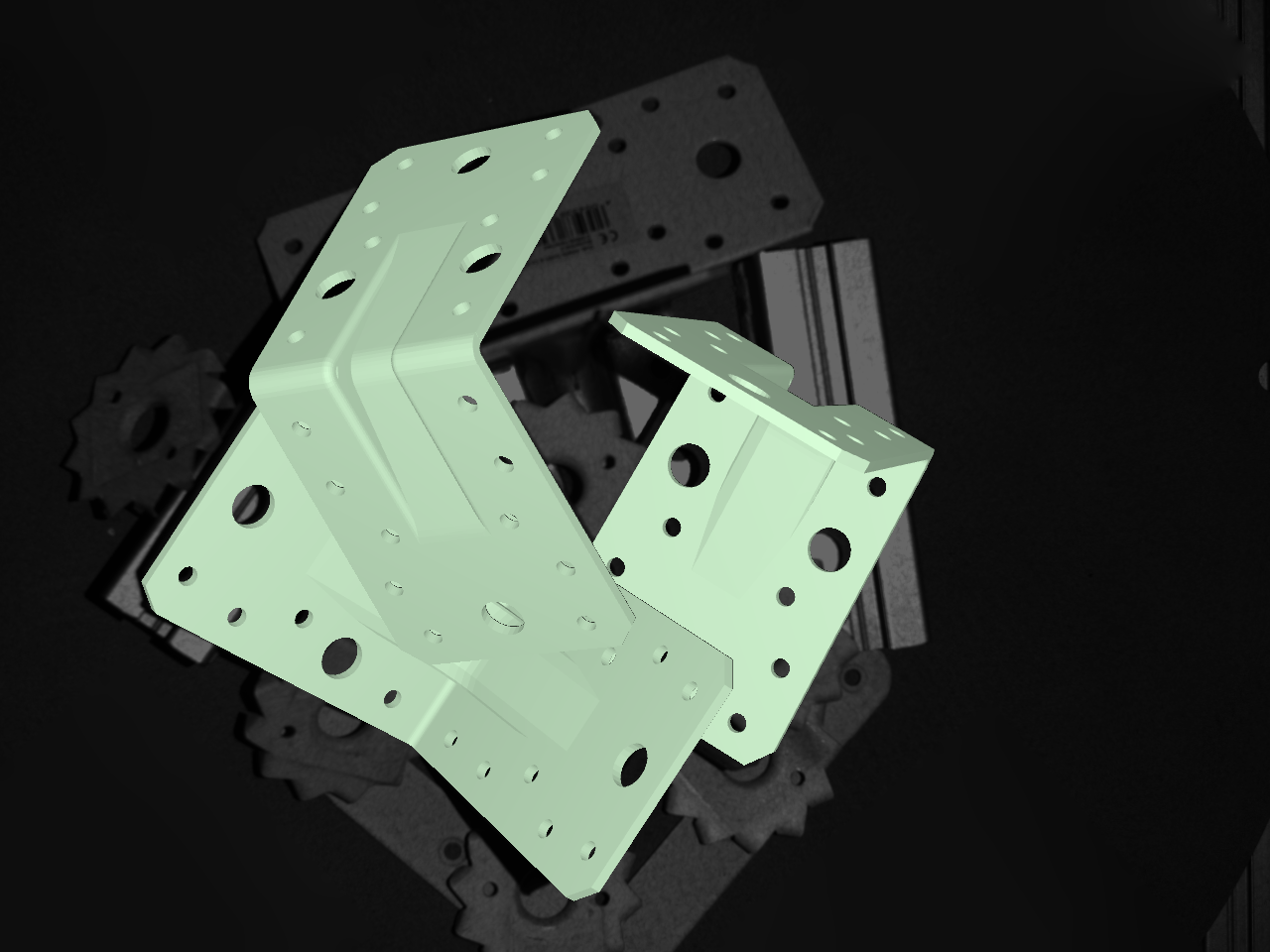}
        \end{overpic} &
        \begin{overpic}[width=0.33\textwidth]
        {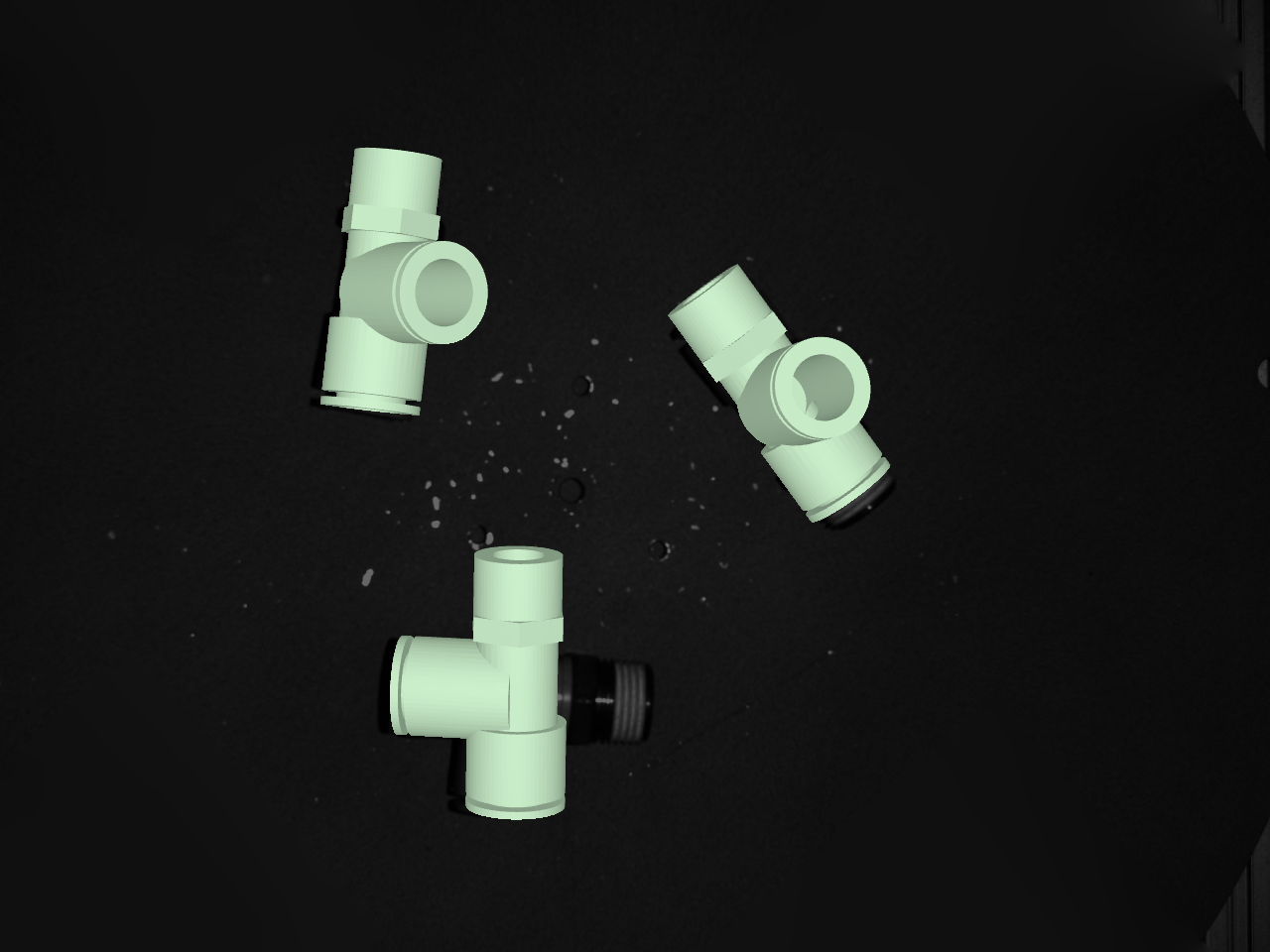}
        \linethickness{2.1pt}
        \put(23, 6){\color{red}\vector(1.4, 1){12}}
        \end{overpic} \\
        
    \end{tabular}
    \vspace{-1mm}
    \caption{
        Qualitative results on ITODD~\cite{itodd}.
        Columns show different examples.
        Rows show a comparison against various methods.
        {\color{red}\textbf{Red}} ({\color{Dandelion}\textbf{yellow}}) arrows highlight wrong (missing) predictions.
        ITODD images are grayscale. The 3D models of texture-less objects are converted to a light-green color for a better contrast.
    }
    \label{fig:supp_qual_itodd}
\end{figure*}

%%%%%%%%%%%%%%%%%%%%%%%%%%%%%%%%%%%%%%%%%%%%%%%%%%%%%%%%%%%%%%%%%%%%%%%
\noindent\textbf{ITODD dataset.}~Fig.~\ref{fig:supp_qual_itodd} shows qualitative results on ITODD~\cite{itodd}.
It features 28 industrial objects captured from multiple views using a grayscale camera.
The 3D models of texture-less objects are colored in light-green for a better contrast.
In column (a), \ourmethod predicts the correct pose for all the objects despite the presence of clutter, while MegaPose and SAM6D struggle on one of them (the one indicated by the red arrow).
In column (b), \ourmethod outperforms all the other methods by predicting the correct pose for all the objects. Instead, SAM6D has the lowest performance, correctly estimating only the pose of the most visible object. In addition to this object, MegaPose also correctly predict the pose of one of the other two less visible objects, even though the pose estimated by MegaPose deviates a bit from the ground-truth one.
In column (c), we present a scene containing three instances of the same object without any occlusions. MegaPose and SAM6D fail to predict the pose of the top-right object, while \ourmethod struggles with predicting the pose of bottom one. The errors of MegaPose, SAM6D and \ourmethod are similar, as they rotate the object by 90 degrees. The estimated pose aligns with the two larger extremities, correctly matching that portion of the surface, while missing alignment with the smaller third extremity.

\begin{figure*}[t]
\centering

    \begin{minipage}{0.32\textwidth}
        \centering
        \footnotesize (a)
    \end{minipage}
    \begin{minipage}{0.32\textwidth}
        \centering
        \footnotesize (b)
    \end{minipage}
    \begin{minipage}{0.32\textwidth}
        \centering
        \footnotesize (c)
    \end{minipage}

    \vspace{1.3 mm}
    \begin{tabular}{@{}c@{\,}c@{\,}c}
    \raggedright

        % first row
        \begin{overpic}[width=0.33\textwidth]{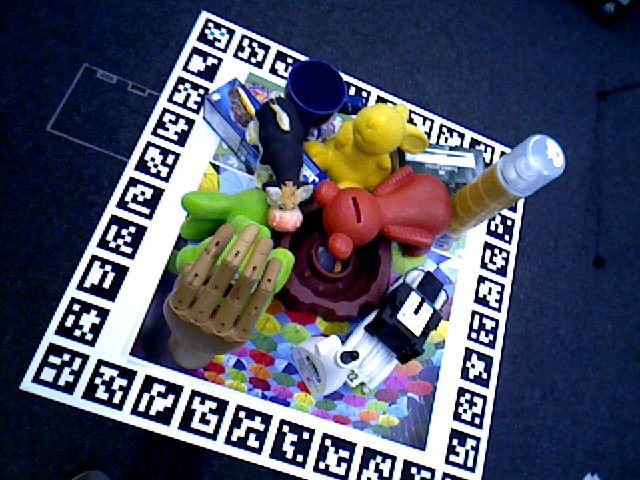}
            \put(-10, 14){\rotatebox{90}{\footnotesize Input images}}
        \end{overpic} &
        \begin{overpic}[width=0.33\textwidth]{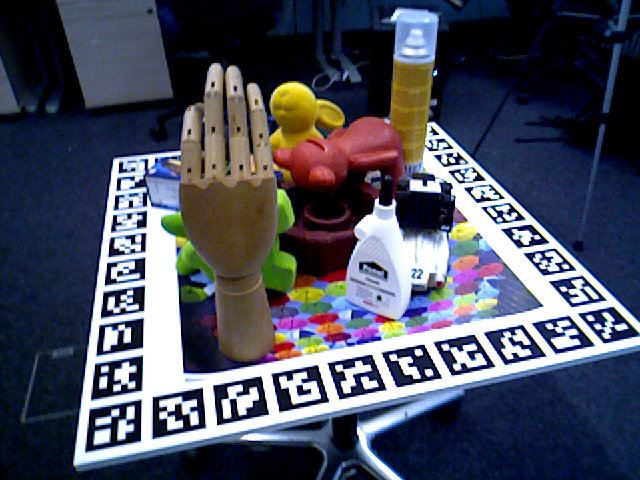}
        \end{overpic} &
        \begin{overpic}[width=0.33\textwidth]{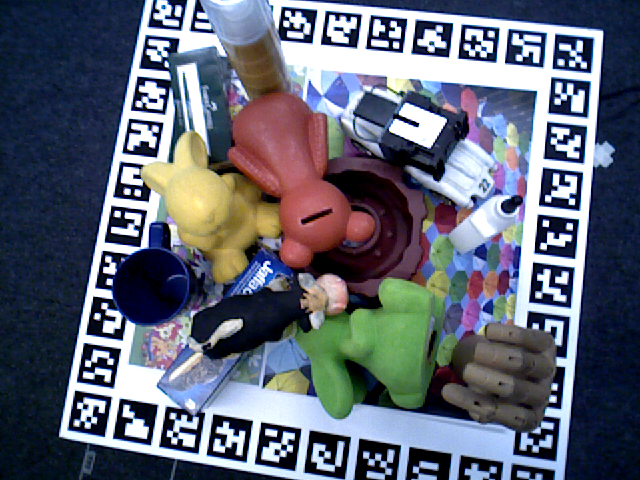}
        \end{overpic} \\

        % second row

        \begin{overpic}[width=0.33\textwidth]{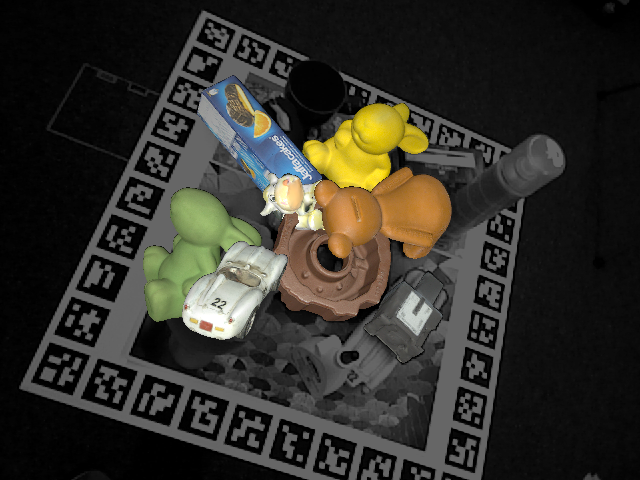}
            \put(-10, 20){\rotatebox{90}{\footnotesize MegaPose}}
            \linethickness{2.1pt}
            \put(29, 53){\color{red}\vector(1.4, -1){12}}
            \put(15, 45){\color{red}\vector(1.4, -1){12}}
            \put(50, 15){\color{red}\vector(-1.4, 1){12}}
        \end{overpic} &
        \begin{overpic}[width=0.33\textwidth]{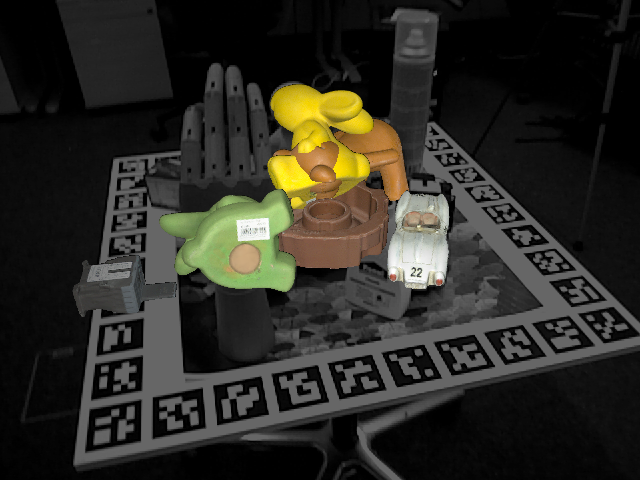}
            \linethickness{2.1pt}
            \put(65, 70){\color{red}\vector(-1.4, -1){12}}
            \put(38, 27){\color{red}\vector(1.4, 1){12}}
            \put(1, 17){\color{red}\vector(1.4, 1){12}}
        \end{overpic} &
        \begin{overpic}[width=0.33\textwidth]{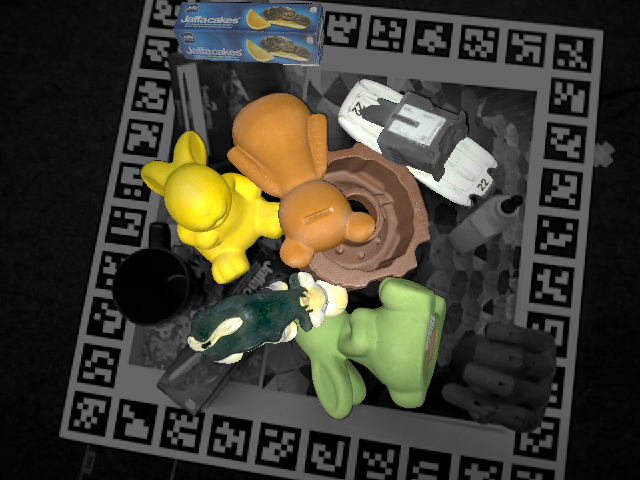}
            \linethickness{2.1pt}
            \put(16, 60){\color{red}\vector(1.4, 1){12}}
            \put(79, 30){\color{red}\vector(-1.4, 1){12}}
        \end{overpic} \\

     % third row

        \begin{overpic}[width=0.33\textwidth]{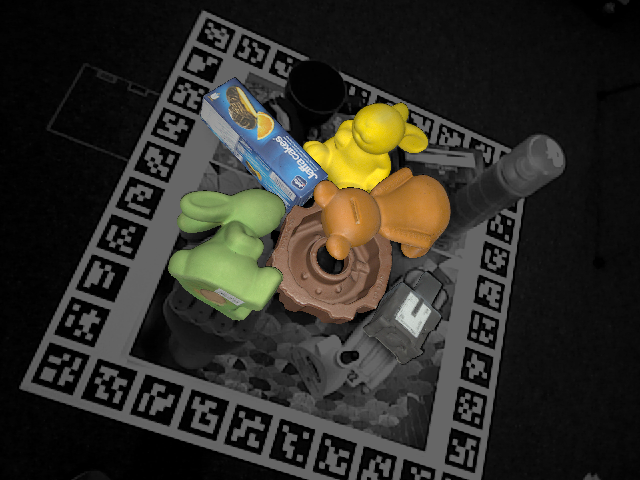}
            \put(-10, 23){\rotatebox{90}{\footnotesize SAM6D}}
            \linethickness{2.1pt}
            \put(57, 65){\color{Goldenrod}\vector(-1.4, -1){12}}
            \put(72, 7){\color{Goldenrod}\vector(-1.4, 1){12}}
        \end{overpic} &
        \begin{overpic}[width=0.33\textwidth]{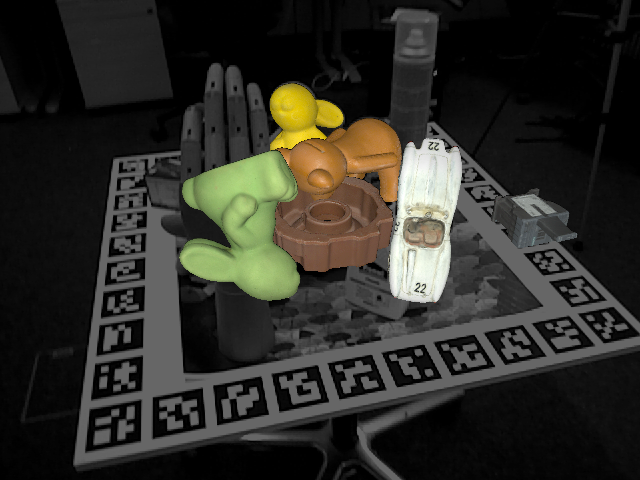}
            \linethickness{2.1pt}
            \put(25, 22){\color{red}\vector(1.4, 1){12}}
            \put(97, 28){\color{red}\vector(-1.4, 1){12}}
            \put(83, 60){\color{red}\vector(-1.4, -1){12}}
        \end{overpic} &
        \begin{overpic}[width=0.33\textwidth]{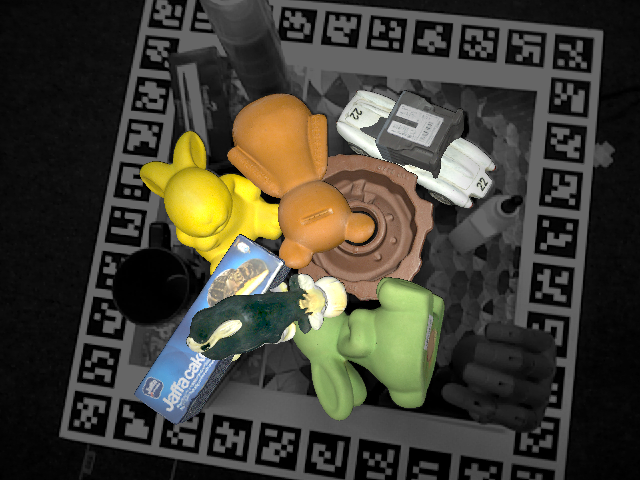}
            \linethickness{2.1pt}
            \put(10, 23){\color{red}\vector(1.4, -1){12}}
        \end{overpic} \\

        % fifth row
        \begin{overpic}[width=0.33\textwidth]{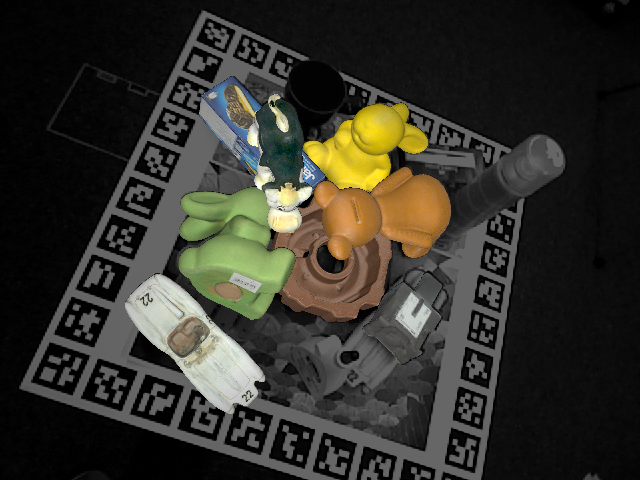}
            \put(-10, 25){\rotatebox{90}{\footnotesize \ourmethod}}
            \linethickness{2.1pt}
            \put(15, 10){\color{red}\vector(1.4, 1){12}}
        \end{overpic} &
        \begin{overpic}[width=0.33\textwidth]{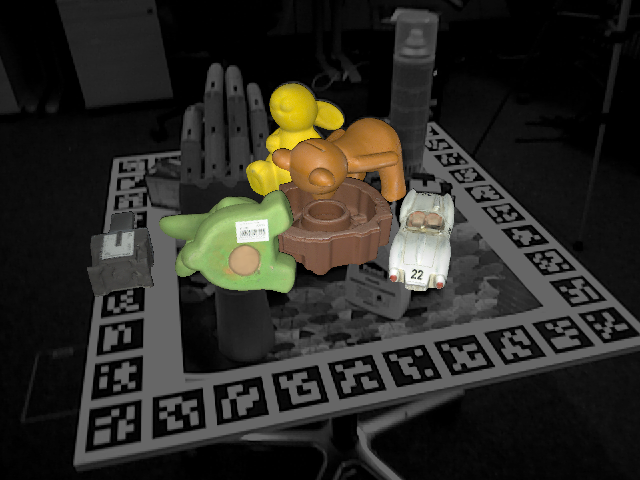}
            \linethickness{2.1pt}
            \put(2, 22){\color{red}\vector(1.4, 1){12}}
        \end{overpic} &
        \begin{overpic}[width=0.33\textwidth]{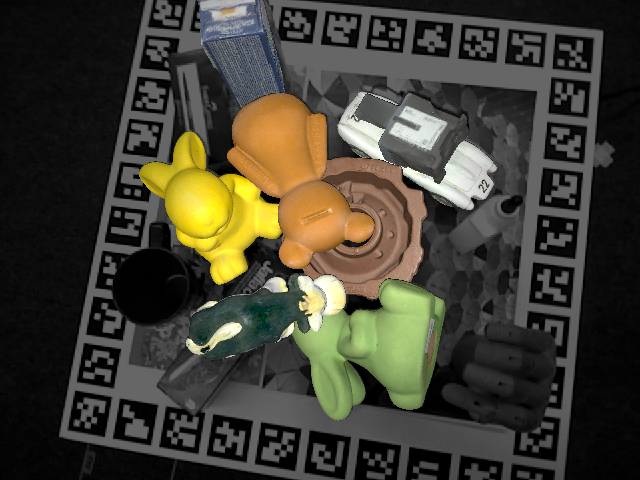}
            \linethickness{2.1pt}
            \put(19.5, 60){\color{red}\vector(1.4, 1){12}}
        \end{overpic} \\
        
    \end{tabular}
    \vspace{-1mm}
    \caption{
        Qualitative results on HB~\cite{hb}.
        Columns show different examples.
        Rows show a comparison against different methods.
        {\color{red}\textbf{Red}} ({\color{Dandelion}\textbf{yellow}}) arrows highlight wrong (missing) predictions.
        Backgrounds are converted to grayscale for a better contrast.
    }
    \label{fig:supp_qual_hb}
\end{figure*}

%%%%%%%%%%%%%%%%%%%%%%%%%%%%%%%%%%%%%%%%%%%%%%%%%%%%%%%%%%%%%%%%%%%%%%%
\noindent\textbf{HB dataset.}~Fig.~\ref{fig:supp_qual_hb} shows qualitative results on HB~\cite{hb}. 
It contains toys, industrial, and household items arranged in highly cluttered scenes.
In column (a), MegaPose performs poorly, by wrongly estimating the poses of three out of seven objects, i.e. the black-white cow, the green rabbit, and the white car. SAM6D correctly estimates the poses of five objects but misses two (indicated by yellow arrows). \ourmethod perform relatively good by correctly predicting the poses of six objects and missing one (the white car).
In column (b), all competitors exhibit poor performance, each making three incorrect predictions. In contrast, \ourmethod outperforms them by making only a single inaccurate prediction for the black box object (red arrow). 
In column (c), all other methods struggle with the blue ``Jaffa cakes'' box: SAM6D successfully locates the object but flips its pose by 180 degrees, whereas both MegaPose and \ourmethod fail to even localize it.

\begin{figure*}[t]
\centering

    \raggedright
    \begin{minipage}{0.32\textwidth}
        \centering
        \footnotesize (a)
    \end{minipage}
    \begin{minipage}{0.32\textwidth}
        \centering
        \footnotesize (b)
    \end{minipage}
    \begin{minipage}{0.32\textwidth}
        \centering
        \footnotesize (c)
    \end{minipage}

    \vspace{1.3 mm}
    \begin{tabular}{@{}c@{\,}c@{\,}c}
    \raggedright

        \begin{overpic}[width=0.33\textwidth]
        {supp/images/qualitatives/ycbv/001571_input.png}
            \put(-10, 14){\rotatebox{90}{\footnotesize Input images}}
        \end{overpic} &
        \begin{overpic}[width=0.33\textwidth]
        {supp/images/qualitatives/ycbv/000077_input.png}
        \end{overpic} &
        \begin{overpic}[width=0.33\textwidth]
        {supp/images/qualitatives/ycbv/000139_input.png}
        \end{overpic}  \\

        \begin{overpic}[width=0.33\textwidth]
        {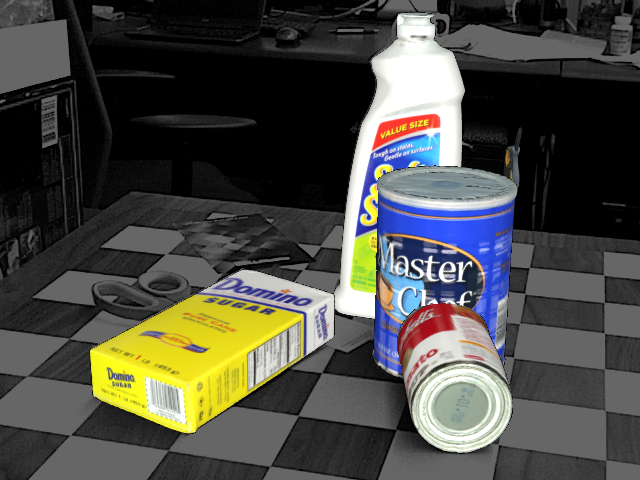}
            \put(-10, 20){\rotatebox{90}{\footnotesize MegaPose}}
            \linethickness{2.1pt}
            \put(94,60){\color{red}\vector(-1.4,-1){12}}
        \end{overpic} &
        \begin{overpic}[width=0.33\textwidth]
        {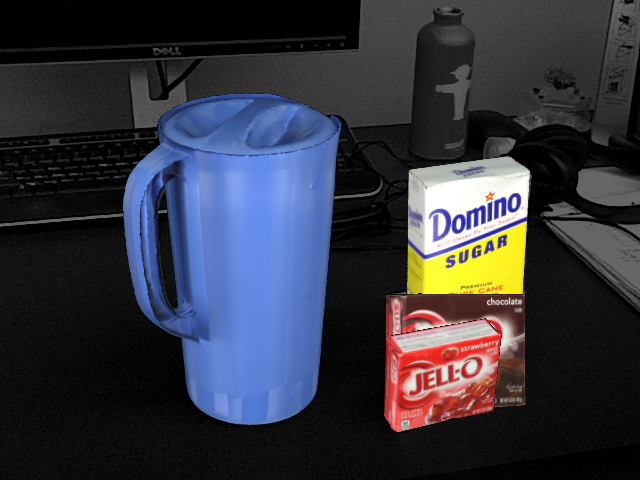}
                \linethickness{2.1pt}
        \put(95,36){\color{red}\vector(-1.4,-1){12}}
        \end{overpic} &
        \begin{overpic}[width=0.33\textwidth]
        {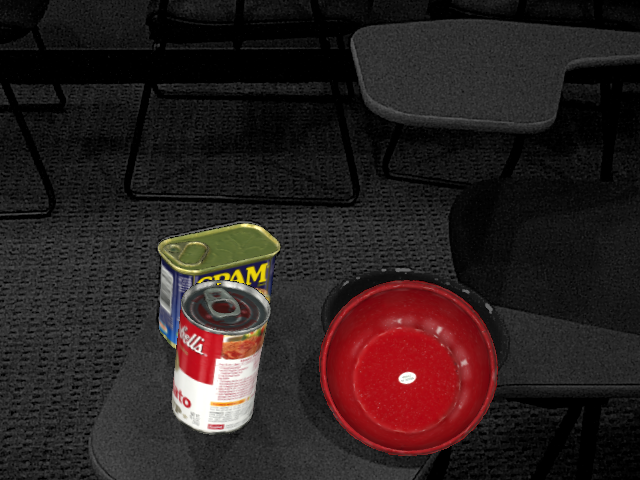}
        \linethickness{2.1pt}
        \put(88,35){\color{red}\vector(-1.4,-1){12}}
        \end{overpic} \\

        \begin{overpic}[width=0.33\textwidth]
        {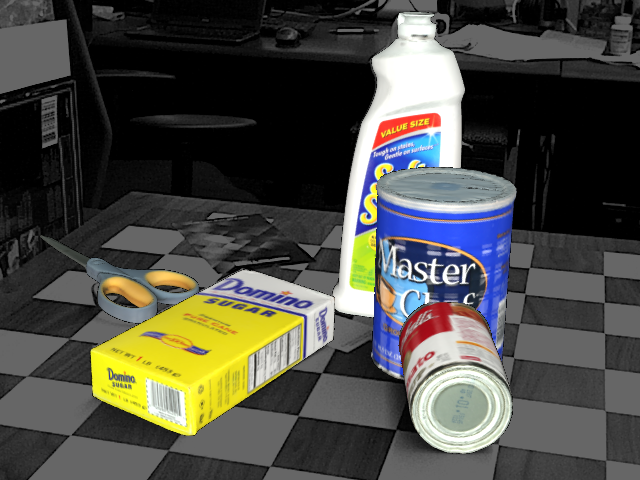}
            \put(-10, 23){\rotatebox{90}{\footnotesize SAM6D}}
            \linethickness{2.1pt}
            \put(30,43){\color{red}\vector(-1.4,-1){12}}
        \end{overpic} &
        \begin{overpic}[width=0.33\textwidth]
        {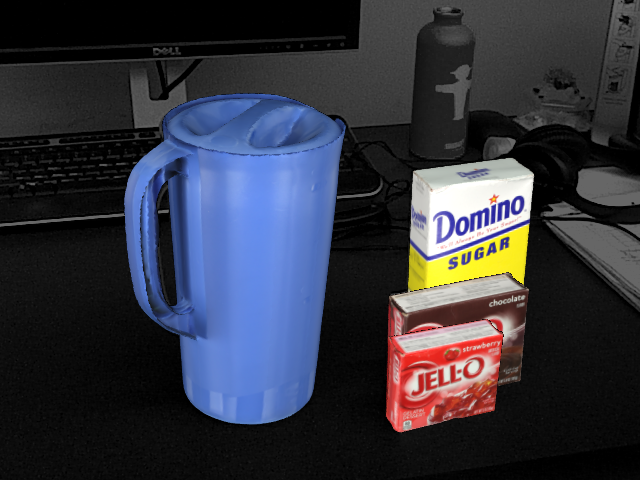}
        \end{overpic} &
        \begin{overpic}[width=0.33\textwidth]
        {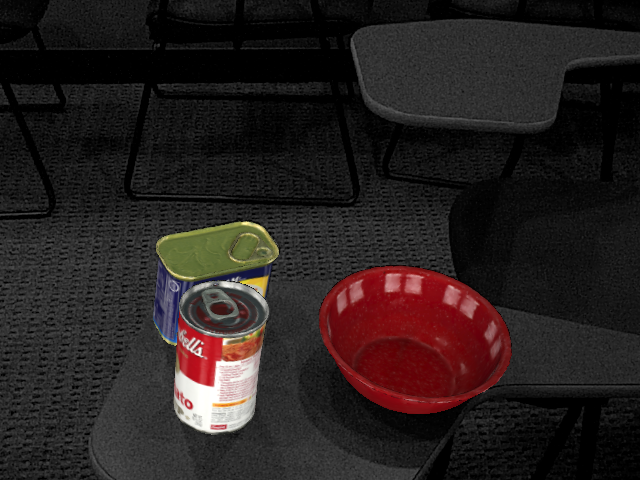}
        \linethickness{2.1pt}
        \put(55,40){\color{red}\vector(-1.4,-0.8){12}}
        \end{overpic} \\

        \begin{overpic}[width=0.33\textwidth]
        {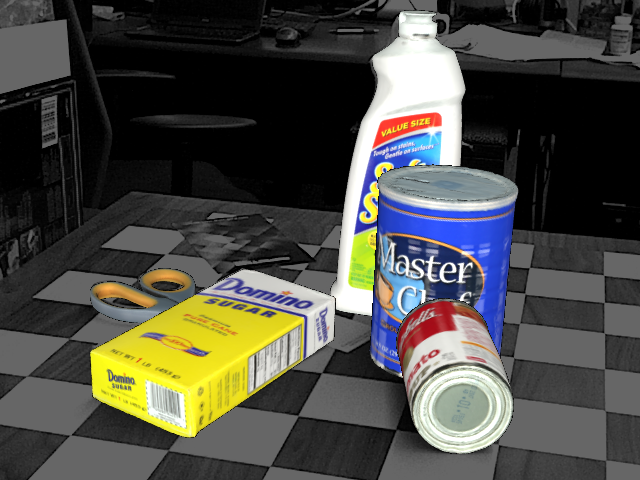}
            \put(-10, 25){\rotatebox{90}{\footnotesize \ourmethod}}
        \end{overpic} &
        \begin{overpic}[width=0.33\textwidth]
        {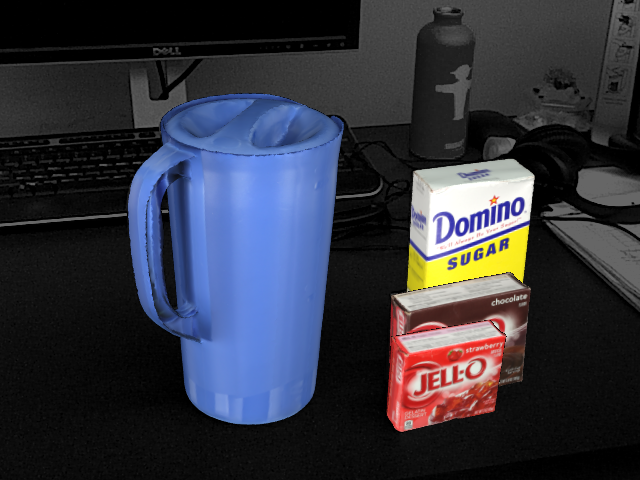}
        \end{overpic} &
        \begin{overpic}[width=0.33\textwidth]
        {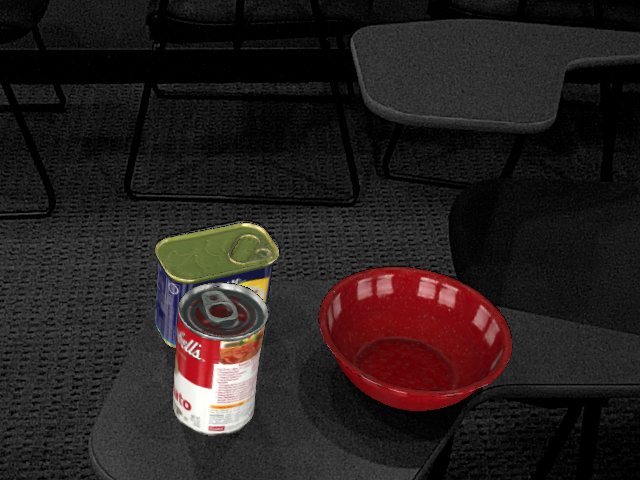}
        \linethickness{2.1pt}
        \put(55,40){\color{red}\vector(-1.4,-0.8){12}}
        \end{overpic} \\
        
    \end{tabular}
    \vspace{-1mm}
    \caption{
        Qualitative results on YCB-V~\cite{ycbv}.
        Columns show different examples.
        Rows show a comparison against different methods.
        {\color{red}\textbf{Red}} arrows highlight wrong predictions.
        Backgrounds are converted to grayscale for a better contrast.
    }
    \label{fig:supp_qual_ycbv}
\end{figure*}

%%%%%%%%%%%%%%%%%%%%%%%%%%%%%%%%%%%%%%%%%%%%%%%%%%%%%%%%%%%%%%%%%%%%%%%
\noindent\textbf{YCB-V dataset.}~Fig.~\ref{fig:supp_qual_ycbv} shows qualitative results on YCB-V~\cite{ycbv}.
Columns show scenarios where \ourmethod performs equivalently (a), surpasses (b), or falls short (c) compared to the other methods.
In column (a), MegaPose and SAM6D incorrectly predict the pose of the heavily-occluded scissors, whereas \ourmethod successfully determine the correct pose for every object.
In column (b), MegaPose encounter difficulties in determining the pose of the brown box, heavily occluded by the red box. Conversely, SAM6D and \ourmethod predict the correct pose for all objects despite the occlusions.
In column (c), MegaPose predicts the wrong pose for the red bowl, despite the object being un-occluded. Both SAM6D and \ourmethod struggle with the potted meat can, flipping the pose by 180 degrees. This discrepancy is evident in the mismatched side texture compared to the input image (highlighted by the red arrow, where the front texture lacks the ``Spam'' logo). However, this case is particularly interesting because of the object's 3D model inconsistency with the real product: the opening tab is positioned in a different location. We believe that both SAM6D and \ourmethod prioritize aligning this visible geometric detail over the heavily-occluded side texture.

\end{document}